\documentclass[10pt,twocolumn,letterpaper]{article}

\usepackage{wacv}
\makeatletter
\@namedef{ver@everyshi.sty}{}
\makeatother

\usepackage{times}
\usepackage{epsfig}
\usepackage{graphicx}
\usepackage{amsmath}
\usepackage{tikz}
\usepackage{amssymb}
\usepackage[export]{adjustbox}
\usepackage{booktabs}
\usepackage{subcaption}
\usepackage{xcolor}
\usepackage{tabularx}
\usepackage{lipsum}
\usepackage{multirow}
\usepackage{numprint}
\usepackage{stmaryrd}
\usepackage{authblk}
\usepackage[page]{appendix}

\definecolor{airforceblue}{rgb}{0.7, 0.7, 0.04}
\definecolor{bostonuniversityred}{rgb}{0.8, 0.0, 0.0}

\wacvfinalcopy

\ifwacvfinal
\def\assignedStartPage{1} 
\fi

\ifwacvfinal
\usepackage[breaklinks=true,bookmarks=false]{hyperref}
\else
\usepackage[pagebackref=true,breaklinks=true,colorlinks,bookmarks=false]{hyperref}
\fi

\ifwacvfinal
\else
\fi

\newcommand\tikzmark[2]{
\tikz[remember picture,baseline] \node[above, outer sep=0em, inner sep=0.5em] (#1){#2};%
}

\newcommand\linkwithlabel[3]{
	\begin{tikzpicture}[remember picture, overlay, >=stealth, shift={(0,0)}]
		\draw[thick,color=bostonuniversityred,->] (#1) -- (#2) node [midway,fill=white,color=white,text=bostonuniversityred] {#3};
	\end{tikzpicture}
}

\newcommand\linkyellowwithlabel[3]{
	\begin{tikzpicture}[remember picture, overlay, >=stealth, shift={(0,0)}]
		\draw[thick,color=airforceblue,->] (#1) -- (#2) node [midway,fill=white,color=white,text=airforceblue] {#3};
	\end{tikzpicture}
}

\usepackage{array}
\newcolumntype{H}{>{\setbox0=\hbox\bgroup}c<{\egroup}@{}}

\begin{document}

\title{Leveraging Local Domains for Image-to-Image Translation}

\author[1]{Anthony Dell'Eva}
\author[1,2]{Fabio Pizzati}
\author[3]{Massimo Bertozzi}
\author[2]{Raoul de Charette}
\affil[1]{VisLab, Parma, Italy}
\affil[2]{Inria, Paris, France}
\affil[3]{University of Parma, Parma, Italy}
\date{}  
\affil[ ]{{\tt\small anthony.delleva2@unibo.it}}
\affil[ ]{{\tt\small \{fabio.pizzati, raoul.de-charette\}@inria.fr}}
\affil[ ]{{\tt\small bertozzi@ce.unipr.it}}

\maketitle

\begin{abstract}
Image-to-image (i2i) networks struggle to capture local changes because they do not affect the global scene structure. 
For example, translating from highway scenes to offroad, i2i networks easily focus on global color features but ignore obvious traits for humans like the absence of lane markings.

In this paper, we leverage human knowledge about spatial domain characteristics which we refer to as 'local domains' and demonstrate its benefit for image-to-image translation. 
Relying on a simple geometrical guidance, we train a patch-based GAN on few source data and hallucinate a new unseen domain which subsequently eases transfer learning to target.

We experiment on three tasks ranging from unstructured environments to adverse weather. 
Our comprehensive evaluation setting shows we are able to generate realistic translations, with minimal priors, and training only on a few images. 
Furthermore, when trained on our translations images we show that all tested proxy tasks are significantly improved, without ever seeing target domain at training.
\end{abstract}

\section{Introduction}
Apart from their appealing translations, image-to-image~(i2i) GAN networks also offer an alternative to the supervised-learning paradigm. 
Indeed, as translations share features characteristics with the target domain they can be used to fine-tune proxy tasks, reducing the need for target annotations. 
However, i2i GANs perform well at learning global scene changes -- winter$\mapsto$summer, {paints}, etc.~\cite{liu2017unsupervised,zhu2017unpaired}, -- 
but struggle to learn subtle local changes. %
Instead, we leverage human domain knowledge to guide i2i and improve proxy tasks on target, \textit{without seeing target images}.
This is of paramount importance for real-world applications like autonomous driving~\cite{schutera2020night,bruls2019generating,romera2019bridging} which must operate safely in all hazardous conditions -- some of which are rarely observed.

\begin{figure}
	\centering
	\includegraphics[width=\linewidth]{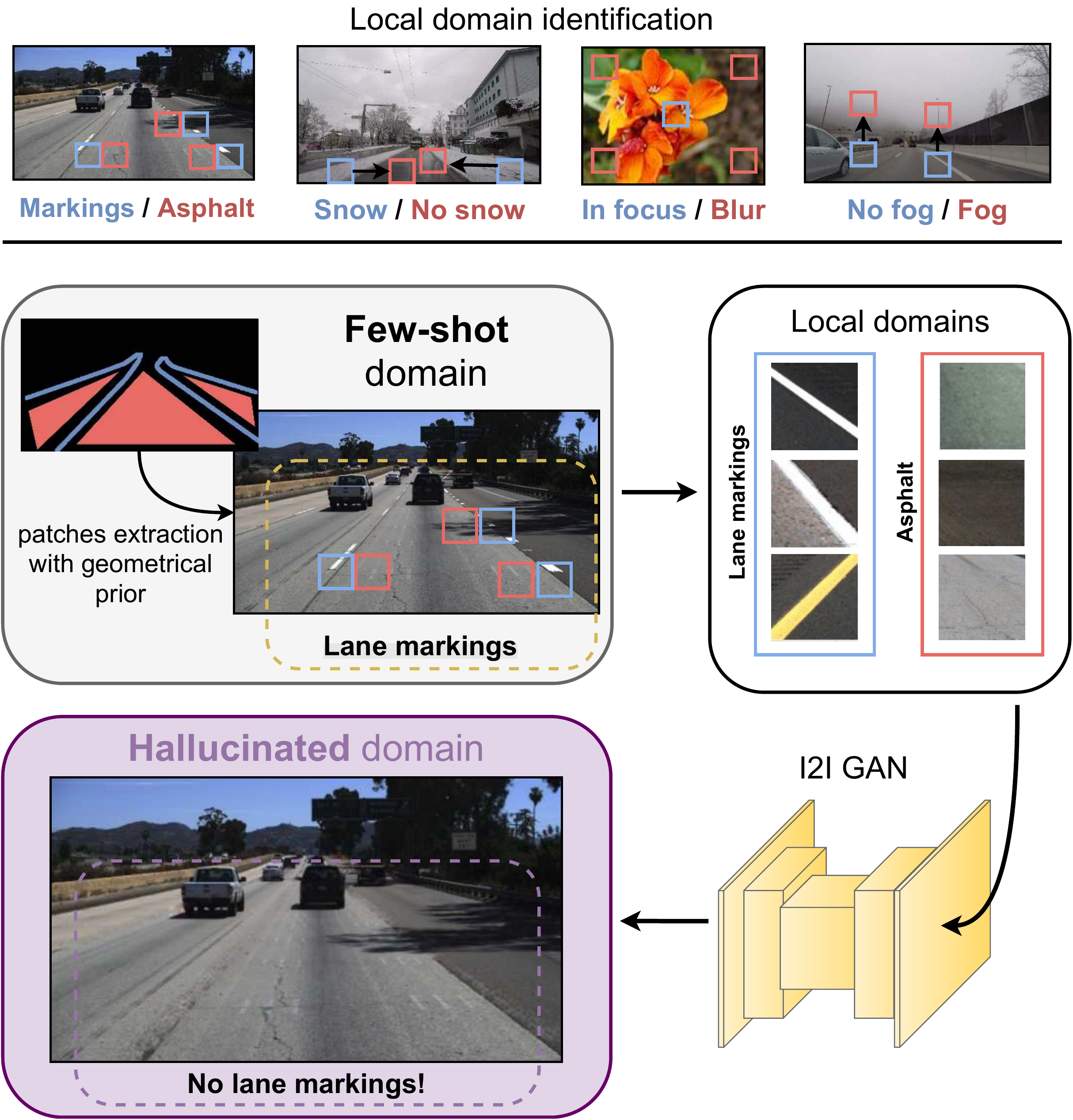}
	\caption{\textbf{Overview.} Our method is able to generate images of unseen domain, leveraging geometrical-guidance to extract patches of \textit{local domains}, i.e. spatially defined sub-domains, on source images. 
		Here, we generate image without any lane markings training only on an extremely small amount of images with well-defined lane markings.
	}
	\label{fig:teaser}
\end{figure}

We propose a method exploiting human knowledge about source and target, to identify domain-specific local characteristics which we call \textit{local domains} (Fig.~\ref{fig:teaser}, top). 
The latter are used as guidance to perform patches translations on \textit{source only}, thus hallucinating a new unseen domain. %
An example in Fig.~\ref{fig:teaser} bottom shows we leverage {local domains} knowledge about `lane markings' and `asphalt' to hallucinate a new domain without lane markings. 
Experimental evidence in this paper shows that our new domain acts as a bridge leading to a performance boost on target.
An interesting characteristic of our method is that it trains few shots on source only, leveraging only minimal human knowledge about the target.\\

\noindent{}In short, the main contributions of this paper are:%
\begin{itemize}
    \item we introduce and define \textit{local domains} as being domain-specific spatial characteristics (Sec.~\ref{sec:meth-localdomains}),
    \item to the best of our knowledge, we propose the first geometrical-guided patch-based i2i, leveraging our \textit{local domains} priors (Sec.~\ref{sec:meth-geom_priors}) and enabling continuous geometrical translation (Sec.~\ref{sec:meth-interp}),
    \item we experiment on three different tasks in a few-shot setting, showing that our translations lead to better performance on all target proxy tasks (Sec.~\ref{sec:exp}).
\end{itemize}
\section{Related works}
\paragraph{Image-to-image translation}
Image-to-image translation has been introduced as an application of conditional GANs in~\cite{isola2017image}, extended by~\cite{liu2017unsupervised,zhu2017toward} for multi-modality or better performances. In more recent approaches, the obvious limitation of requiring paired images for training has been removed~\cite{huang2018multimodal,zhu2017unpaired,lee2019drit++,yi2017dualgan}. Recently, there has been an emergence of attention-based~\cite{mejjati2018unsupervised,ma2018gan,tang2019attention,kim2019u,Lin_2021_WACV} or memory-based~\cite{jeong2021memory} methods, which further guarantee more realism or increased network capabilities. Some methods guarantee multi-domain translations~\cite{choi2018stargan,choi2020stargan}. The first approach efficiently exploiting a patch-wise decomposition of images has been CUT~\cite{park2020contrastive}, which exploits patches from different domains to impose contrastive learning constraints. All these methods use different data as source and target and are unable to identify inter-domain transformation by default. 

\paragraph{Image translation with less supervision}
A recent field of study focus on reducing the number of images necessary for training i2i networks. For instance, in BalaGAN~\cite{patashnik2021balagan} they exploit domain clustering in order to make the training robust to classes with few examples. Other strategies use self-supervision~\cite{wang2020semi} or latent space interpolations~\cite{cao2021remix} in order to avoid the discriminator overfitting and train on extremely small datasets. Differently, FUNIT~\cite{liu2019few} and COCO-FUNIT~\cite{saito2020coco} generalize to few-shot domains at inference stage. Some other works try to work with less supervision at the domain level, on a mixed target domain~\cite{pizzati2021comogan} or without even source and target domain distinctions~\cite{baek2020rethinking,lee2021contrastive}. It is worth noticing that some methods are trained on single images, as SinGAN~\cite{shaham2019singan}, employable for image editing tasks. Finally, ZstGAN~\cite{lin2021zstgan} exploits textual inputs for zero-shot image translation.

\paragraph{Prior-guided image translation}
Several priors could be exploited to increase image translation effectiveness, with several degrees of supervision as bounding boxes~\cite{shen2019towards,bhattacharjee2020dunit}, semantic maps~\cite{li2018semantic,ramirez2018exploiting,tang2020multi,cherian2019sem,zhu2020semantically,zhu2020sean,lin2020multimodal,ma2018exemplar,park2019semantic} or instance labels~\cite{mo2018instagan,xu2021instance}. Another line of works exploits physical models as priors for translation enhancement~\cite{halder2019physics,tremblay2020rain}, disentanglement~\cite{pizzati2021guided}, or guidance~\cite{pizzati2021comogan}. Importantly, scene geometry could be used as a prior, with learned correspondencies~\cite{wu2019transgaga} or by exploiting additional modalities~\cite{arar2020unsupervised}. Some use text for image editing purposes~\cite{liu2020describe}. Bruls et al.~\cite{bruls2019generating} exploit full semantic maps for road randomization, to generalize across challenging lane detection scenarios. However, they are limited to annotated road layouts and constrained by expensive complete segmentation maps.
\begin{figure*}
	\centering
	\includegraphics[width=\linewidth]{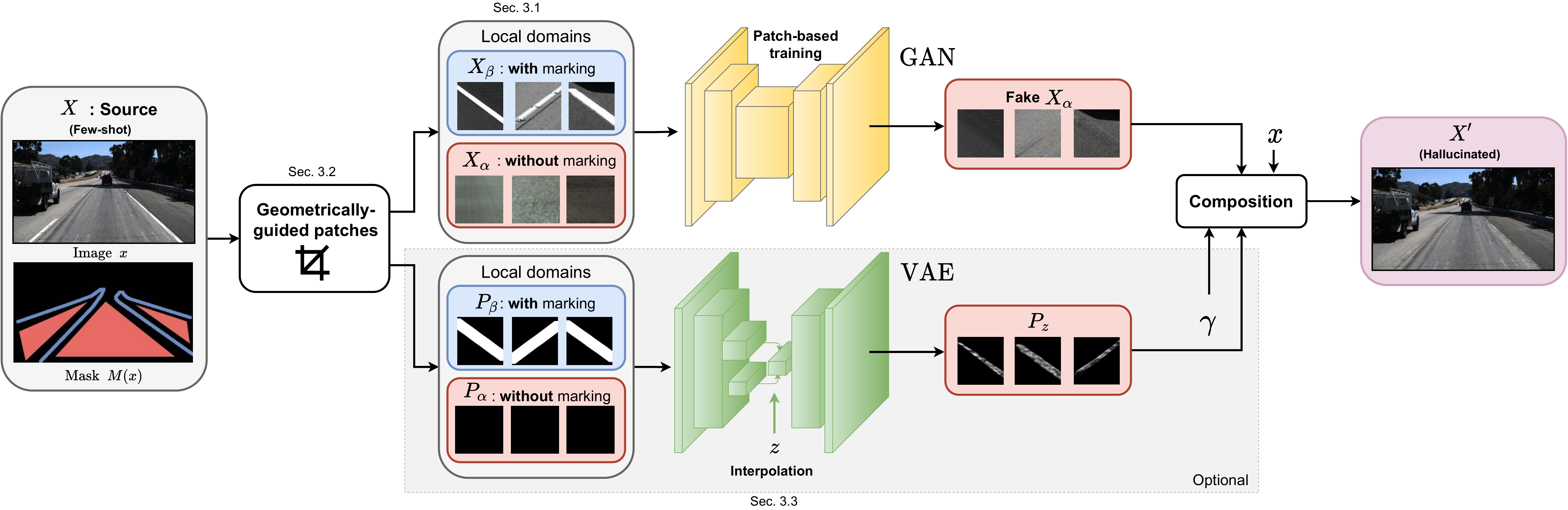}
	\caption{\textbf{Architecture pipeline.} Our method exploits knowledge about \textit{local domains} (Sec.~\ref{sec:meth-localdomains}) and relies on geometrical-prior to extract samples of local domains in \textit{source only} (Sec.~\ref{sec:meth-geom_priors}) that train a patch-based GAN.
	Here, source is having ``lane markings'' and ``asphalt'' local domains ($X_\alpha$ and $X_\beta$, respectively) while target have only ``asphalt'' ($X_\beta$), learning $X_\alpha\mapsto{}X_\beta$ further reduces the gap with target.
	An optional local domain interpolation strategy (Sec.~\ref{sec:meth-interp}) is added for generating geometrically continuous translation between local domains (here, simulating lane degradation).}
	\label{fig:pipeline}
\end{figure*}

\begin{figure}
	\centering
		\setlength{\tabcolsep}{0.003\linewidth}
		\footnotesize
		\begin{tabular}{c c c}
		    \includegraphics[width=0.32\linewidth, valign=m]{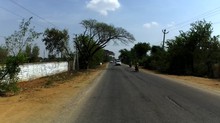}
			& \includegraphics[width=0.32\linewidth, valign=m]{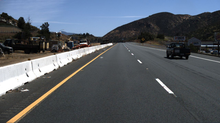}
			& \includegraphics[width=0.32\linewidth, valign=m]{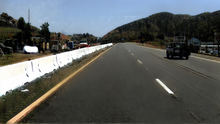}\vspace{0.3em}\\
			IDD~\cite{varma2018idd} & TuSimple~\cite{tusimple} & TuSimple$\mapsto$IDD\\
			
	\end{tabular}
	\caption{\textbf{Translation with CycleGAN~\cite{zhu2017unpaired}.} Sample output shows that i2i is prone to transfer global features (here, sky color) but neglects evident local features for humans as the street structure (note that IDD has no lane markings).} \label{fig:example-tusimple2IDD}
\end{figure}

\section{Method}

We address the problem of image to image translation accounting for \textit{source} and \textit{target} domains having predominant local transformations. As such, leveraging \textit{only} \textit{source} data, our proposal hallucinates a new \textit{unseen} intermediate domain which can be used to ease transfer learning towards \textit{target}. An overview of our pipeline is in Fig.~\ref{fig:pipeline}.

In the following, we introduce our definition of local domains (Sec.~\ref{sec:meth-localdomains}) and propose a geometrical-guided patch-based strategy to learn translation between the latter (Sec.~\ref{sec:meth-geom_priors}). For some local domains, we also show that a continuous geometrical translation can be learned from the interpolation of a mask (Sec.~\ref{sec:meth-interp}). 
Finally, we describe our training strategy showing few shot capabilities (Sec.~\ref{sec:meth-training}).

\subsection{Local domains} \label{sec:meth-localdomains}

Image-to-image (i2i) networks learn a mapping function $G:X\mapsto Y$ from a source domain $X$ to a target domain $Y$, such that the distribution $P_{G(X)}$ approximates $P_{Y}$. The goal is to transfer the features of domain $Y$ to samples from $X$ while preserving their content. 
This works well for transformation globally affecting the scene (eg. summer to winter) but struggles to capture the mappings of local changes due to the under-constrained settings of the system. 
A simple failure example, shown in Fig.~\ref{fig:example-tusimple2IDD}, is the translation from outdoor images having lane markings, to images having no (or degraded) lane markings. 
As it seeks global changes, the i2i is likely to transfer unintended characteristics while missing the subtle -- but consistent -- local changes (here, the lane markings).

To overcome this, we introduce \textit{local domains} which are sub-domains \textit{spatially defined} -- for example, lane markings, asphalt, etc.
Formally, we define domain $X$ as the composition of local domains, denoted $\{X_{\alpha},...,X_{\omega}\}$, and the remaining sub-domains written $X_\text{O}$. Considering only two local domains of interest, it writes:
\begin{equation}
	X = \{X_\text{O}, X_{\alpha}, X_{\beta}\}\,.
\end{equation}
Because we consider only source and target domains sharing at least one local domain, say $_{\alpha}$, we write $Y$ as:
\begin{equation}
	Y = \{Y_\text{O}, Y_{\alpha}\}\,,
\end{equation}
so that the Kullback-Leibler divergence $\text{KL}(X_{\alpha}, Y_{\alpha})$ is close to 0.
Instead of learning the direct mapping of $X\mapsto Y$, we propose to learn local domain mappings, such as $X_{\beta}\mapsto{}X_{\alpha}$. 
If such mapping is applied systematically on all samples from $X$, we get a new domain $X'$ without $_\beta$, so:
\begin{equation}
	X' = \{X_\text{O}, X_{\alpha}\}\,,
\end{equation}
where domain $X'$ is unseen and thus hallucinated. 
Considering that $X'$ and $Y$ share the same local domains, they are subsequently closer: $\text{KL}(Y, X') < \text{KL}(Y,X)$. 

Our intuition is that when training target data is hard to get, our hallucinated domain $X'$ can ease transfer learning. 
Notably here, our method only requires a priori knowledge of the shared local domains in source and target. 

\subsection{Geometrically-guided patches} \label{sec:meth-geom_priors}
Learning the mapping between local domains requires extracting local domain samples. 
To do so we leverage patches corresponding to either local domains in the source dataset only.
We rely here on a simple geometrical guidance from a mask $M(.)$ to extract random patches centered around a given local domain.

Considering $x$ an image in source domain $X$, we extract $\mathcal{X}_{\alpha}$ the unordered set of patches of fixed dimension, so that:\\
\begin{equation}
\label{eq:patches_xalpha}
	\mathcal{X}_{\alpha} = \{\{x_{p_0}, x_{p_1}, \dots, x_{p_m} |\, p \in M_\alpha(x)\}\, |\, \forall x \in X\}\,,
\end{equation}
\noindent{}having $m$ the number of patches per image, and $M_\alpha(x)=\llbracket{}M(x)=\alpha\rrbracket$ with $\llbracket.\rrbracket$ the Iverson brackets. Literally, $M(x)$ is our geometrical prior -- a 2D mask of same size than $x$ -- 
encoding the position of local domains. Subsequently, $M_{\alpha}(x)$ is filled with ones where local domain $X_\alpha$ is and zeros elsewhere. 
Similarly to Eq.~\ref{eq:patches_xalpha}, we extract the set $X_{\beta}$ from $M_{\beta}(x)$ and $X$.

In practice, the geometrical prior $M(x)$ is often simply derivable from the image labels. For example, the position of lane marking and asphalt can both be extracted from image labels. 
In some cases, the position of local domains is constant dataset-wise and we use a fixed geometrical prior, so $M(x) = M$. This is for example the case for portraits datasets, where faces are likely to be centered and background located along the image edges. 

Having collected the two sets of patches $\mathcal{X}_{\alpha}$ and $\mathcal{X}_{\beta}$, a straightforward patch-based GAN can learn ${X}_{\alpha}\mapsto{}{X}_{\beta}$. In some cases, $\mathcal{X}_{\alpha}$ and $\mathcal{X}_{\beta}$ being of similar nature we demonstrate spatial interpolation is beneficial.

\subsection{Local domains interpolation} \label{sec:meth-interp}

Continuous i2i are extensively studied~\cite{gong2019dlow, wang2019deep, lample2018fader}, but existing methods are not suitable for translation affecting only local regions as in our problem setting~(see~Sec.~\ref{sec:exp}). 
Instead, we learn a non-linear geometrical interpolation of patch masks, leveraging a variational autoencoder (VAE).

Previously we described each patch as encompassing a single local domain but, in reality, patches often mix multiple local domains. This is the case of lane markings patches, shown in Fig.~\ref{fig:pipeline}, that contain asphalt too.
Hence, along with the set of local domains patches 
we extract the sets $\mathcal{P}_{\alpha}$ and $\mathcal{P}_{\beta}$ directly from our geometrical guidance $M(.)$, and seek to continuously interpolate $P_{\alpha}\mapsto{}P_{\beta}$.

In practice, our VAE having encoder $E(.)$ and decoder $D(.)$ is trained in the standard fashion, but at inference it yields the latent representation $h_Z$ corresponding to the linear combination of $E(p_{\alpha})$ and $E(p_{\beta})$, having $p_{\alpha} \in \mathcal{P}_{\alpha}$ and $p_{\beta} \in \mathcal{P}_{\beta}$, respectively\footnote{In our formalism, we include the VAE reparametrization in $E(.)$}.
Formally:
\begin{equation}
    \begin{split}
        h_Z = E(p_{\alpha}) \, z + E(p_{\beta}) \, (1 - z), \\
        p_{z} = D(h_Z),
    \end{split}
\end{equation}
where $z \in [0, 1]$ encodes the progress along $P_{\alpha}\mapsto{}P_{\beta}$. 
The final interpolated patch $x_z$ is the composite between $x_{\alpha}$ and $x_{\beta}$ patches, following the VAE output. It writes:
\begin{equation}\label{eq:blending}
	\begin{split}
    x_z = x_{\alpha} \, m + x_{\beta} \, (1 - m)\,, \\
    \text{with}\;m=\gamma{}\,p_{z}\,,
	\end{split}
\end{equation}
$\gamma\in[0,1]$ being an arbitrary controlled blending parameter adding a degree of freedom to our model.
Furthermore, notice that the stochastic VAE behavior further increases variability, beneficial for proxy tasks.

\subsection{Training} \label{sec:meth-training}

We train our pipeline, the patch-based GAN and the optional VAE, leveraging only images from the source domain and geometrical priors about local domains. The patch-based GAN is trained on ${X}_{\alpha}\mapsto{}{X}_{\beta}$ (Sec.~\ref{sec:meth-geom_priors}) minimizing the LSGAN~\cite{mao2017squares} adversarial loss:
\begin{equation}
    \begin{split}
        y_f &= G(x), \\
        \mathcal{L}_G(y_f) &= \mathbb{E}_{x\sim P_X(x)} \left[ (D(y_f)-1)^2\right],\\
        \mathcal{L}_D(y_f,y) &= \mathbb{E}_{x\sim P_X(x)} \left[ (D(y_f))^2 \right] + \\
        &+ \mathbb{E}_{y\sim P_Y(y)} \left[ (D(y)-1)^2\right],
    \end{split}
\end{equation}
along with task-specific losses. 
If used, the VAE interpolation (Sec.~\ref{sec:meth-interp}) is trained with standard ELBO strategy~\cite{blei2017variational}, minimizing reconstruction loss along with a regularizer:
\begin{equation}
\begin{split}
    \mathcal{L}_{VAE} &= -\mathbb{E}_{q_{\phi} (z|x)}\log p_{\theta}(x|z) +\\
    &+ D_{KL}(q_{\phi} (z|x) ||p(z)).
\end{split}
\end{equation}

At inference time, the full image is fed to the GAN backbone to produce the translated image, while the corresponding full interpolation mask is obtained processing mask patches independently and then stitching them together with a simple algorithm. Of note, our method has important few-shot capabilities. As we train only on source patches a reduced number of image samples is sufficient to get reasonable data diversity, which we further demonstrated in the following section.

\begin{figure*}
	\newcommand{\flowerimg}[1]{\includegraphics[width=7em, valign=m]{figures/qualit_deblurring/#1.png}\llap{\makebox[7em][l]{\raisebox{-2.13em}{\textcolor{white}{\frame{\includegraphics[height=2em, valign=m]{figures/qualit_deblurring/blurmaps/blurmap_#1.png}}}}}}}
	\centering
	\resizebox{\linewidth}{!}{
		\setlength{\tabcolsep}{0.003\linewidth}
		\tiny
		\centering
		\begin{tabular}{c c c c c c c c c c c}

          \multicolumn{11}{c}{{Task with interpolation}}\\
		    \toprule
		    && Original & \multicolumn{8}{c}{Ours}\\			\adjustbox{valign=m}{\multirow{3}{*}[-1.5em]{\rotatebox{90}{\textbf{Lane degradation}}}}
			& \adjustbox{valign=m}{{\rotatebox{90}{}}}
			& \multicolumn{1}{c | }{\includegraphics[width=7em, valign=m]{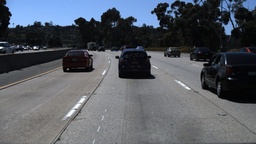}}
			& \includegraphics[width=7em, valign=m]{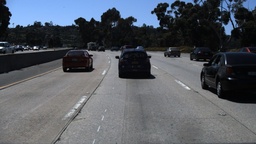}
			& \includegraphics[width=7em, valign=m]{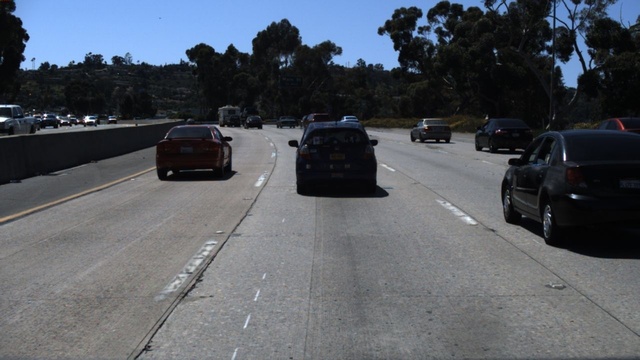}
			& \includegraphics[width=7em, valign=m]{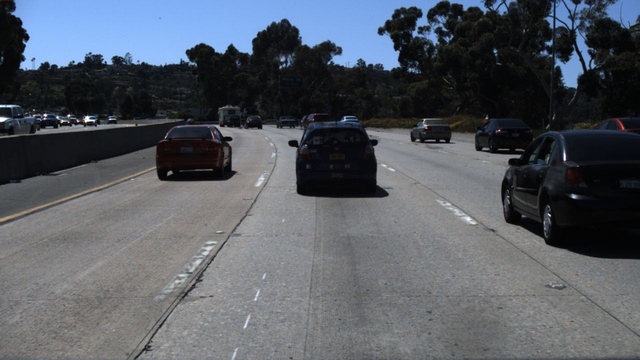}
			& \includegraphics[width=7em, valign=m]{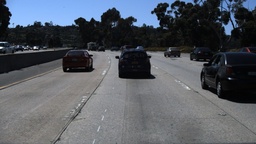}
			& \includegraphics[width=7em, valign=m]{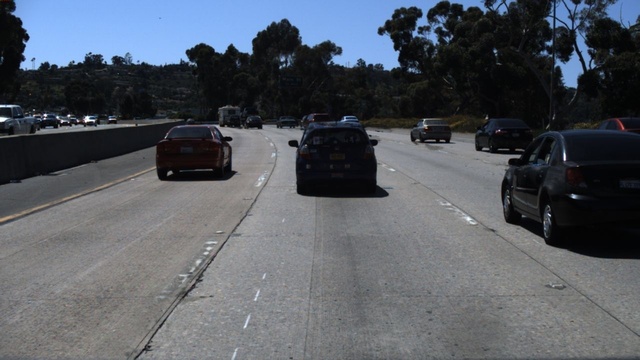}
			& \includegraphics[width=7em, valign=m]{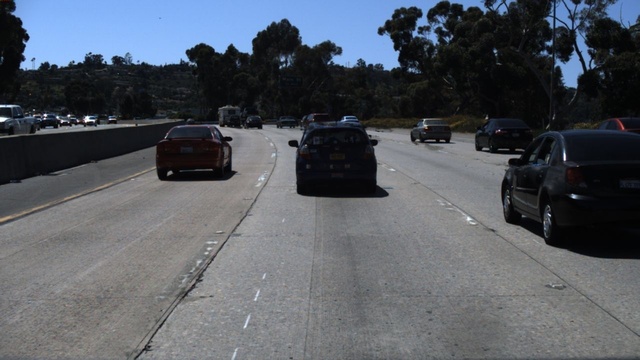}
			& \includegraphics[width=7em, valign=m]{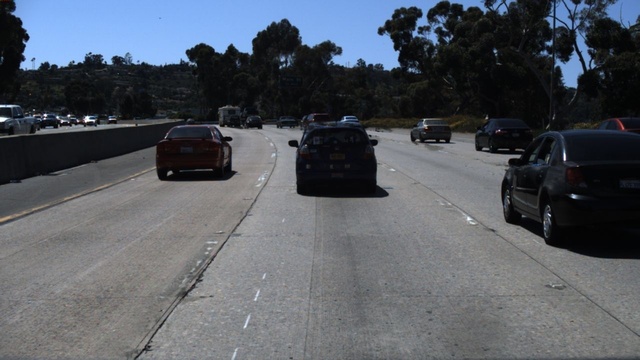}\vspace{0.3em}\\
			& \adjustbox{valign=m}{{\rotatebox{90}{}}}
			& \multicolumn{1}{c | }{\includegraphics[width=7em, valign=m]{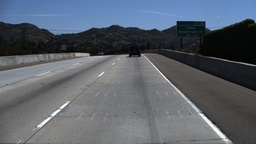}}
			& \includegraphics[width=7em, valign=m]{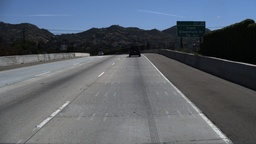}
			& \includegraphics[width=7em, valign=m]{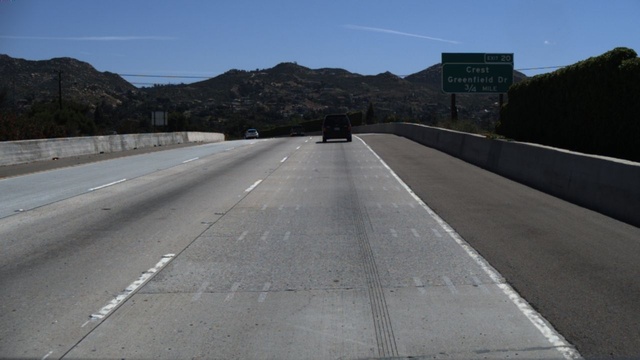}
			& \includegraphics[width=7em, valign=m]{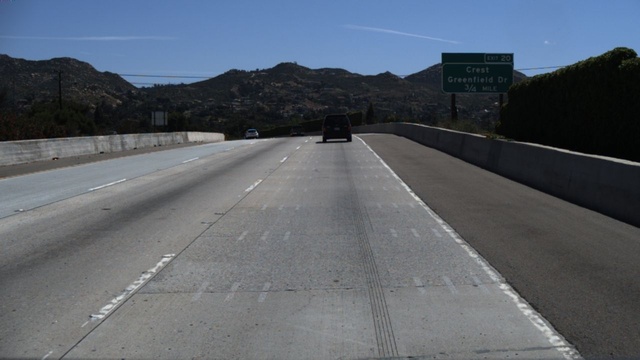}
			& \includegraphics[width=7em, valign=m]{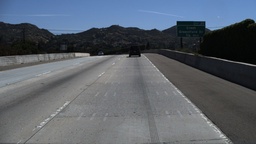}
			& \includegraphics[width=7em, valign=m]{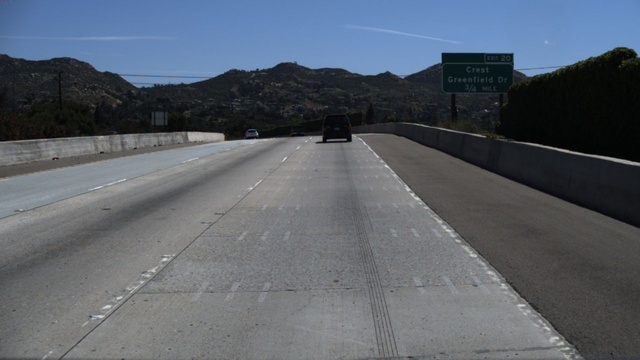}
			& \includegraphics[width=7em, valign=m]{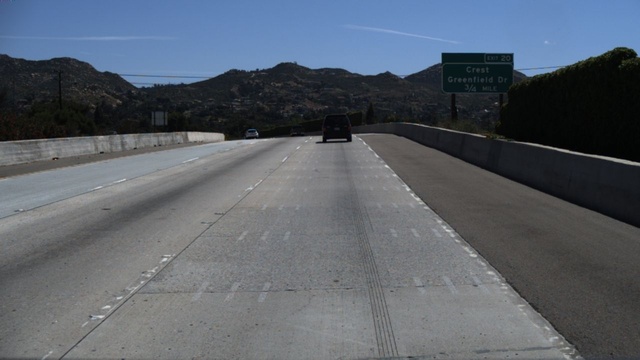}
			& \includegraphics[width=7em, valign=m]{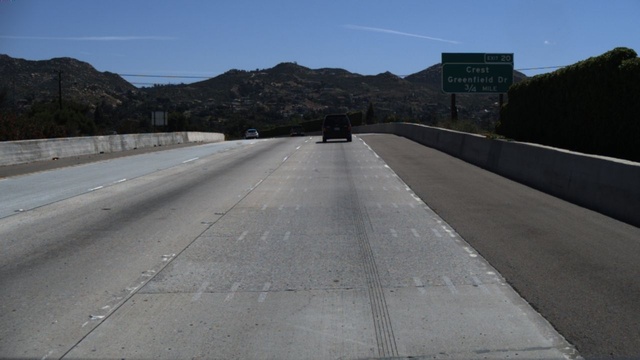}\vspace{0.3em}\\\vspace{-5px}
			& \adjustbox{valign=m}{{\rotatebox{90}{}}}
			& \multicolumn{1}{c | }{\includegraphics[width=7em, valign=m]{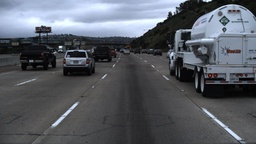}}
			& \includegraphics[width=7em, valign=m]{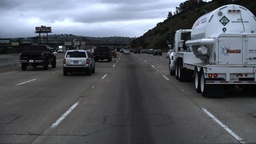}
			& \includegraphics[width=7em, valign=m]{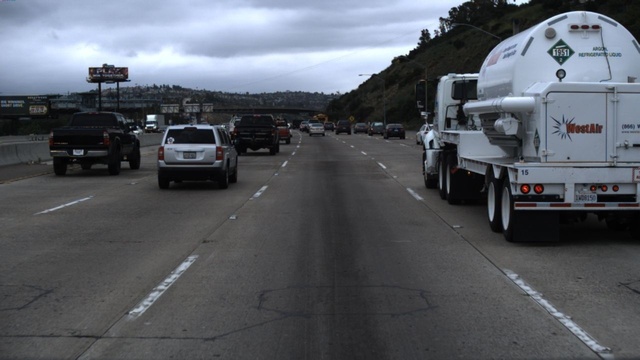}
			& \includegraphics[width=7em, valign=m]{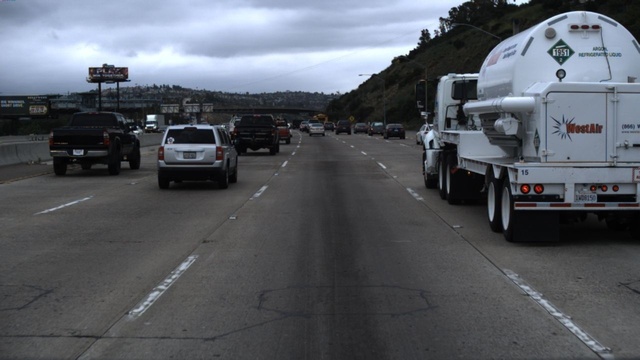}
			& \includegraphics[width=7em, valign=m]{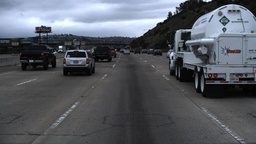}
			& \includegraphics[width=7em, valign=m]{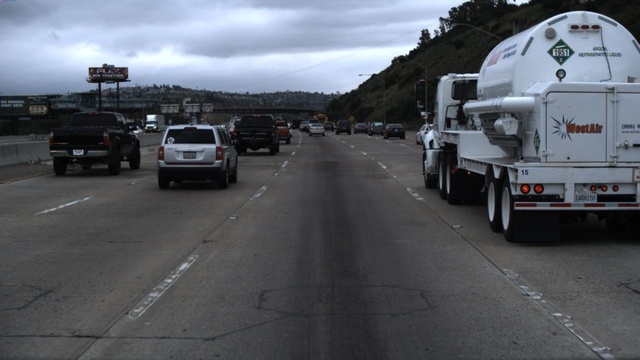}
			& \includegraphics[width=7em, valign=m]{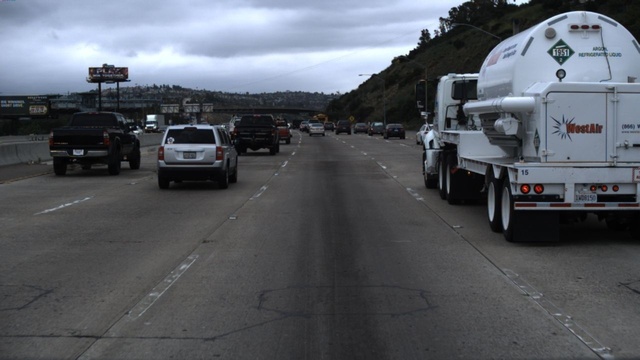}
			& \includegraphics[width=7em, valign=m]{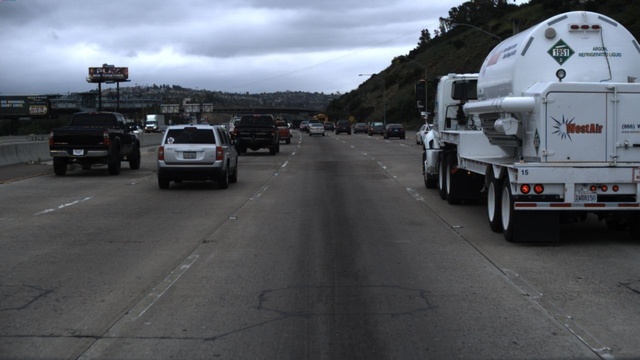}\vspace{1em}\\\vspace{-5px}
		    && & \tikzmark{a}{0.35} &&&&&& \tikzmark{b}{0.95} \\
		    &&&&&\multicolumn{2}{c}{\textcolor{bostonuniversityred}{}}&&&&\\

			\multicolumn{11}{c}{{Task without interpolation}}\\
			\toprule
			
			\adjustbox{valign=m}{\multirow{2}{*}{\rotatebox{90}{\textbf{Snow addition}}}}
			& \adjustbox{valign=m}{{\rotatebox{90}{Original}}}
			& \includegraphics[width=7em, valign=m]{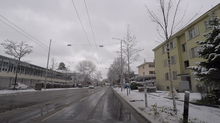}
			& \includegraphics[width=7em, valign=m]{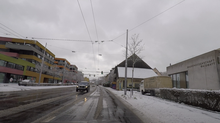}
			& \includegraphics[width=7em, valign=m]{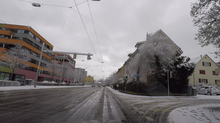}
			& \includegraphics[width=7em, valign=m]{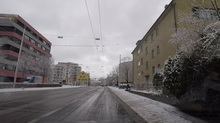}
			& \includegraphics[width=7em, valign=m]{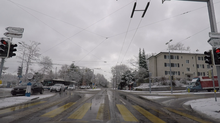}
			& \includegraphics[width=7em, valign=m]{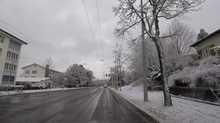}
			& \includegraphics[width=7em, valign=m]{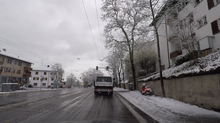}
			& \includegraphics[width=7em, valign=m]{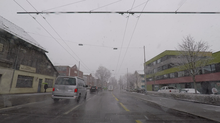}\vspace{0.3em}\\
			& \adjustbox{valign=m}{{\rotatebox{90}{Ours}}}
			& \includegraphics[width=7em, valign=m]{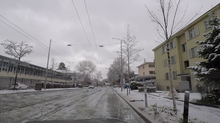}
			& \includegraphics[width=7em, valign=m]{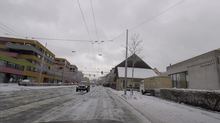}
			& \includegraphics[width=7em, valign=m]{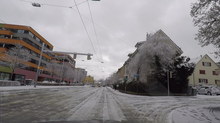}
			& \includegraphics[width=7em, valign=m]{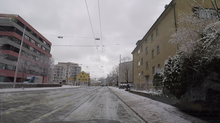}
			& \includegraphics[width=7em, valign=m]{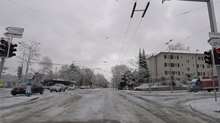}
			& \includegraphics[width=7em, valign=m]{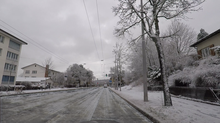}
			& \includegraphics[width=7em, valign=m]{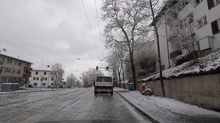}
			& \includegraphics[width=7em, valign=m]{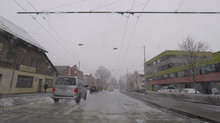}\vspace{0.3em}\\
			\midrule
			
			\adjustbox{valign=m}{\multirow{4}{*}[-1.3em]{\rotatebox{90}{\textbf{Deblurring}}}}
			& \adjustbox{valign=m}{{\rotatebox{90}{Original}}}
			& \flowerimg{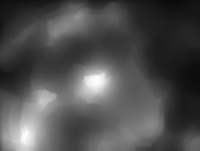}
			& \flowerimg{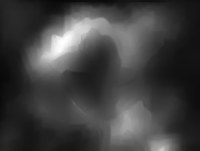}
			& \flowerimg{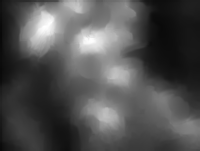}
			& \flowerimg{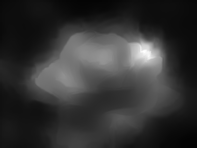}
			& \flowerimg{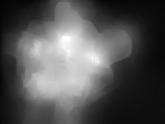}
			& \flowerimg{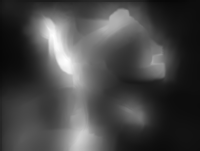}
			& \flowerimg{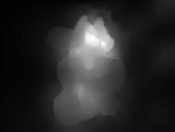}
			& \flowerimg{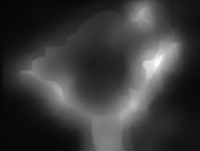}\vspace{0.3em}\\
			
			& \adjustbox{valign=m}{{\rotatebox{90}{Ours}}}
			& \flowerimg{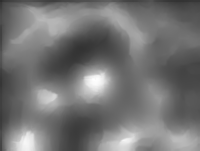}
			& \flowerimg{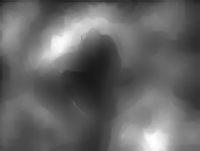}
			& \flowerimg{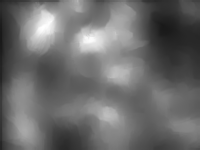}
			& \flowerimg{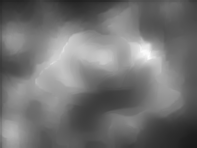}
			& \flowerimg{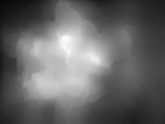}
			& \flowerimg{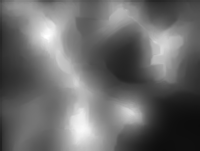}
			& \flowerimg{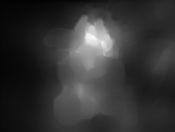}
			& \flowerimg{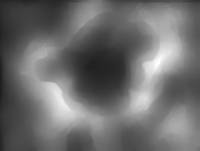}\vspace{0.3em}\\
			\bottomrule
	\end{tabular}\linkwithlabel{a}{b}{$z$}}
	\caption{\textbf{Qualitative results.} For each task we show the original input image and our output with the $X_\beta\mapsto{}X_\alpha$ local domains translation. \textbf{Lane degradation}: sample translations on TuSimple~\cite{tusimple} test set with increasing degradation $z \in [0.35, 0.95]$ from left to right, blending variable $\gamma = 0.75$. \textbf{Snow addition}: augmentation of ACDC~\cite{sakaridis2021acdc} validation set, only road is involved in the transformation. \textbf{Deblurring}: Original flower images~\cite{oxfordflowersdataset} and our deblurred version. Bottom left insets show the in-focus map~\cite{golestaneh2017spatiallyvarying} which are whiter on average (ie. less blurred) for ours.} \label{fig:qualitative-alltasks}
\end{figure*}

\section{Experiments}
\label{sec:exp}
We evaluate our method on 3 different tasks, namely lane markings degradation, snow addition and deblurring, leveraging 5 recent datasets~\cite{tusimple,varma2018idd,sakaridis2021acdc,cordts2016cityscapes,oxfordflowersdataset}, and evaluating our translation both against i2i baselines and on proxy tasks.
In Sec.~\ref{sec:exps-settings} we provide details on our tasks, while Secs.~\ref{sec:evaluation},~\ref{sec:ablation} report extensive qualitative and quantitative evaluation. 

\subsection{Tasks definitions}\label{sec:exps-settings}
We describe our three task below, detailing the learned local domains translation $X_\beta\mapsto{}X_\alpha$.

\paragraph{Lane degradation.}\label{sec:lane-degradation}
Here, we leverage the highway TuSimple~\cite{tusimple} dataset having clear lane markings.
For local domains, we chose \textit{lane marking} ($X_\beta$) and \textit{asphalt} ($X_\alpha$) exploiting geometrical priors from the provided lane labels and assuming near by asphalt. 
We use our interpolation strategy (Sec.~\ref{sec:meth-interp}) accounting for both degradation and blending. 
Importantly, we train only on 15 images (1280x720) to demonstrate few-shot capabilities, with 30 patches per image of size 128x128, 200x200 and 256x256. Backbones are DeepFillv2~\cite{yu2019freeform} as GAN and IntroVAE~\cite{huang2018introvae} for interpolation. The latter is trained with a binarized difference mask from lane inpainting and original image.

We evaluate our translations on the standard 358/2782 val/test sets of TuSimple.
In addition to demonstrating generalization, we evaluate several lane detectors on 110 images from the India Driving Dataset (IDD)~\cite{varma2018idd} -- never seen during training -- having degraded lane markings which we manually annotated.

\paragraph{Snow addition.}\label{sec:snow-addition}
Here, we rely on snowy images from the recent Adverse Driving Conditions Dataset (ACDC)~\cite{sakaridis2021acdc}, which typically have snow only on sidewalk and \textit{not} on road. The task is to add snow on the road. 
Logically, local domains are \textit{road} ($X_\beta$) and snowy \textit{sidewalk} ($X_\alpha$), exploiting semantic labels as priors.
Again, we train only with 15 images with 30 patches (128x128) per image , using CycleGAN~\cite{zhu2017unpaired} with default hyperparameters. No interpolation is used.

We evaluate on the original val/test set of ACDC having 100/500 images.
To increase generalization for the segmentation task in snowy weather, we also augment Cityscapes~\cite{cordts2016cityscapes} with the same trained network.

\paragraph{Deblurring.}\label{sec:deblurring}
We leverage the Oxford 102 Flower dataset~\cite{oxfordflowersdataset} to learn turning shallow Depth of Field (DoF) photos to deep DoF, therefore seeking to deblur the image.
As blur is not labeled, we rely on a simple \textit{dataset-wise} geometrical prior that image center is always in-focus and image corners are always out-of-focus. Local domains are \textit{out-of-focus} ($X_\beta$) and \textit{in-focus} ($X_\alpha$).
Since we use only 8 patches per image (4 in-focus, 4 out-of-focus, 128x128), we train our CycleGAN~\cite{zhu2017unpaired} with 400 images, %
adding a task-specific objective function defined as the composition of a color consistency loss and an in-focus loss:
\begin{equation}
    \mathcal{L}_{deblur} = D_{KL}(H[x]||H[G(x)]) + \frac{1}{\sigma^2_{LoG(G(x))}},
\end{equation}
with $H[.]$ the image histogram and $\sigma^2_{LoG(.)}$ the Laplacian of Gaussian variance.
A color jitter augmentation is applied to ensure discriminator invariance to color.
We do not use interpolation. At inference, we exclude foreground since it impacts translation quality due to the identity loss in CycleGAN~\cite{zhu2017unpaired}.

\subsection{Evaluation}\label{sec:evaluation}
\subsubsection{Translation quality}\label{sec:transl-quality}

Qualitative results are visible in Fig.~\ref{fig:qualitative-alltasks} and show our method outputs realistic translations for all tasks. 
In details, we are able to modify lanes (first three rows) on TuSimple  with different degrees of degradation (from left to right).
On snow addition, images show plausible snow on ACDC roads (middle two rows), preserving shadows.
\begin{table}
	\centering
	\setlength{\tabcolsep}{0.01\linewidth}
	
	\footnotesize
	\begin{tabular}{cc}
		\toprule
		\textbf{Images} & \textbf{In-focus avg}$\uparrow$ \\\midrule
		Original & {1.28} \\ %
		Ours \textit{(deblurred)} & \textbf{1.53} \\
		\bottomrule
		
	\end{tabular}%
	\caption{\textbf{Deblurring performance.} Average of the in-focus maps~\cite{golestaneh2017spatiallyvarying} on the Oxford Flowers~\cite{oxfordflowersdataset} test set show our method efficiently deblur the input images despite a trivial dataset-wise geometrical prior.}
	\label{tab:quantit-deblurring}
\end{table}
Finally, on our deblurring task (bottom two rows) the flowers background appear in-focus, exhibiting sharper edges. To better size the benefit on this last task, flower images have as inset the in-focus map computed with~\cite{golestaneh2017spatiallyvarying} (ie. white means in-focus).

Of note, evaluating GAN metrics on target images would be biased since we use only source images -- unlike existing i2i --. They are reported in the supp for the sake of completeness.
To provide a quantitative quality evaluation, Tab.~\ref{tab:quantit-deblurring} reports the in-focus average proving our translations are significantly more in-focus (+0.24) than original images.

\begin{figure}
	\centering
	\subcaptionbox{Qualitative\label{fig:qualitative-lanedegr}}{
		\resizebox{1.0\linewidth}{!}{
		\setlength{\tabcolsep}{0.003\linewidth}
		
		\begin{tabular}{c c c H c H c}
		    &&&&\Large Baselines&&\\
		    \toprule 
		    && \tikzmark{c}{\Large 0} &&&& \tikzmark{d}{\Large 1}\\
			&\adjustbox{valign=m}{\rotatebox{90}{\Large DLOW~\cite{gong2019dlow}}}
			& \includegraphics[width=7em, valign=m]{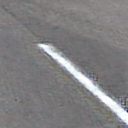}
			& \includegraphics[width=7em, valign=m]{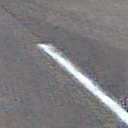}
			& \includegraphics[width=7em, valign=m]{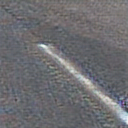}
			& \includegraphics[width=7em, valign=m]{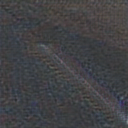}
			& \includegraphics[width=7em, valign=m]{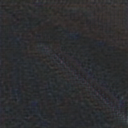}\vspace{0.3em}\\
			&\adjustbox{valign=m}{{\rotatebox{90}{\Large DLOW+}}}
			& \includegraphics[width=7em, valign=m]{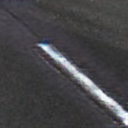}
			& \includegraphics[width=7em, valign=m]{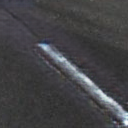}
			& \includegraphics[width=7em, valign=m]{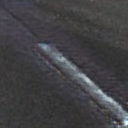}
			& \includegraphics[width=7em, valign=m]{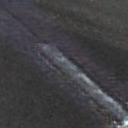}
			& \includegraphics[width=7em, valign=m]{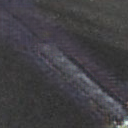}\vspace{0.3em}\\
			\bottomrule
		\end{tabular}\linkwithlabel{c}{d}{\huge $z$}%
		\hspace{20px}
		\begin{tabular}{c c c H c H c}
            &&&&\Large {Ours}&&\\
		    \toprule
		    && \tikzmark{e}{\Large 0} &&&& \tikzmark{f}{\Large 1}\\

			\tikzmark{h}{\Large 1}&
			& \includegraphics[width=7em, valign=m]{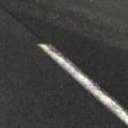}
			& \includegraphics[width=7em, valign=m]{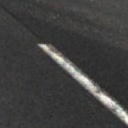}
			& \includegraphics[width=7em, valign=m]{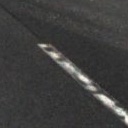}
			& \includegraphics[width=7em, valign=m]{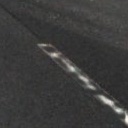}
			& \includegraphics[width=7em, valign=m]{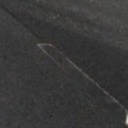}\vspace{0.3em}\\
			\tikzmark{g}{\Large 0.35} &
			& \includegraphics[width=7em, valign=m]{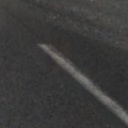}
			& \includegraphics[width=7em, valign=m]{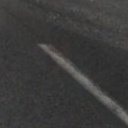}
			& \includegraphics[width=7em, valign=m]{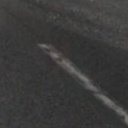}
			& \includegraphics[width=7em, valign=m]{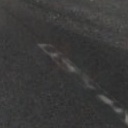}
			& \includegraphics[width=7em, valign=m]{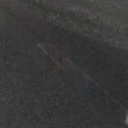}\vspace{0.3em}\\

			\bottomrule
		\end{tabular}\linkwithlabel{e}{f}{\huge $z$}\linkyellowwithlabel{g}{h}{\huge $\gamma$}}
	}\\\vspace{1em}%
	\subcaptionbox{GAN metrics\label{fig:gan-metrics-lanedegrad}}{
		\scriptsize
		\setlength{\tabcolsep}{0.015\linewidth}
		\begin{tabular}{ccc}
			\toprule
			\textbf{Network} & \textbf{FID$\downarrow$} & \textbf{LPIPS$\downarrow$} \\\hline
			DLOW~\cite{gong2019dlow} & {211.7} & {0.4942}\\ %
			DLOW+ & {155.6} & {0.4206}\\ %
			Ours w/o blending & {154.7} & {0.3434}\\ %
			Ours & \textbf{135.4} & \textbf{0.3254}\\ %
			\bottomrule
		\end{tabular}
	}
	\caption{\textbf{Lane translations.} (\subref{fig:qualitative-lanedegr}) Qualitative comparison of lane degradation on patches with baselines. Our method is the only one to output a realistic degradation. (\subref{fig:gan-metrics-lanedegrad}) GAN metrics on the lane degradation task prove the benefit of our method.} 
\end{figure}
\begin{figure}
	\newcommand{\laneimg}[3]{\setlength{\fboxrule}{0pt}%
	\includegraphics[width=7em, valign=m]{#1}\llap{\raisebox{-2.4em}{\makebox[7em][l]{\textcolor{bostonuniversityred}{\fbox{\small \textbf{#2}}}}}}\llap{\raisebox{-2.4em}{\makebox[7em][r]{\textcolor{airforceblue}{\fbox{\small \textbf{#3}}}}}}}
	
	\centering
	\resizebox{\linewidth}{!}{
		\setlength{\tabcolsep}{0.003\linewidth}
		\tiny
		\begin{tabular}{c c c c c c c}\toprule
			\adjustbox{valign=m}{{\rotatebox{90}{2nd best}}}
			& \laneimg{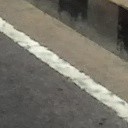}{0.3}{1.0}
			& \laneimg{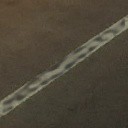}{0.5}{1.0}
			& \laneimg{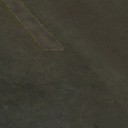}{0.5}{0.2}
			& \laneimg{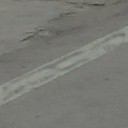}{0.5}{0.8}
			& \laneimg{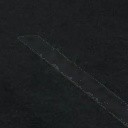}{0.4}{0.2}
			& \laneimg{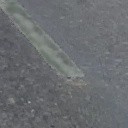}{0.6}{0.4}
			\vspace{0.3em}\\
			\adjustbox{valign=m}{{\rotatebox{90}{Best matching}}}
			& \laneimg{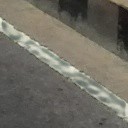}{0.6}{1.0}
			& \laneimg{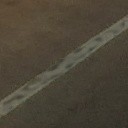}{0.4}{0.6}
			& \laneimg{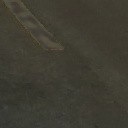}{0.6}{0.4}
			& \laneimg{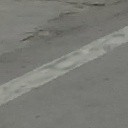}{0.4}{1.0}
			& \laneimg{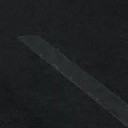}{0.3}{0.4}
			& \laneimg{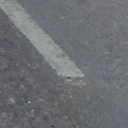}{0.3}{0.6}
			\vspace{0.3em}\\
			\midrule
			\adjustbox{valign=m}{{\rotatebox{90}{Degraded reference}}}
			& \includegraphics[width=7em, valign=m]{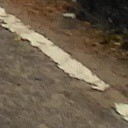}
			& \includegraphics[width=7em, valign=m]{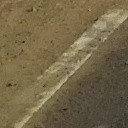}
			& \includegraphics[width=7em, valign=m]{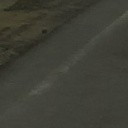}
			& \includegraphics[width=7em, valign=m]{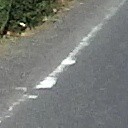}
			& \includegraphics[width=7em, valign=m]{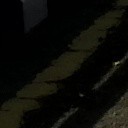}
			& \includegraphics[width=7em, valign=m]{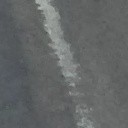}\vspace{0.3em}\\
			\adjustbox{valign=m}{{\rotatebox{90}{Clear}}}
			& \includegraphics[width=7em, valign=m]{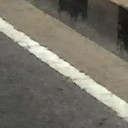}
			& \includegraphics[width=7em, valign=m]{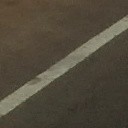}
			& \includegraphics[width=7em, valign=m]{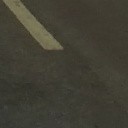}
			& \includegraphics[width=7em, valign=m]{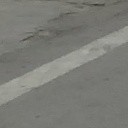}
			& \includegraphics[width=7em, valign=m]{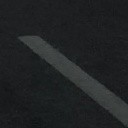}
			& \includegraphics[width=7em, valign=m]{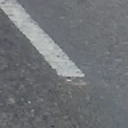}\vspace{0.3em}\\

			\bottomrule
	\end{tabular}}
	\caption{\textbf{Evaluation of lane degradation on patches taken from IDD dataset~\cite{varma2018idd}}. We associate clear patches (bottom row) to degraded ones (third row) by minimizing LPIPS. Applying our method to clean images variating the \textcolor{bostonuniversityred}{$z$} and \textcolor{airforceblue}{$\gamma$} parameters (shown in the images), we subsequently lower the LPIPS. We display the best and second best translation in terms of LPIPS. The similarities of our results with the degraded patches prove the efficacy of our LPIPS-based evaluation.} \label{fig:lpips-rankings}
\end{figure}

\subsubsection{Interpolation quality}\label{sec:interp-quality}
For the lane degradation task, we compare our interpolations against the continuous i2i DLOW~\cite{gong2019dlow} baseline, trained on the same data. 
As it suffers from evident color artifacts, we introduce DLOW+: a custom version using lane mask as additional channel input, masked reconstruction loss, and masked input-output blending.
For DLOW/DLOW+, we regulate the walk on the discovered manifold of each network with a \textit{domainness} variable $z$ \mbox{-- which} amounts to our lane degradation.

With respect to baselines, it is visible in Fig.~\ref{fig:qualitative-lanedegr} that our degraded lane translations are more realistic for all $z$ since DLOW and DLOW+ discover simpler transformations, just regulating color homogeneously. 

For quantification, we compare translations against real degraded lane markings from IDD and report FID and LPIPS in Fig.~\ref{fig:gan-metrics-lanedegrad}. 
In details, we select 35/62 clear/degraded lane patches from IDD test set, and couple those with minimum LPIPS~\cite{zhang2018unreasonable} distance. 
Intuitively, we pair similar clear and degraded lane markings together. Pairs are shown in the two bottom rows of Fig.~\ref{fig:lpips-rankings}.
We then degrade each clear image with ours / DLOW / DLOW+, generating several degraded versions, and use the best degrading version in terms of LPIPS w.r.t. its clear match to compute GAN metrics. 
Fig.~\ref{fig:gan-metrics-lanedegrad} shows we outperform baseline on both metrics significantly (roughly, -20 FID, and -0.1 LPIPS), demonstrating the realism of our lane degradation. 
Since baselines are not using any explicit blending as us (see $\gamma$ in Eq.~\ref{eq:blending}), we also evaluate ``ours w/o blending'' using $m=p_z$ in Eq.~\ref{eq:blending}, which still outperforms baselines.

\subsubsection{Proxy tasks}\label{sec:proxy-tasks}
Here, we study the applicability of our pipeline to increase the robustness of existing lane detection and semantic segmentation networks.

\begin{table}
	\centering
	\footnotesize
	\setlength{\tabcolsep}{0.01\linewidth}
	\resizebox{\linewidth}{!}{%
		
		\begin{tabular}{cccccccccc}
			\toprule
			\multirow{2}{*}{\textbf{Detector}} & \multirow{2}{*}{\textbf{Translation}} & \multicolumn{3}{c}{\textbf{{TuSimple~\cite{tusimple}}}} & \multicolumn{3}{c}{\textbf{IDD~\cite{varma2018idd}}} \\
			& & Acc. $\uparrow$ & FP $\downarrow$ & FN $\downarrow$ & Acc. $\uparrow$ & FP $\downarrow$ & FN $\downarrow$ \\
			\midrule
			\multirow{2}{*}{SCNN~\cite{pan2017spatial}} & none (source) &  \textbf{0.946}& \textbf{0.052}& \textbf{0.069}& {0.617}& {0.538} & {0.741}\\
			& Ours& {0.945}& {0.058}& {0.072}& \textbf{0.730}& \textbf{0.453} & \textbf{0.577}\\
			\midrule
			\multirow{2}{*}{RESA~\cite{zheng2021resa}} & none (source) & \textbf{0.952}& \textbf{0.056}& \textbf{0.065}& {0.639}& {0.720} & {0.800}\\
			& Ours& {0.951}& {0.059}& {0.068}& \textbf{0.671}& \textbf{0.686} & \textbf{0.761}\\
			\bottomrule
		\end{tabular}
	}%
	\caption{\textbf{Lane detection on TuSimple and IDD.} Performance of lane detectors when trained on TuSimple source (\textit{none}) or our degraded translations (\textit{ours}). The latter significantly outperforms baseline, while retaining equivalent performances on TuSimple images.}\label{tab:quantit-lanedetector}
\end{table}

\begin{figure}
	\centering
	\resizebox{\linewidth}{!}{
		\setlength{\tabcolsep}{0.003\linewidth}
		\tiny
		\begin{tabular}{c c c c c}
			\adjustbox{valign=m}{{\rotatebox{90}{Input}}}
			& \includegraphics[width=7em, valign=m]{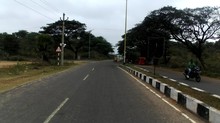}
			& \includegraphics[width=7em, valign=m]{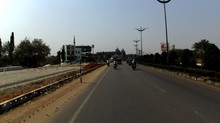}
			& \includegraphics[width=7em, valign=m]{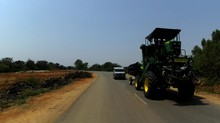}
			& \includegraphics[width=7em, valign=m]{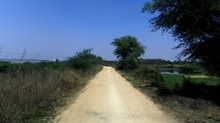}\vspace{0.3em}\\
			\adjustbox{valign=m}{{\rotatebox{90}{none}}}
			& \includegraphics[width=7em, valign=m]{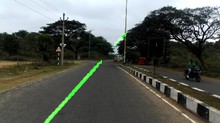}
			& \includegraphics[width=7em, valign=m]{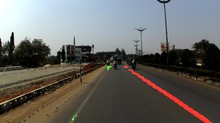}
			& \includegraphics[width=7em, valign=m]{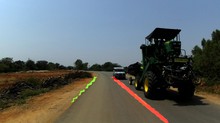}
			& \includegraphics[width=7em, valign=m]{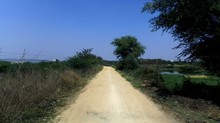}\vspace{0.3em}\\
			\adjustbox{valign=m}{{\rotatebox{90}{Ours}}}
			& \includegraphics[width=7em, valign=m]{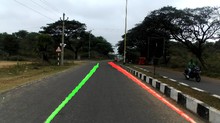}
			& \includegraphics[width=7em, valign=m]{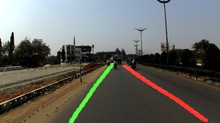}
			& \includegraphics[width=7em, valign=m]{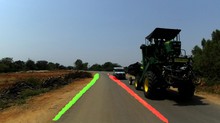}
			& \includegraphics[width=7em, valign=m]{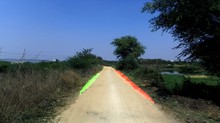}\\
			
	\end{tabular}}
	\caption{\textbf{SCNN~\cite{pan2017spatial} lane detection on IDD~\cite{varma2018idd}.} Training on generated images with degraded lanes makes existing lane detectors (e.g. SCNN~\cite{pan2017spatial}) resistant to scenes with damaged (first three columns) or no (last column) lane markings.} \label{fig:qualit-lanedetIDD}
\end{figure}

\paragraph{Lane detection.}
We aim here to make lane detectors robust to \textit{unseen} degraded lane markings. 
To do so, we train two state-of-the-art detectors (SCNN~\cite{pan2017spatial} and RESA~\cite{zheng2021resa}) on both TuSimple original images and our translated version (mixing with 5\% probability and randomizing $z$ and~$\gamma$). 
The models are tested on both the TuSimple test set and our 110 labeled IDD images, the latter having severely degraded lane markings.

From the quantitative results in Tab.~\ref{tab:quantit-lanedetector}, we observe that with our source degraded translations both detectors severely outperform the baselines using clear source on the challenging IDD, while maintaining on-par performances on TuSimple with clear markings. 
In particular, for SCNN we improve by $+11.3\%$ the accuracy, $-8.5\%$ the false positives and $-16.4\%$ the false negatives. Sample qualitative results are in Fig.~\ref{fig:qualit-lanedetIDD} and showcase the robustness of our method on degraded or even absent street lines. 
We conjecture that our degraded translations forced the network to rely on stronger contextual information.

\begin{figure}
	\centering
	\resizebox{\linewidth}{!}{
		\setlength{\tabcolsep}{0.003\linewidth}
		\tiny
		\centering
		\begin{tabular}{c c c c c}
		    \adjustbox{valign=m}{{\rotatebox{90}{Original}}}
			& \includegraphics[width=7em, valign=m]{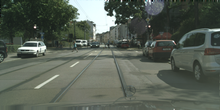}
			& \includegraphics[width=7em, valign=m]{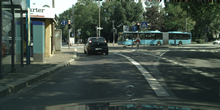}
			& \includegraphics[width=7em, valign=m]{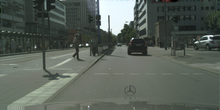}
			& \includegraphics[width=7em, valign=m]{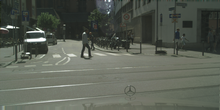}\vspace{0.3em}\\
			\adjustbox{valign=m}{{\rotatebox{90}{Ours}}}
			& \includegraphics[width=7em, valign=m]{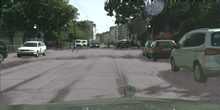}
			& \includegraphics[width=7em, valign=m]{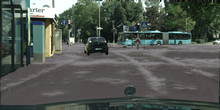}
			& \includegraphics[width=7em, valign=m]{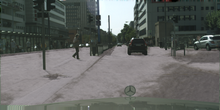}
			& \includegraphics[width=7em, valign=m]{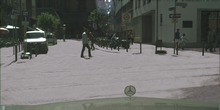}\\

		\end{tabular}}
		\caption{\textbf{Cityscapes images with snow added}. We add snow on roads and sidewalks of the Cityscapes training set to train semantic segmentation networks robust to snow. Cityscapes exhibits a domain shift with respect to ACDC, but our method is still able to generate acceptable snow.}\label{fig:cityscapes-snowy}
\end{figure}
\begin{table}
	\centering
	\footnotesize
	\setlength{\tabcolsep}{0.015\linewidth}
	\begin{tabular}{ccccc}
		\toprule
		\textbf{Model} & \textbf{Translations} & \textbf{road IoU $\uparrow$}  & \textbf{sidewalk IoU $\uparrow$} & \textbf{mIoU $\uparrow$} \\
		\midrule
		\multirow{2}{*}{DeepLabv3+~\cite{chen2018encoderdecoder}} & none (source) & {74.95} & {39.52} & {45.31} \\
		& Ours & \textbf{80.56} & \textbf{49.52} & \textbf{47.64}\\
		\midrule
		\multirow{2}{*}{PSANet~\cite{Zhao2018PSANetPS}} & none (source) & \textbf{74.29} & {30.71} & {42.97}\\
		& Ours & {74.01} & \textbf{36.28} & \textbf{43.85}\\
		\midrule
		\multirow{2}{*}{OCRNet~\cite{YuanCW20}} & none (source) & {82.30} & {45.60} & {54.54}\\
		& Ours & \textbf{82.78} & \textbf{54.69} & \textbf{55.48}\\
		\bottomrule
	\end{tabular}
	\label{tab:snowseg}
	\caption{\textbf{Semantic segmentation on ACDC~\cite{sakaridis2021acdc} snow.} We train multiple segmentation network on Cityscapes~\cite{cord2011towards} with added snow with our method and test on ACDC~\cite{sakaridis2021acdc} snow validation, consistently improving generalization capabilities.} \label{tab:quantit-semanticseg}
\end{table}

\begin{figure}
	\centering
	\resizebox{\linewidth}{!}{
		\setlength{\tabcolsep}{0.003\linewidth}
		\tiny
		\begin{tabular}{c c c c c}
			\adjustbox{valign=m}{{\rotatebox{90}{Input}}}
			& \includegraphics[width=7em, valign=m]{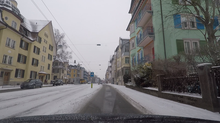}
			& \includegraphics[width=7em, valign=m]{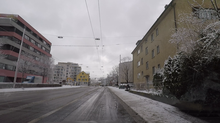}
			& \includegraphics[width=7em, valign=m]{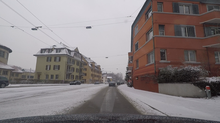}
			& \includegraphics[width=7em, valign=m]{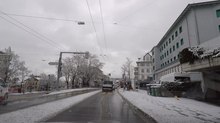}\vspace{0.3em}\\
			\adjustbox{valign=m}{{\rotatebox{90}{GT}}}
			& \includegraphics[width=7em, valign=m]{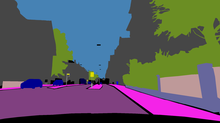}
			& \includegraphics[width=7em, valign=m]{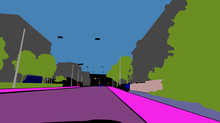}
			& \includegraphics[width=7em, valign=m]{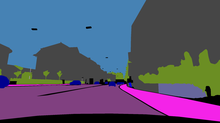}
			& \includegraphics[width=7em, valign=m]{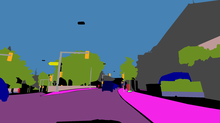}\vspace{0.3em}\\
			\adjustbox{valign=m}{{\rotatebox{90}{none}}}
			& \includegraphics[width=7em, valign=m]{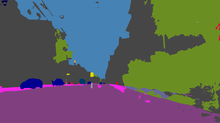}
			& \includegraphics[width=7em, valign=m]{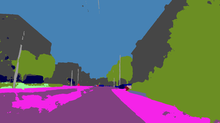}
			& \includegraphics[width=7em, valign=m]{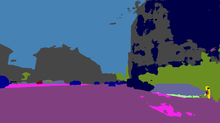}
			& \includegraphics[width=7em, valign=m]{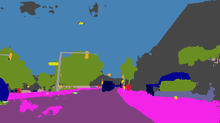}\vspace{0.3em}\\
			\adjustbox{valign=m}{{\rotatebox{90}{{Ours}}}}
			& \includegraphics[width=7em, valign=m]{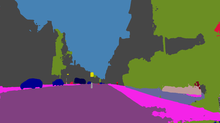}
			& \includegraphics[width=7em, valign=m]{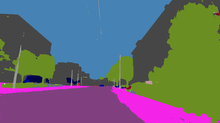}
			& \includegraphics[width=7em, valign=m]{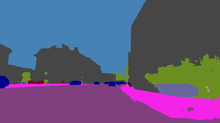}
			& \includegraphics[width=7em, valign=m]{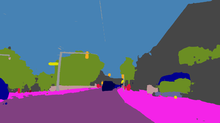}\\
			
	\end{tabular}}
	\caption{\textbf{DeepLabv3+~\cite{chen2018encoderdecoder} on ACDC~\cite{sakaridis2021acdc} snow.} Training with our generated images brings improvements in segmentation of snowy scenes in ACDC~\cite{sakaridis2021acdc}, especially in the road and sidewalk classes.} \label{fig:qualit-semanticseg}
\end{figure}

\paragraph{Semantic segmentation.}
Here, we seek to improve segmentation in snowy driving conditions. 
We train three state-of-the-art semantic segmentation models, namely DeepLabv3+~\cite{chen2018encoderdecoder}, PSANet~\cite{Zhao2018PSANetPS} and OCRNet~\cite{YuanCW20}, with either clear Cityscapes images and snowy Cityscapes images translated with our method. Translation examples are available in Fig.~\ref{fig:cityscapes-snowy}, where we add snow on sidewalks and road by using Cityscapes semantic maps. Visual results remain acceptable and snow is added uniformly on both semantic classes, even if inference on Cityscapes brings a consistent domain shift with respect to training patches on ACDC.
In detail, for the latter we augment images with $10\%$~(DeepLabv3+, PSANet) or $5\%$~(OCRNet) probability. 
The models are evaluated on the ACDC snow validation set.

Tab.~\ref{tab:quantit-semanticseg} shows the benefit of our augmented images (\textit{Ours}) 
to consistently improve the performance on road or sidewalk (our two local domains) and mean IoU for all networks. 
From Fig.~\ref{fig:qualit-semanticseg} it is visible that the model trained with our augmentation strategy is able to better detect roads and footpaths in difficult weather conditions with respect to the baseline, which is not capable of properly discriminating between them if they are covered with snow.

\subsection{Ablation study}\label{sec:ablation}
\begin{table}
	\centering
	\setlength{\tabcolsep}{0.01\linewidth}
	
	\footnotesize
	\begin{tabular}{cccc}
		\toprule
		\textbf{TuSimple samples (\%)} & \textbf{Patches/img} & \textbf{LPIPS}$\downarrow$   & \textbf{FID}$\downarrow$\\\midrule
		15 (0.4\%) & 1 & {0.3296} & {148.22} \\
		15 (0.4\%) & 5 & {0.3295} & {135.53} \\
		15 (0.4\%) & 30 & {0.3254} & {131.73} \\
		15 (0.4\%) & 60 & {0.3246} & {127.94} \\
		15 (0.4\%) & 150 & \textbf{0.3222} & \textbf{126.94} \\
		\midrule
		50 (1.4\%) & 30 & {0.3236} & {129.42} \\
		150 (4.1\%) & 30 & {0.3221} & \textbf{124.79} \\
		500 (14\%) & 30 & {0.3234} & {128.56} \\
		3626 (100\%) & 30 & \textbf{0.3218} & {125.56} \\
		\bottomrule
		
	\end{tabular}%
	\caption{\textbf{Data ablation on TuSimple.} The use of data on the lane degradation task ($\text{TuSimple}\mapsto\text{IDD}$) is ablated by varying the number of images and patches per images in the training set, and evaluating GAN similarity metrics (see Sec.~\ref{sec:interp-quality}) on IDD.}\label{tab:ablation-imgs_patches}
\end{table}

\paragraph{Training images and patches.} As mentioned our method requires very little images to train. 
Here, we study the effect of number of images and patches per image on the lane degradation task. To measure its impact, we use LPIPS~\cite{zhang2018unreasonable} and FID~\cite{heusel2018gans} following Sec.~\ref{sec:interp-quality}.

Results in Tab.~\ref{tab:ablation-imgs_patches} show, as expected, better translation with the increase of both the number of images and the number of patches extracted per each image. However, we also denote the few-shot capability of our method, and the minimal benefit of using a large number of images.

\begin{table}
	\centering
	\setlength{\tabcolsep}{0.01\linewidth}
	
	\footnotesize
	\begin{tabular}{cccc}
		\toprule
		\textbf{Augmented images (\%)} & \textbf{road IoU}$\uparrow$ & \textbf{sidewalk IoU}$\uparrow$   & \textbf{mIoU}$\uparrow$\\\midrule
		100 & {36.43} & {35.26} & {39.83} \\
		66 & {60.43} & {43.81} & {46.84} \\
		50 & {68.78} & {45.79} & \textbf{50.31} \\
		20 & {75.85} & {44.60} & {47.36} \\
		10 & \textbf{80.56} & \textbf{49.52} & {47.64}\\\midrule
		0 (none) & {74.95} & {39.52} & {45.31}\\
		\bottomrule
		
	\end{tabular}%
	\caption{\textbf{Augmentation percentage ablation on Cityscapes.} The effectiveness of our snow addition translation is ablated by varying the probability of Cityscapes augmented images shown to DeepLabv3+~\cite{chen2018encoderdecoder} during training. Segmentation evaluation is reported on ACDC~\cite{sakaridis2021acdc} validation set.} \label{tab:ablation-augment_seg}
\end{table}

\paragraph{Augmentation percentage.} We study also how the percentage of augmented images shown to DeepLabv3+ network at training impacts performances on semantic segmentation in snowy conditions.

As indicated in Tab.~\ref{tab:ablation-augment_seg}, we achieve best performances with an augmentation probability of $50\%$ ($+5\%$ mIoU w.r.t. no augmentation), still we use for evaluation in Sec.~\ref{sec:evaluation} the model obtained with $10\%$ for its higher accuracy on road and sidewalk -- crucial for autonomous navigation tasks.

\section{Conclusion}
In this work, we proposed a patch-based image-to-image translation model which relies on a GAN backbone trained on patches and an optional VAE to interpolate non-linearly between domains. Along with the definition of \textit{local domains}, we introduced a dataset-based geometrical guidance strategy to ease the patches extraction process. Our few-shot method outperformed the literature on all tested metrics on several tasks (lane degradation, snow addition, deblurring), and its usability has been demonstrated on proxy tasks. In particular, our translation pipeline led to higher performances on lane detection in scenes with degraded or absent markings and on semantic segmentation in snowy conditions. 

{\small
\bibliographystyle{ieee_fullname}
\bibliography{bibliography}
}







%
%


\onecolumn

\begin{appendices}
\section{Introduction}
In this document, we provide additional results of our method. In Sec.~\ref{sec:gan-metrics}, we propose an additional comparison with baselines for the snow addition task, described in the main paper, Sec.~4.1. Also, in Sec.~\ref{sec:additional-qualit}, we display additional qualitative results for image-to-image translation (Sec.~\ref{sec:i2i}) and the proposed proxy tasks (Sec.~\ref{sec:proxy-tasks}).

\section{Experiments}\label{sec:experiments}
\subsection{GAN metrics}\label{sec:gan-metrics}

In the main paper, we intentionally omit GAN metrics as they have important biases for two reasons we explain now. First, our method leverages only local domains translation while standard i2i applies a global transformation. Second, while we leverage high-level domain priors about local domains, we do not use any target images unlike standard i2i.

For completeness we still report GAN metrics against CycleGAN~\cite{zhu2017unpaired} and CycleGAN-15, which are trained on {$\text{ACDC}_\text{clear}\mapsto\text{ACDC}_\text{snow}$} using respectively 400/400 or 400/15 source/target images.

Comparatively, we \textit{only} use the same 15 cherry picked $\text{ACDC}_\text{snow}$ images.
The quantitative evaluation is obtained by performing roads and sidewalks translation of 100 $\text{ACDC}_\text{clear}$ images relying on segmentation masks from OCRNet~\cite{YuanCW20} pretrained on Cityscapes~\cite{cordts2016cityscapes}. 
Since ACDC provides images (weakly) paired, we compute the pair-wise average LPIPS metric between each fake translation and its paired real snow image. We also evaluate FID between the fake and real snow datasets.
It is important to note that we do not seek to outperform the baselines since they have access to $\text{ACDC}_\text{clear}$ images while our method does not. Results in Tab.~\ref{tab:gan-metrics-snow} however show we perform reasonably good given the additional domain gap, even on par with baselines on LPIPS metric.

In addition, in Fig.~\ref{fig:qualit-snowaddcmp} our translations are shown to be significantly more homogeneous than baselines.

\begin{table}[b]
        \centering
        \footnotesize
    	\setlength{\tabcolsep}{0.015\linewidth}
            \begin{tabular}{c|cc|cc}
            \toprule
            \textbf{Network} & \multicolumn{2}{c|}{\textbf{Training samples}} & \textbf{FID$\downarrow$} & \textbf{LPIPS$\downarrow$} \\
             & clear & snow & \\\hline
            CycleGAN~\cite{zhu2017unpaired} & 400 & 400 & \textbf{110.30} & \textbf{0.6225}\\
            CycleGAN-15 & 400 & 15 & {111.29} & {0.6271}\\
            Ours & 0 & 15 & {123.14} & {0.6283}\\
            \bottomrule
            \end{tabular}
        \caption{\textbf{Snow translation similarity on $\text{ACDC}_\text{snow}$.} GAN metrics on the snow addition task confirm the validity of our model. Without using any target images, our model yields acceptable results on FID, attaining even almost on-par performances with baselines on LPIPS.}
        \label{tab:gan-metrics-snow}
\end{table}

\begin{figure*}[hb]
	\centering
	\resizebox{\linewidth}{!}{
		\setlength{\tabcolsep}{0.003\linewidth}

		\scriptsize
		\begin{tabular}{c c c c c c c}

			\adjustbox{valign=m}{{\rotatebox{90}{\tiny Input}}}
			& \includegraphics[width=7em, valign=m]{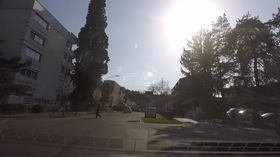}
			& \includegraphics[width=7em, valign=m]{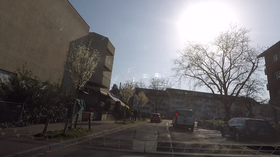}
			& \includegraphics[width=7em, valign=m]{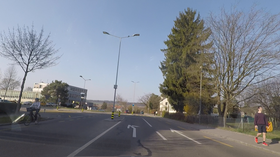}
			& \includegraphics[width=7em, valign=m]{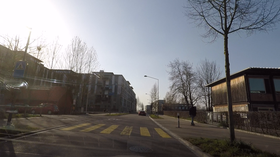}
			& \includegraphics[width=7em, valign=m]{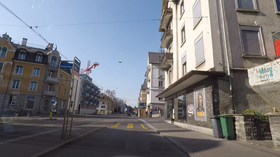}
			& \includegraphics[width=7em, valign=m]{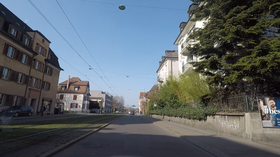}\vspace{0.5em}\\
			\adjustbox{valign=m}{{\rotatebox{90}{\tiny CycleGAN}}}
			& \includegraphics[width=7em, valign=m]{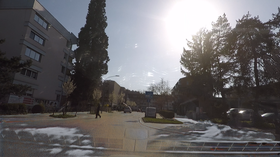}
			& \includegraphics[width=7em, valign=m]{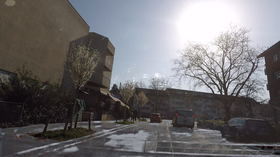}
			& \includegraphics[width=7em, valign=m]{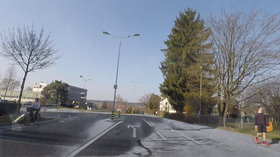}
			& \includegraphics[width=7em, valign=m]{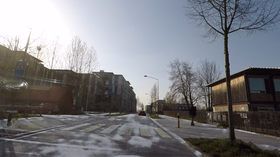}
			& \includegraphics[width=7em, valign=m]{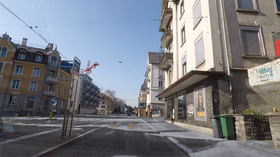}
			& \includegraphics[width=7em, valign=m]{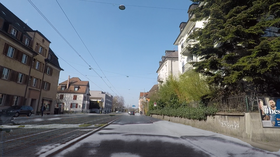}\vspace{0.5em}\\
			\adjustbox{valign=m}{{\rotatebox{90}{\tiny CycleGAN-15}}}
			& \includegraphics[width=7em, valign=m]{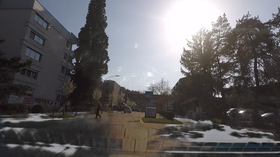}
			& \includegraphics[width=7em, valign=m]{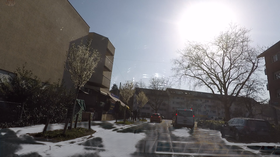}
			& \includegraphics[width=7em, valign=m]{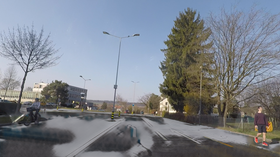}
			& \includegraphics[width=7em, valign=m]{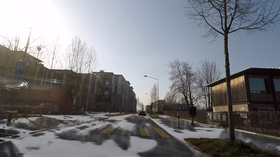}
			& \includegraphics[width=7em, valign=m]{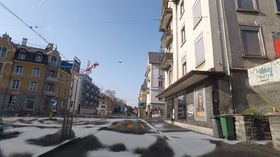}
			& \includegraphics[width=7em, valign=m]{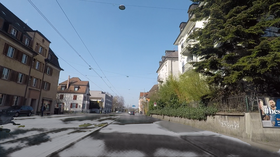}\vspace{0.5em}\\
			\adjustbox{valign=m}{{\rotatebox{90}{\tiny Ours}}}
			& \includegraphics[width=7em, valign=m]{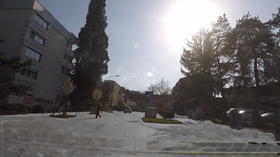}
			& \includegraphics[width=7em, valign=m]{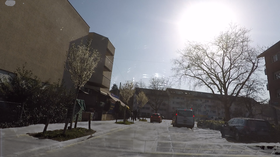}
			& \includegraphics[width=7em, valign=m]{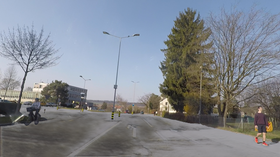}
			& \includegraphics[width=7em, valign=m]{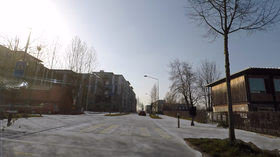}
			& \includegraphics[width=7em, valign=m]{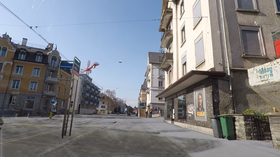}
			& \includegraphics[width=7em, valign=m]{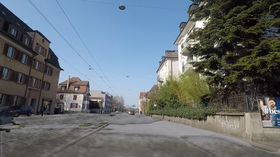}\\

	\end{tabular}}
	\caption{\textbf{Qualitative snow addition comparison on ACDC$_{\text{clear}}$.} Our method (last row) yields acceptable performances compared with different strategies (rows 2-3). Please note that we are the only one to \textit{not} use ACDC$_{\text{clear}}$ images for training. } \label{fig:qualit-snowaddcmp}
\end{figure*}

\subsection{Additional qualitative results}\label{sec:additional-qualit}

\subsubsection{Image-to-image translation}\label{sec:i2i}

\begin{figure*}
	\newcommand{\flowerimg}[1]{\includegraphics[width=7em, valign=m]{figures/supplementary_materials/deblurring/#1.png}\llap{\makebox[7em][l]{\raisebox{-1.9em}{\textcolor{white}{\frame{\includegraphics[height=2em, valign=m]{figures/supplementary_materials/deblurring/blurmaps/blurmap_#1.png}}}}}}}
	\centering
	\resizebox{\linewidth}{!}{
		\setlength{\tabcolsep}{0.003\linewidth}

		\tiny
		\centering
		\begin{tabular}{c c c c c c c c c c c}

          \multicolumn{11}{c}{{Task with interpolation}}\\
		    \toprule
		    && Original & \multicolumn{8}{c}{Ours}\\			\adjustbox{valign=m}{\multirow{8}{*}[-11.5em]{\rotatebox{90}{\textbf{Lane degradation}}}}
			& \adjustbox{valign=m}{{\rotatebox{90}{}}}
			& \multicolumn{1}{c | }{\includegraphics[width=7em, valign=m]{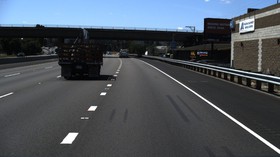}}
			& \includegraphics[width=7em, valign=m]{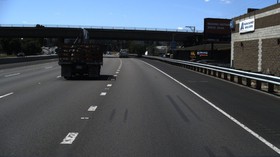}
			& \includegraphics[width=7em, valign=m]{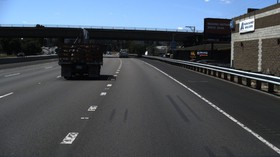}
			& \includegraphics[width=7em, valign=m]{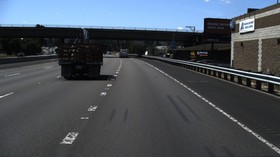}
			& \includegraphics[width=7em, valign=m]{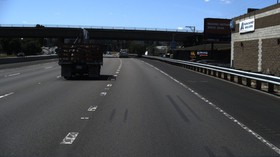}
			& \includegraphics[width=7em, valign=m]{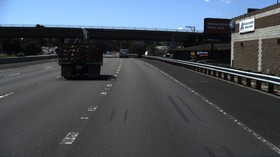}
			& \includegraphics[width=7em, valign=m]{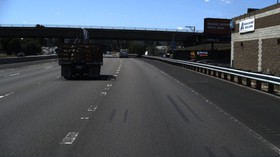}
			& \includegraphics[width=7em, valign=m]{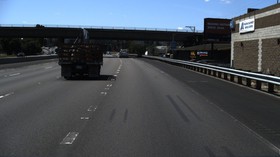}\vspace{0.3em}\\
	        & \adjustbox{valign=m}{{\rotatebox{90}{}}}
	        & \multicolumn{1}{c | }{\includegraphics[width=7em, valign=m]{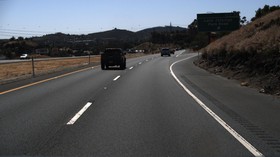}}
			& \includegraphics[width=7em, valign=m]{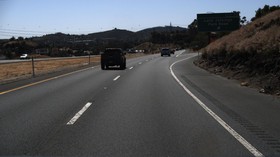}
			& \includegraphics[width=7em, valign=m]{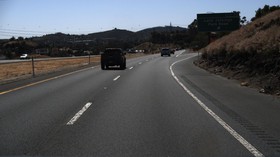}
			& \includegraphics[width=7em, valign=m]{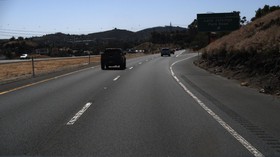}
			& \includegraphics[width=7em, valign=m]{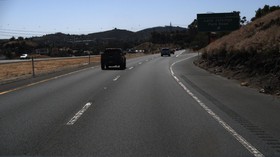}
			& \includegraphics[width=7em, valign=m]{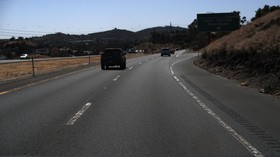}
			& \includegraphics[width=7em, valign=m]{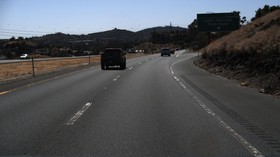}
			& \includegraphics[width=7em, valign=m]{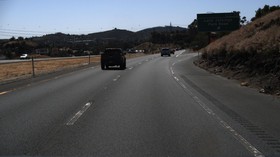}\vspace{0.3em}\\
			& \adjustbox{valign=m}{{\rotatebox{90}{}}}
			& \multicolumn{1}{c | }{\includegraphics[width=7em, valign=m]{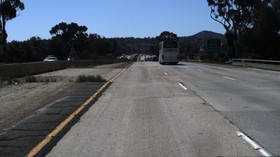}}
			& \includegraphics[width=7em, valign=m]{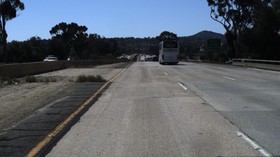}
			& \includegraphics[width=7em, valign=m]{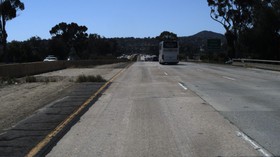}
			& \includegraphics[width=7em, valign=m]{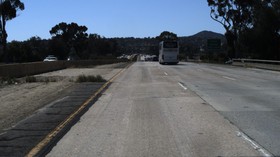}
			& \includegraphics[width=7em, valign=m]{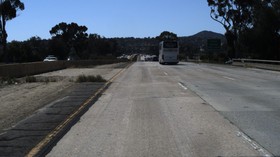}
			& \includegraphics[width=7em, valign=m]{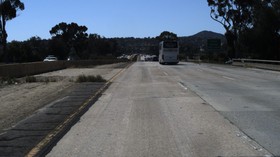}
			& \includegraphics[width=7em, valign=m]{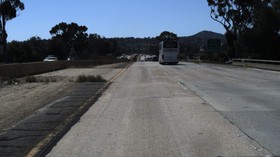}
			& \includegraphics[width=7em, valign=m]{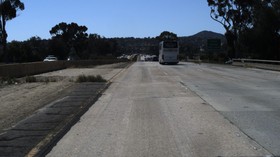}\vspace{0.3em}\\
			& \adjustbox{valign=m}{{\rotatebox{90}{}}}
			& \multicolumn{1}{c | }{\includegraphics[width=7em, valign=m]{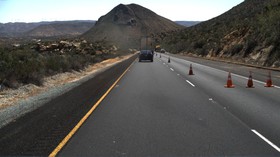}}
			& \includegraphics[width=7em, valign=m]{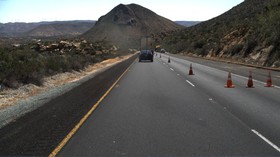}
			& \includegraphics[width=7em, valign=m]{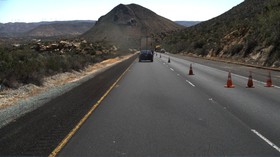}
			& \includegraphics[width=7em, valign=m]{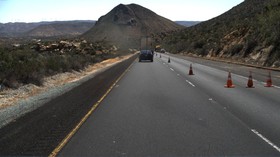}
			& \includegraphics[width=7em, valign=m]{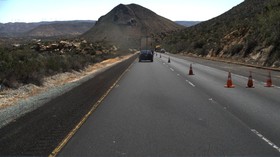}
			& \includegraphics[width=7em, valign=m]{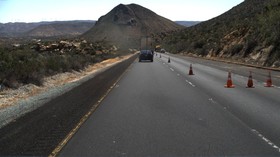}
			& \includegraphics[width=7em, valign=m]{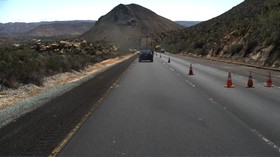}
			& \includegraphics[width=7em, valign=m]{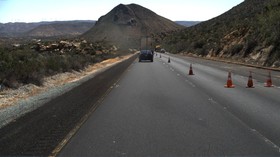}\vspace{0.3em}\\
			& \adjustbox{valign=m}{{\rotatebox{90}{}}}
			& \multicolumn{1}{c | }{\includegraphics[width=7em, valign=m]{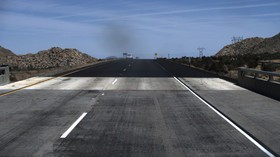}}
			& \includegraphics[width=7em, valign=m]{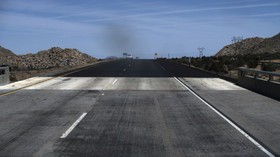}
			& \includegraphics[width=7em, valign=m]{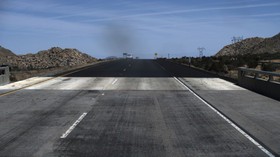}
			& \includegraphics[width=7em, valign=m]{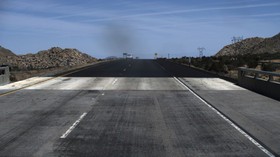}
			& \includegraphics[width=7em, valign=m]{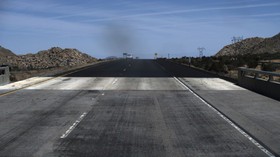}
			& \includegraphics[width=7em, valign=m]{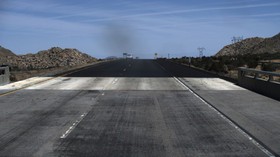}
			& \includegraphics[width=7em, valign=m]{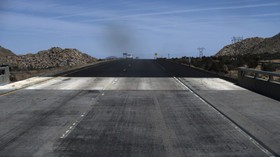}
			& \includegraphics[width=7em, valign=m]{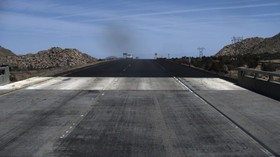}\vspace{0.3em}\\
			& \adjustbox{valign=m}{{\rotatebox{90}{}}}
			& \multicolumn{1}{c | }{\includegraphics[width=7em, valign=m]{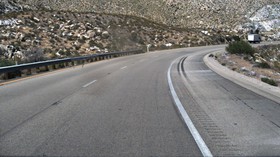}}
			& \includegraphics[width=7em, valign=m]{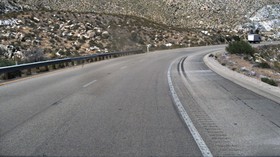}
			& \includegraphics[width=7em, valign=m]{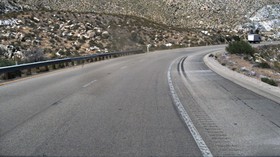}
			& \includegraphics[width=7em, valign=m]{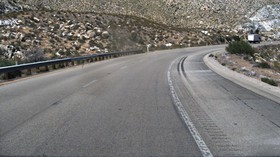}
			& \includegraphics[width=7em, valign=m]{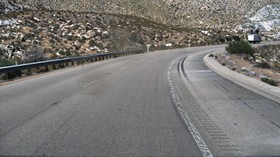}
			& \includegraphics[width=7em, valign=m]{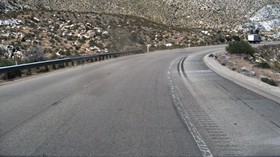}
			& \includegraphics[width=7em, valign=m]{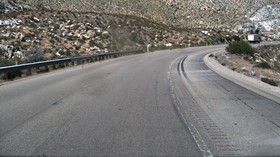}
			& \includegraphics[width=7em, valign=m]{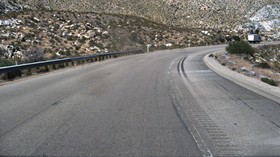}\vspace{0.3em}\\
			& \adjustbox{valign=m}{{\rotatebox{90}{}}}
			& \multicolumn{1}{c | }{\includegraphics[width=7em, valign=m]{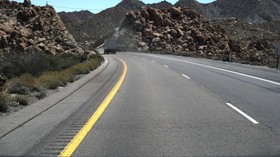}}
			& \includegraphics[width=7em, valign=m]{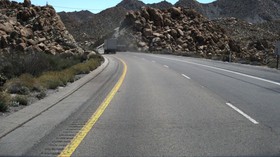}
			& \includegraphics[width=7em, valign=m]{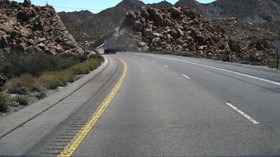}
			& \includegraphics[width=7em, valign=m]{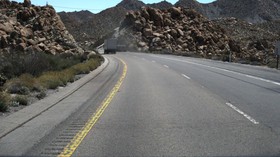}
			& \includegraphics[width=7em, valign=m]{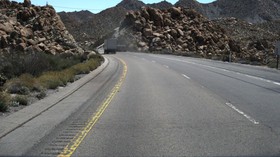}
			& \includegraphics[width=7em, valign=m]{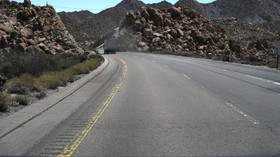}
			& \includegraphics[width=7em, valign=m]{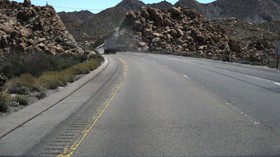}
			& \includegraphics[width=7em, valign=m]{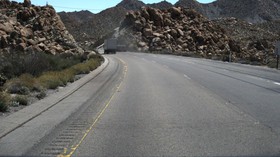}\vspace{0.3em}\\\vspace{-5px}
			& \adjustbox{valign=m}{{\rotatebox{90}{}}}
			& \multicolumn{1}{c | }{\includegraphics[width=7em, valign=m]{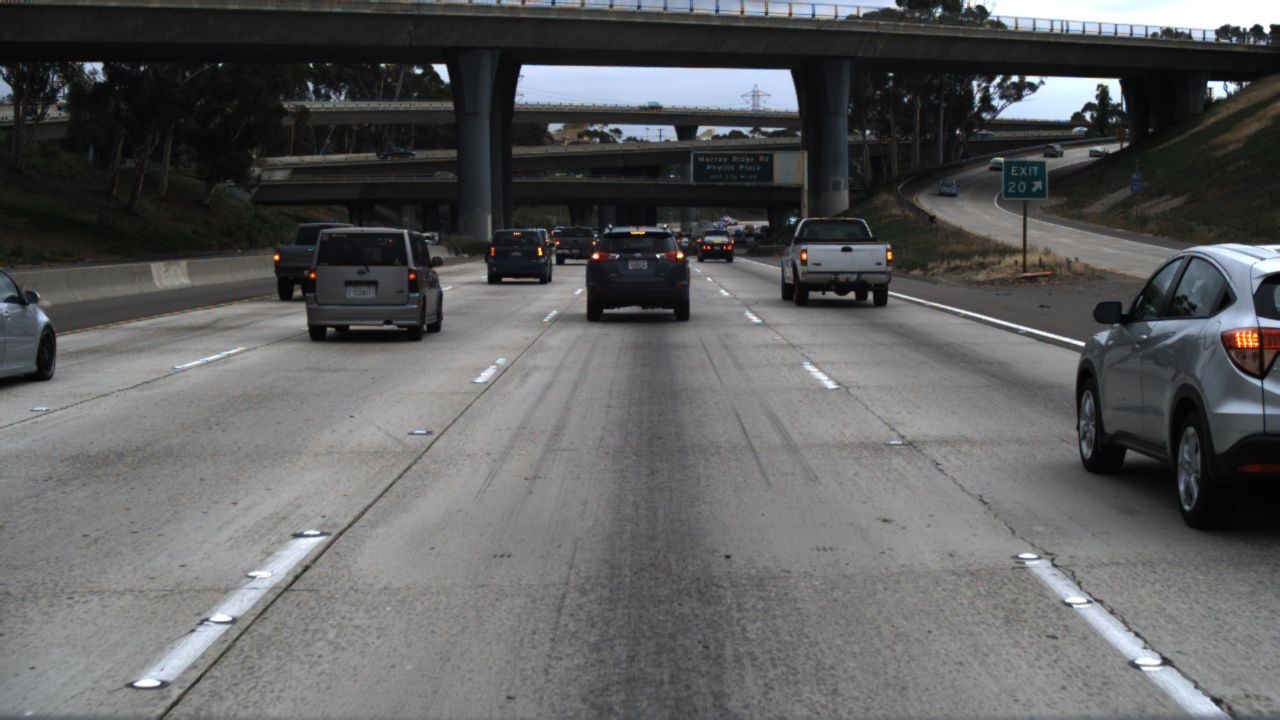}}
			& \includegraphics[width=7em, valign=m]{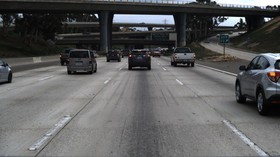}
			& \includegraphics[width=7em, valign=m]{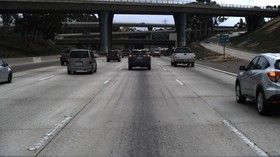}
			& \includegraphics[width=7em, valign=m]{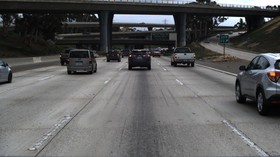}
			& \includegraphics[width=7em, valign=m]{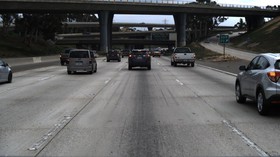}
			& \includegraphics[width=7em, valign=m]{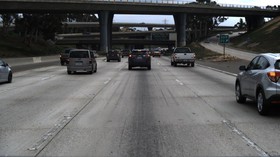}
			& \includegraphics[width=7em, valign=m]{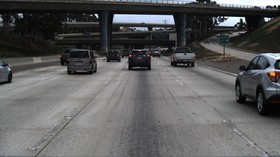}
			& \includegraphics[width=7em, valign=m]{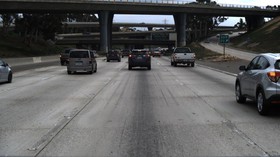}\vspace{1em}\\\vspace{-5px}
		    && & \tikzmark{a}{0.35} &&&&&& \tikzmark{b}{0.95} \\
		    &&&&&\multicolumn{2}{c}{\textcolor{bostonuniversityred}{}}&&&&\\

			\multicolumn{11}{c}{{Task without interpolation}}\\
			\toprule

			\adjustbox{valign=m}{\multirow{2}{*}{\rotatebox{90}{\textbf{Snow addition}}}}
			& \adjustbox{valign=m}{{\rotatebox{90}{Original}}}
			& \includegraphics[width=7em, valign=m]{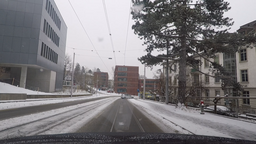}
			& \includegraphics[width=7em, valign=m]{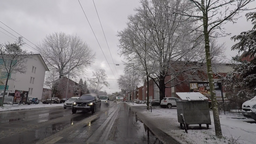}
			& \includegraphics[width=7em, valign=m]{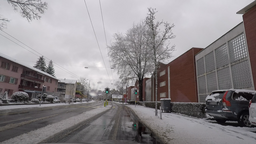}
			& \includegraphics[width=7em, valign=m]{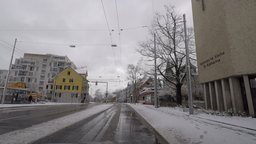}
			& \includegraphics[width=7em, valign=m]{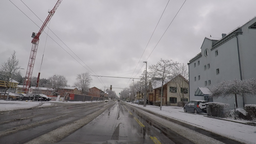}
			& \includegraphics[width=7em, valign=m]{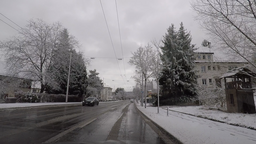}
			& \includegraphics[width=7em, valign=m]{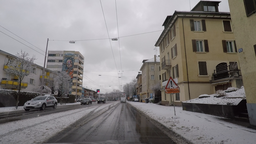}
			& \includegraphics[width=7em, valign=m]{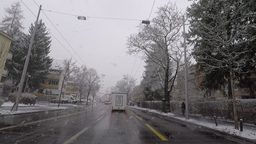}\vspace{0.3em}\\
			& \adjustbox{valign=m}{{\rotatebox{90}{Ours}}}
			& \includegraphics[width=7em, valign=m]{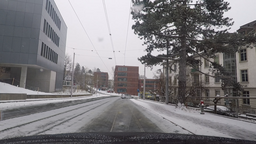}
			& \includegraphics[width=7em, valign=m]{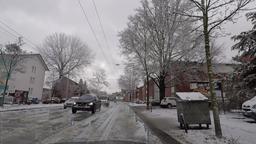}
			& \includegraphics[width=7em, valign=m]{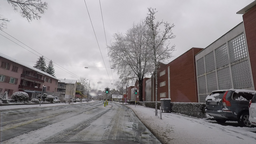}
			& \includegraphics[width=7em, valign=m]{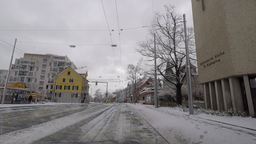}
			& \includegraphics[width=7em, valign=m]{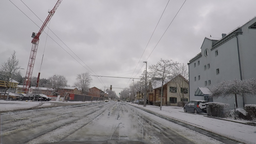}
			& \includegraphics[width=7em, valign=m]{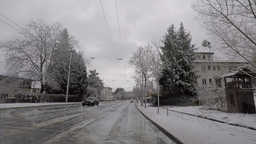}
			& \includegraphics[width=7em, valign=m]{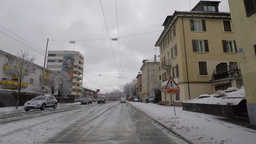}
			& \includegraphics[width=7em, valign=m]{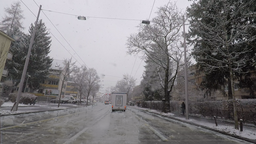}\vspace{0.3em}\\\midrule

			\adjustbox{valign=m}{\multirow{4}{*}[-1.3em]{\rotatebox{90}{\textbf{Deblurring}}}}
			& \adjustbox{valign=m}{{\rotatebox{90}{Original}}}
			& \flowerimg{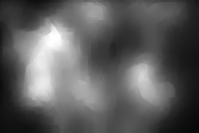}
			& \flowerimg{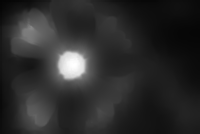}
			& \flowerimg{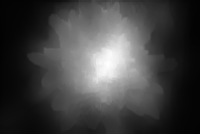}
			& \flowerimg{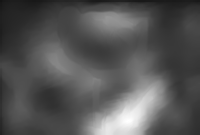}
			& \flowerimg{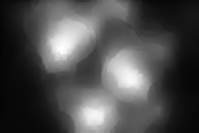}
			& \flowerimg{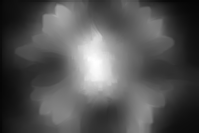}
			& \flowerimg{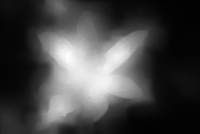}
			& \flowerimg{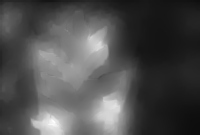}\vspace{0.3em}\\

			& \adjustbox{valign=m}{{\rotatebox{90}{Ours}}}
			& \flowerimg{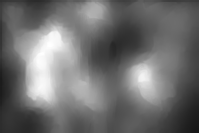}
			& \flowerimg{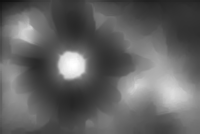}
			& \flowerimg{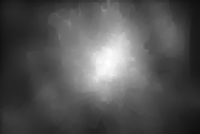}
			& \flowerimg{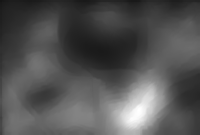}
			& \flowerimg{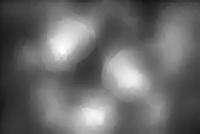}
			& \flowerimg{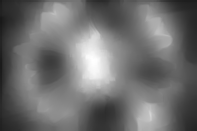}
			& \flowerimg{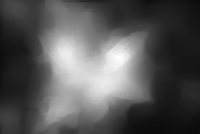}
			& \flowerimg{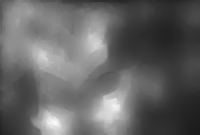}\vspace{0.3em}\\

			\bottomrule
	\end{tabular}\linkwithlabel{a}{b}{$z$}}
	\caption{\textbf{Qualitative results.} Additional qualitative results for the lane degradation, snow addition and deblurring tasks.} \label{fig:qualitative-alltasks_supp}
\end{figure*}

Additional i2i qualitative results are shown in Fig.~\ref{fig:qualitative-alltasks_supp}. With our pipeline, we are able to realistically modify TuSimple~\cite{tusimple} street lines with different degrees of degradation, independently from the lane marking type (continuous, dashed, etc.) and its color (white, yellow, etc.). Furthermore, we can add plausible snow on ACDC~\cite{sakaridis2021acdc} roads while preserving shadows. In addition, our model can turn Oxford Flowers dataset~\cite{oxfordflowersdataset} shallow DoF photos into deep DoF ones, yielding in-focus backgrounds with sharper edges.

\subsubsection{Proxy tasks}\label{sec:proxy-tasks}
\begin{figure*}
	\centering
	\resizebox{\linewidth}{!}{
		\setlength{\tabcolsep}{0.003\linewidth}

		\tiny
		\begin{tabular}{c c c c c c c c c c}

            \toprule
			\adjustbox{valign=m}{\multirow{3}{*}[-2.5em]{\rotatebox{90}{\textbf{SCNN}~\cite{pan2017spatial}}}}
			& \adjustbox{valign=m}{{\rotatebox{90}{Input}}}
			& \includegraphics[width=7em, valign=m]{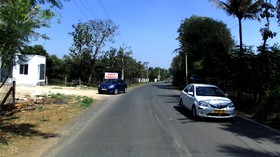}
            & \includegraphics[width=7em, valign=m]{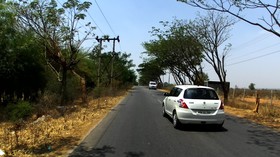}
            & \includegraphics[width=7em, valign=m]{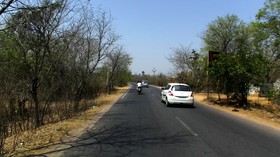}
            & \includegraphics[width=7em, valign=m]{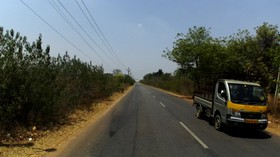}
            & \includegraphics[width=7em, valign=m]{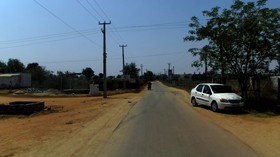}
            & \includegraphics[width=7em, valign=m]{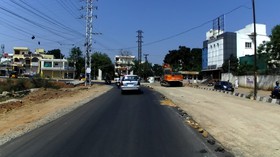}
            & \includegraphics[width=7em, valign=m]{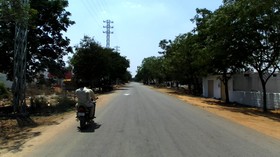}
            & \includegraphics[width=7em, valign=m]{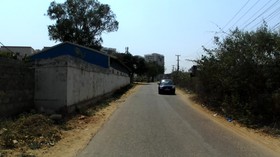}\vspace{0.3em}\\
			& \adjustbox{valign=m}{{\rotatebox{90}{none}}}
			& \includegraphics[width=7em, valign=m]{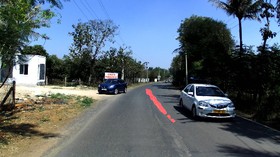}
			& \includegraphics[width=7em, valign=m]{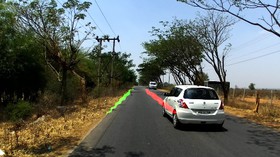}
			& \includegraphics[width=7em, valign=m]{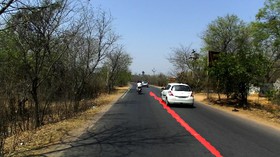}
			& \includegraphics[width=7em, valign=m]{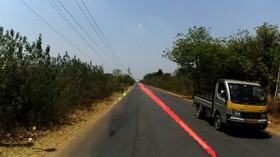}
			& \includegraphics[width=7em, valign=m]{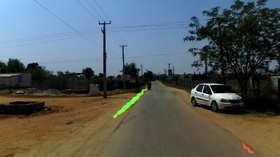}
			& \includegraphics[width=7em, valign=m]{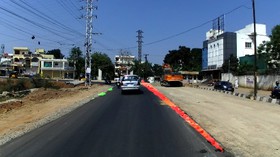}
			& \includegraphics[width=7em, valign=m]{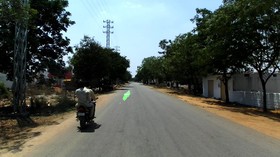}
			& \includegraphics[width=7em, valign=m]{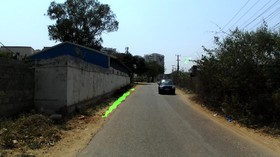}\vspace{0.3em}\\
			& \adjustbox{valign=m}{{\rotatebox{90}{Ours}}}
			& \includegraphics[width=7em, valign=m]{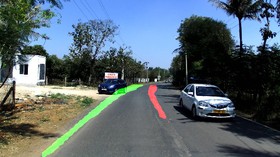}
			& \includegraphics[width=7em, valign=m]{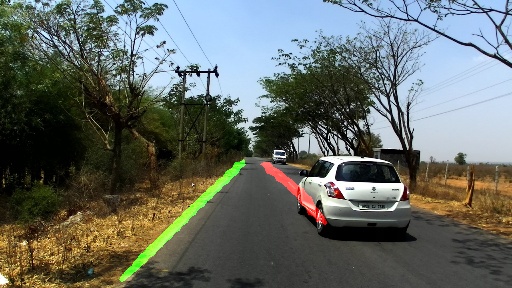}
			& \includegraphics[width=7em, valign=m]{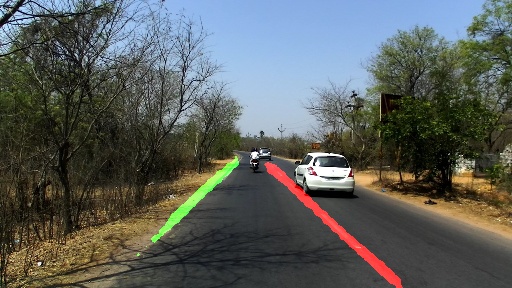}
			& \includegraphics[width=7em, valign=m]{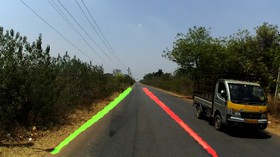}
			& \includegraphics[width=7em, valign=m]{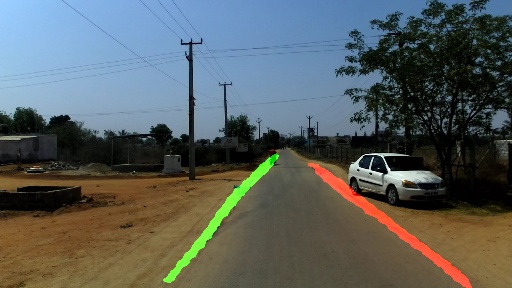}
			& \includegraphics[width=7em, valign=m]{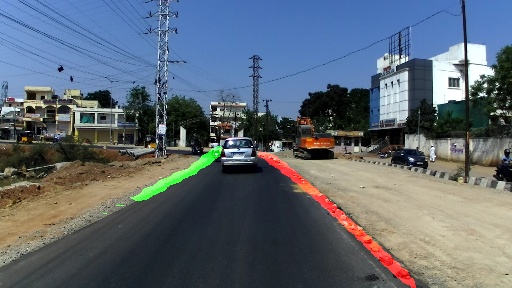}
			& \includegraphics[width=7em, valign=m]{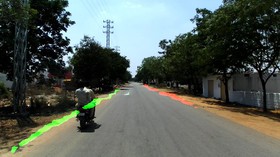}
			& \includegraphics[width=7em, valign=m]{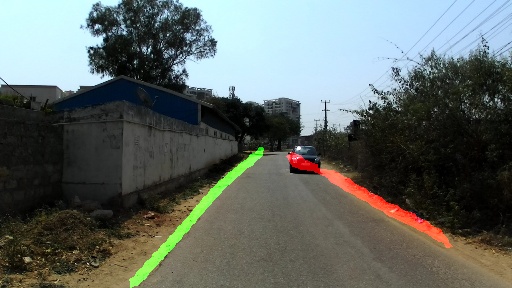}\vspace{0.3em}\\
			
			\midrule
			\adjustbox{valign=m}{\multirow{3}{*}[-2.5em]{\rotatebox{90}{\textbf{RESA}~\cite{zheng2021resa}}}}
			& \adjustbox{valign=m}{{\rotatebox{90}{Input}}}
			& \includegraphics[width=7em, valign=m]{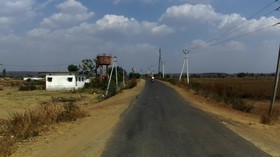}
			& \includegraphics[width=7em, valign=m]{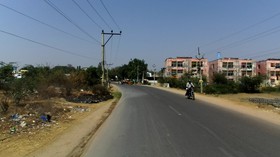}
			& \includegraphics[width=7em, valign=m]{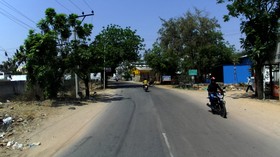}
			& \includegraphics[width=7em, valign=m]{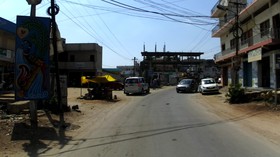}
			& \includegraphics[width=7em, valign=m]{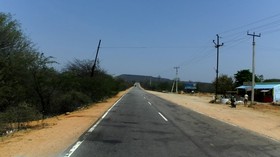}
			& \includegraphics[width=7em, valign=m]{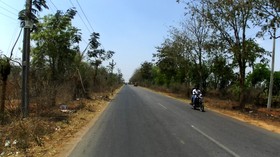}
			& \includegraphics[width=7em, valign=m]{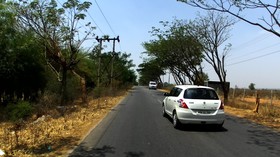}
			& \includegraphics[width=7em, valign=m]{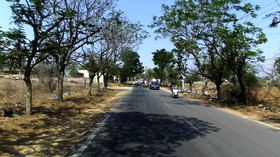}\vspace{0.3em}\\
			
			& \adjustbox{valign=m}{{\rotatebox{90}{none}}}
			& \includegraphics[width=7em, valign=m]{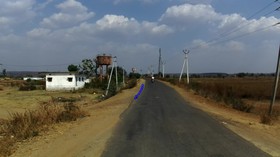}
			& \includegraphics[width=7em, valign=m]{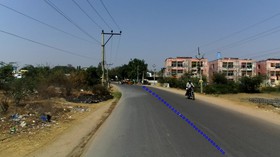}
			& \includegraphics[width=7em, valign=m]{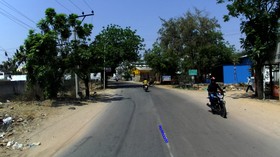}
			& \includegraphics[width=7em, valign=m]{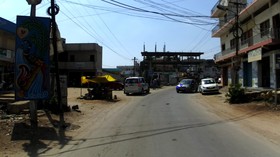}
			& \includegraphics[width=7em, valign=m]{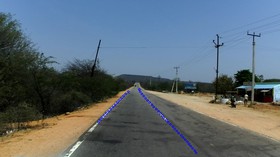}
			& \includegraphics[width=7em, valign=m]{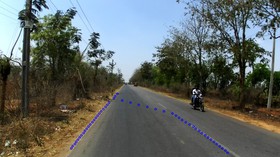}
			& \includegraphics[width=7em, valign=m]{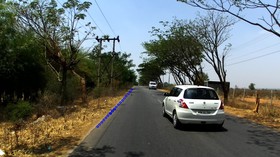}
			& \includegraphics[width=7em, valign=m]{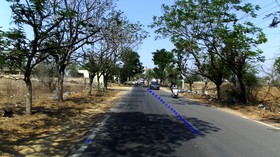}\vspace{0.3em}\\
			& \adjustbox{valign=m}{{\rotatebox{90}{Ours}}}
			& \includegraphics[width=7em, valign=m]{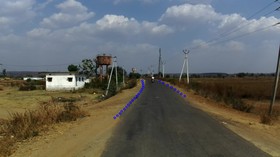}
			& \includegraphics[width=7em, valign=m]{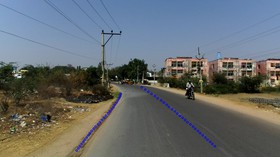}
			& \includegraphics[width=7em, valign=m]{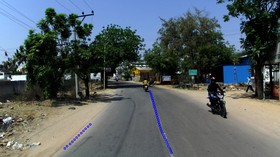}
			& \includegraphics[width=7em, valign=m]{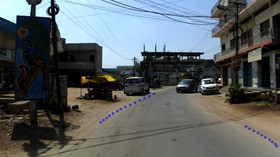}
			& \includegraphics[width=7em, valign=m]{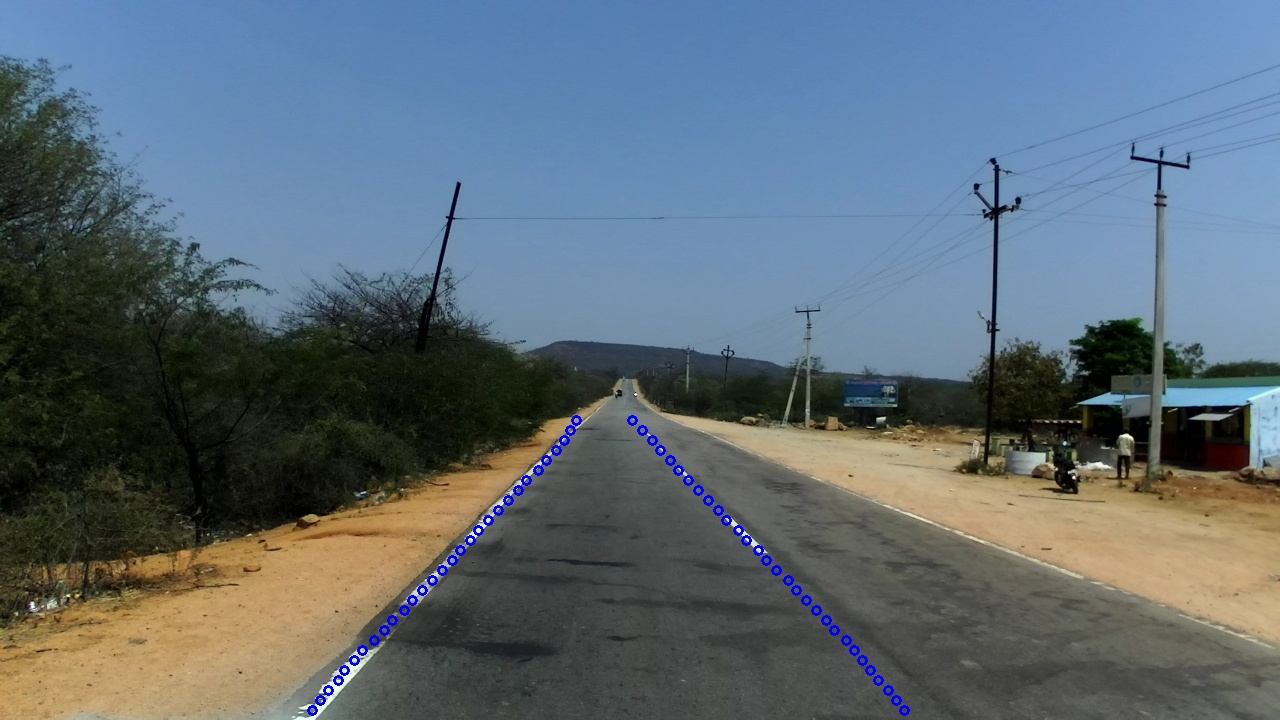}
			& \includegraphics[width=7em, valign=m]{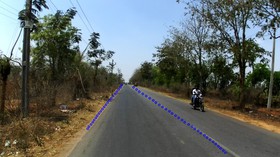}
			& \includegraphics[width=7em, valign=m]{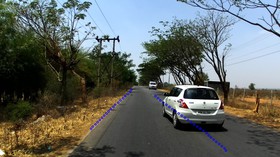}
			& \includegraphics[width=7em, valign=m]{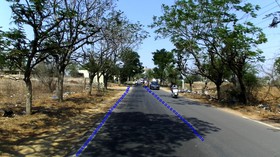}\vspace{0.3em}\\
			\bottomrule
	\end{tabular}}
	\caption{\textbf{Additional qualitative results on lane detection.} We train SCNN~\cite{pan2017spatial} and RESA~\cite{zheng2021resa} on images generated with our method and test on real images with damaged lane markings of IDD~\cite{varma2018idd}, improving performances.} \label{fig:qualit-lanedetIDD_supp}
\end{figure*}

\begin{figure*}
	\centering
	\resizebox{\linewidth}{!}{
		\setlength{\tabcolsep}{0.003\linewidth}

		\tiny
		\begin{tabular}{c c c c c c c c c c}

            \toprule
			\adjustbox{valign=m}{{\rotatebox{90}{}}}
			& \adjustbox{valign=m}{{\rotatebox{90}{Input}}}
			& \includegraphics[width=7em, valign=m]{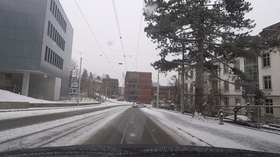}
			& \includegraphics[width=7em, valign=m]{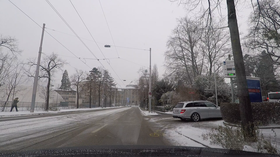}
			& \includegraphics[width=7em, valign=m]{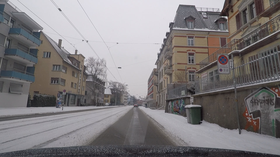}
			& \includegraphics[width=7em, valign=m]{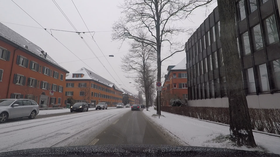}
			& \includegraphics[width=7em, valign=m]{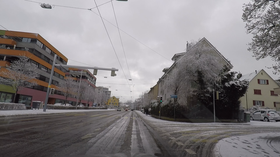}
			& \includegraphics[width=7em, valign=m]{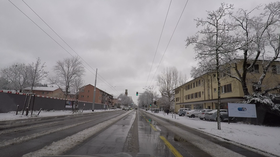}
			& \includegraphics[width=7em, valign=m]{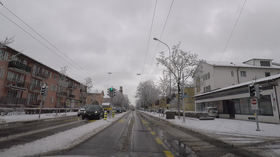}
			& \includegraphics[width=7em, valign=m]{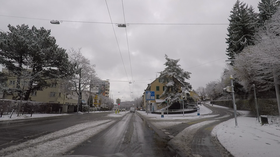}\vspace{0.3em}\\
            \adjustbox{valign=m}{{\rotatebox{90}{}}}
			& \adjustbox{valign=m}{{\rotatebox{90}{GT}}}
			& \includegraphics[width=7em, valign=m]{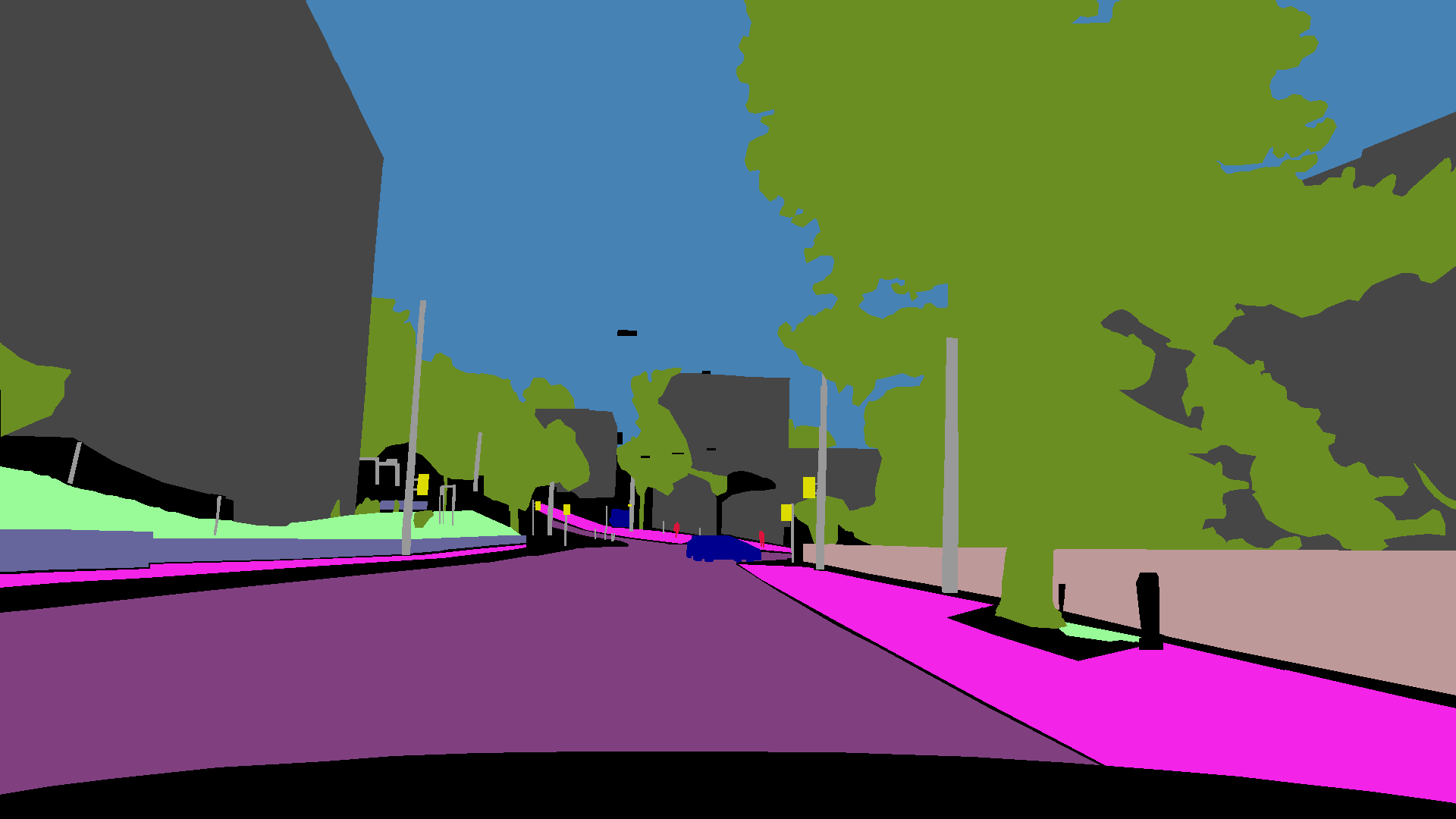}
			& \includegraphics[width=7em, valign=m]{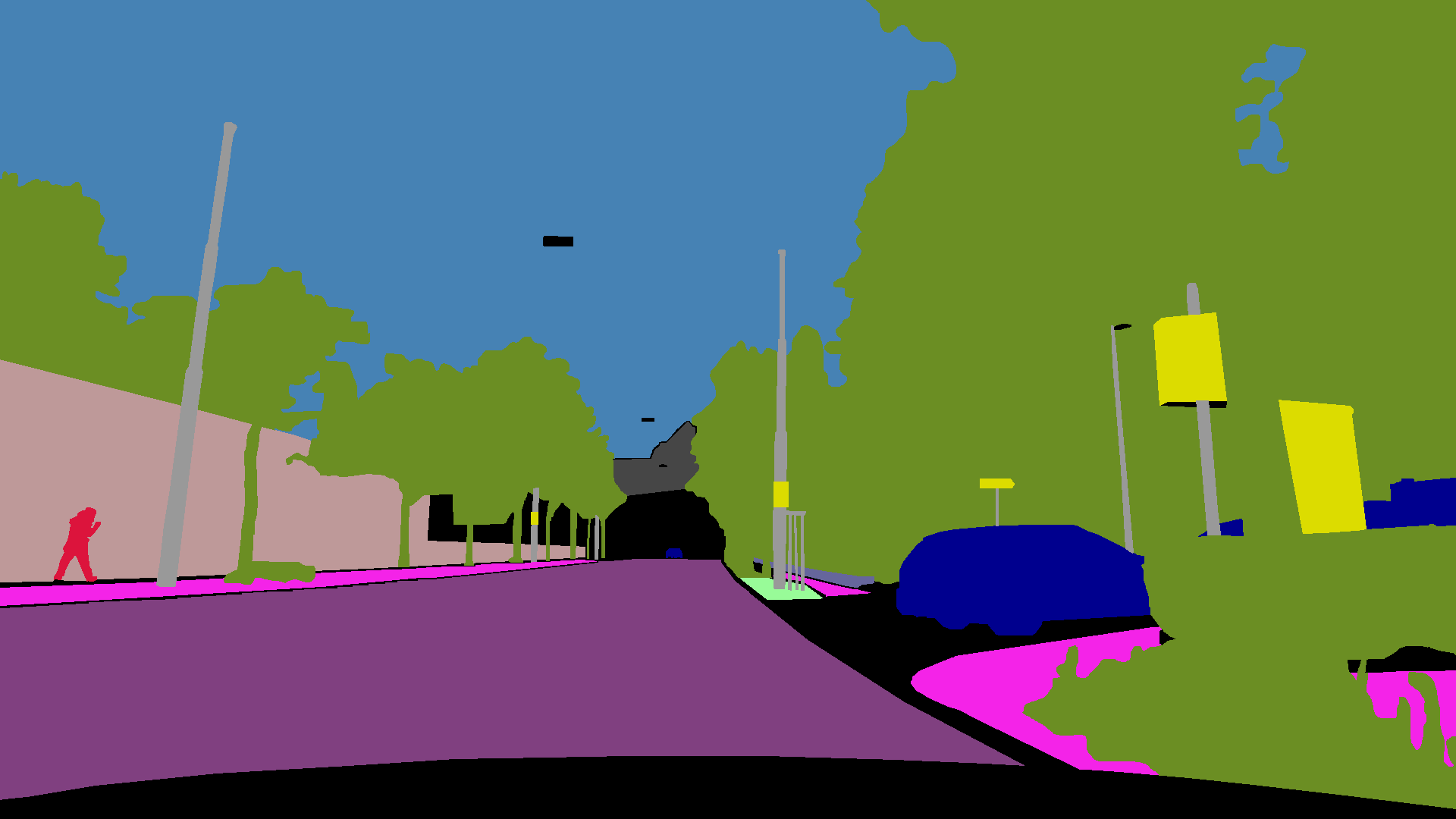}
			& \includegraphics[width=7em, valign=m]{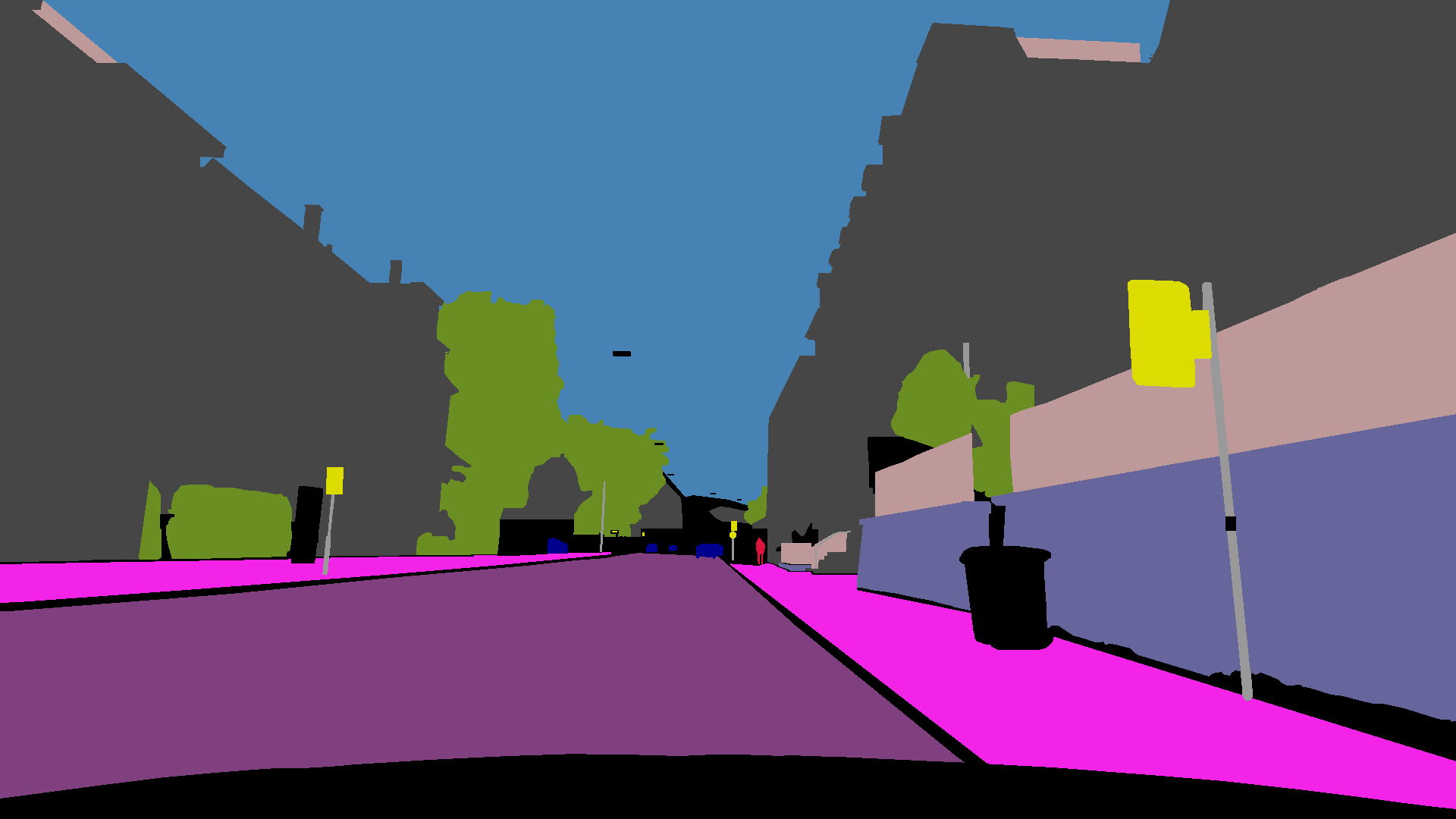}
			& \includegraphics[width=7em, valign=m]{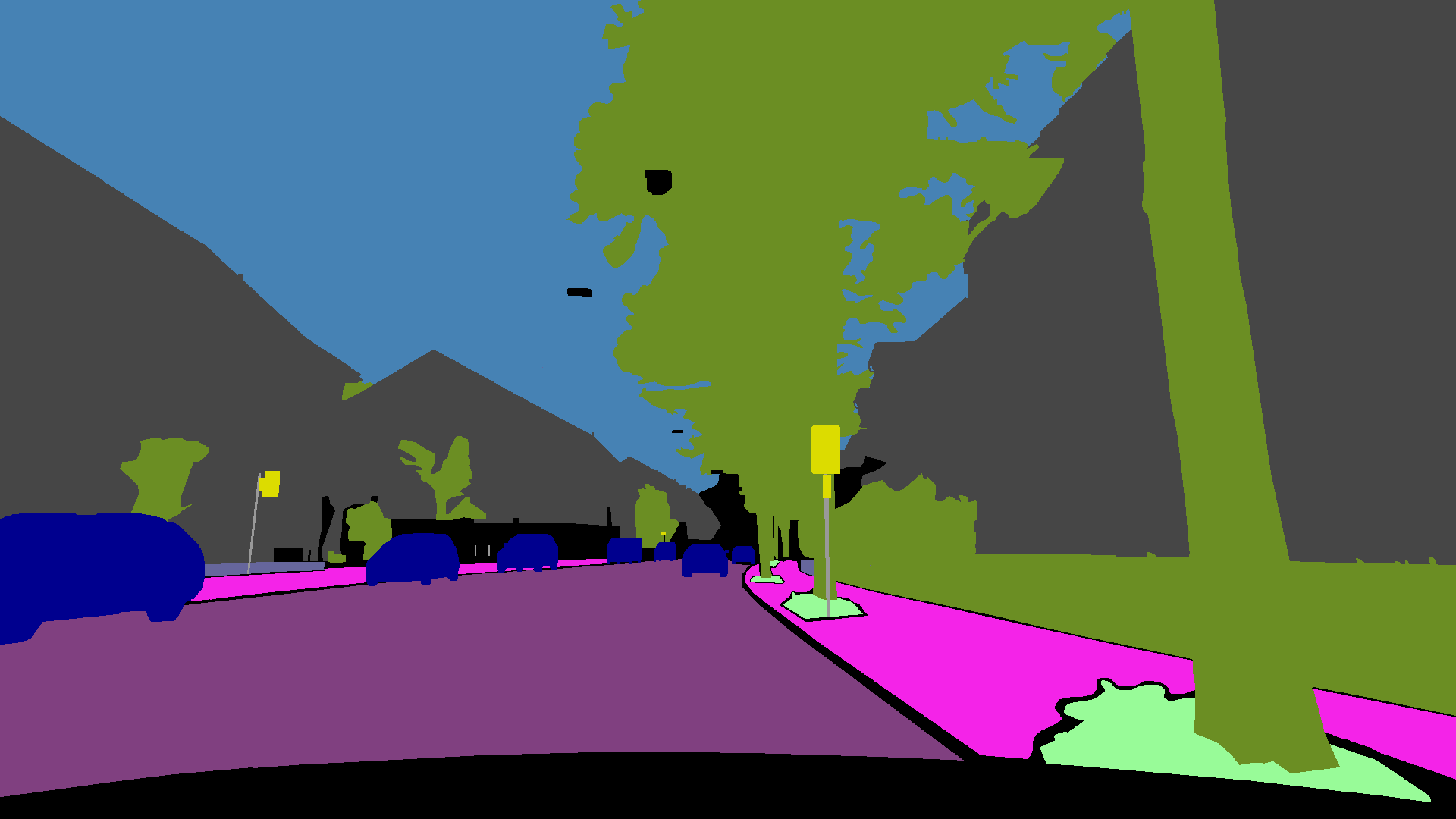}
			& \includegraphics[width=7em, valign=m]{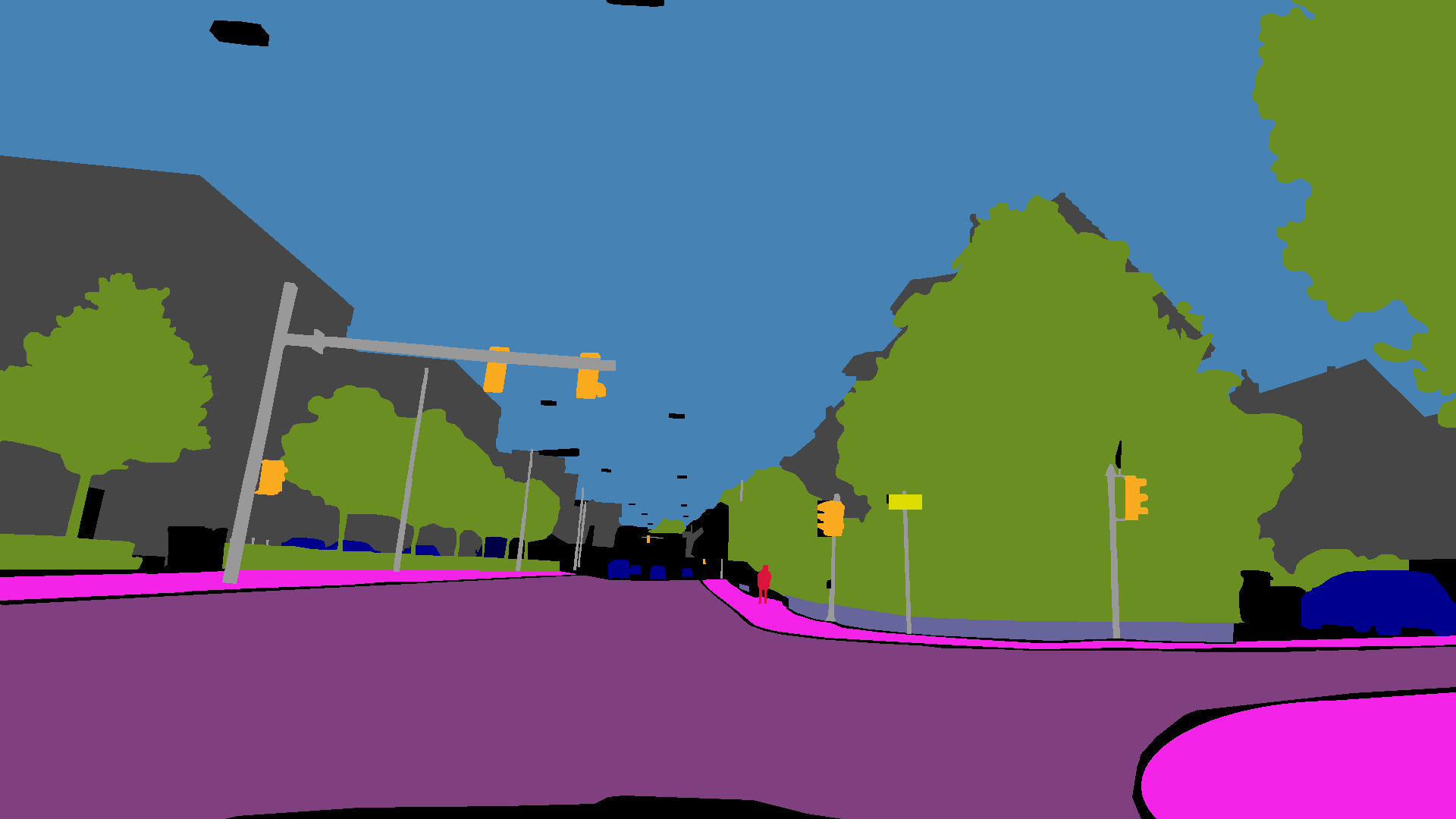}
			& \includegraphics[width=7em, valign=m]{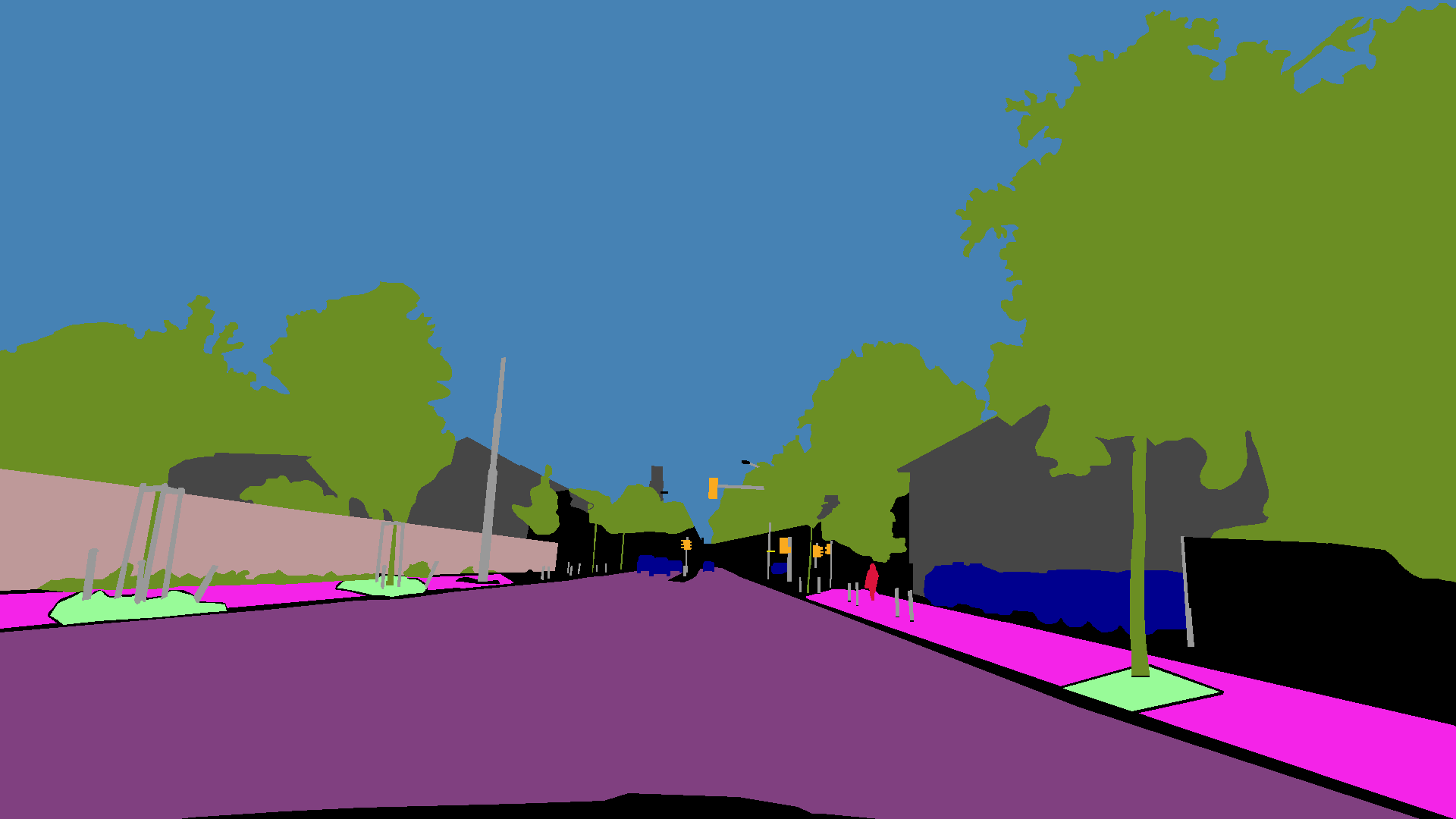}
			& \includegraphics[width=7em, valign=m]{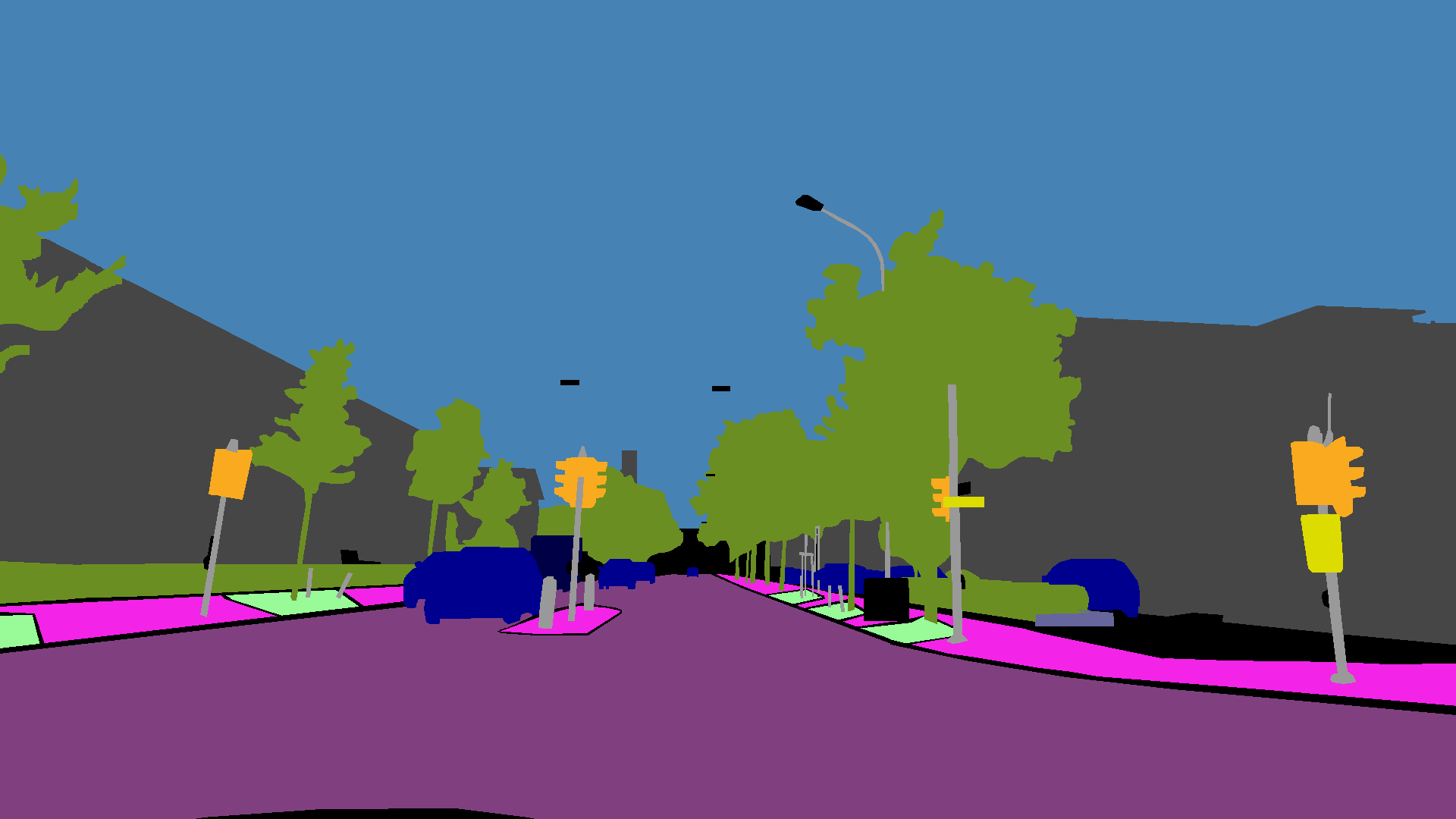}
			& \includegraphics[width=7em, valign=m]{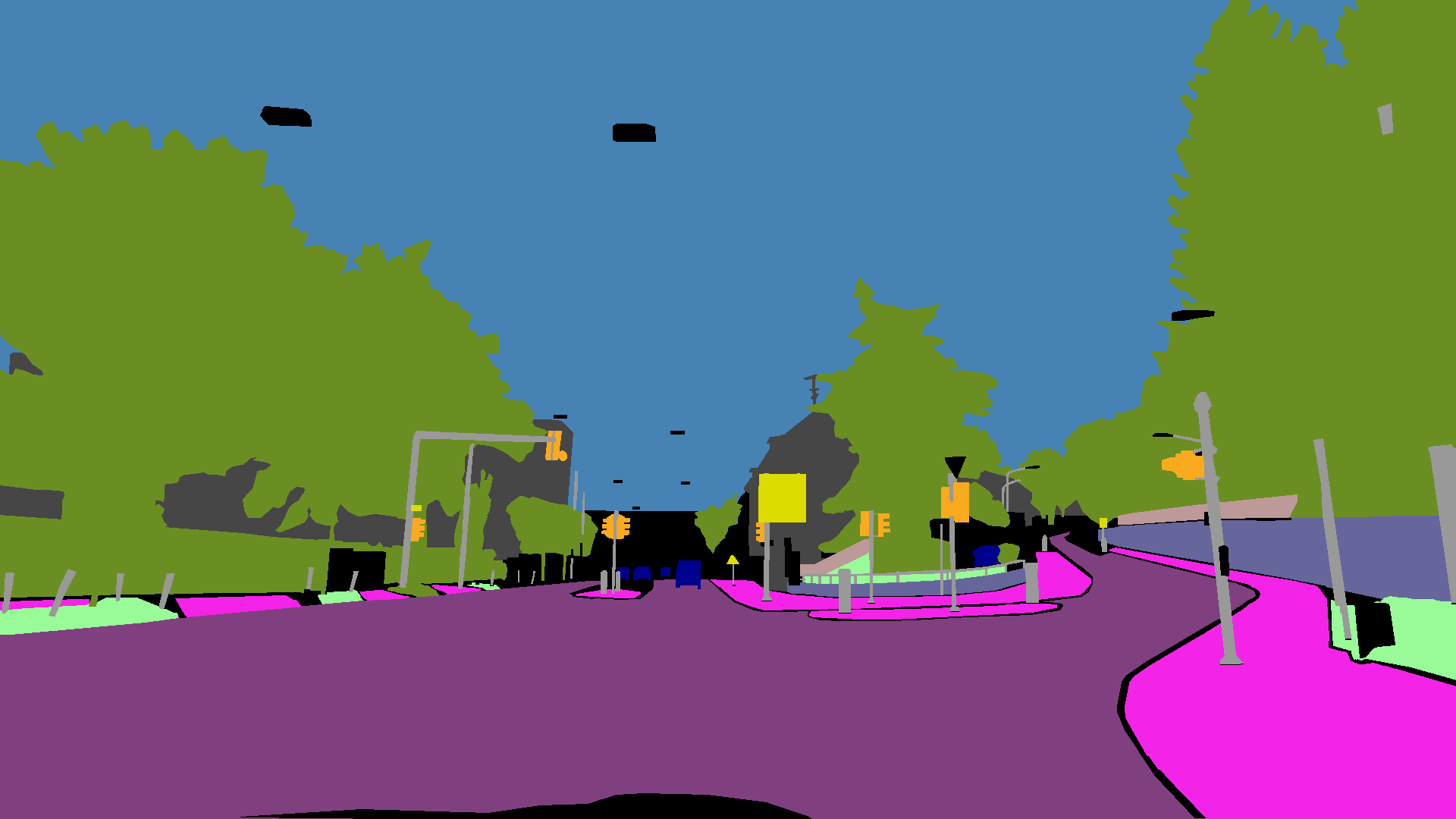}\vspace{0.3em}\\
            
            \midrule
            \adjustbox{valign=m}{\multirow{2}{*}[0.5em]{\rotatebox{90}{\textbf{DeepLabv3+}~\cite{chen2018encoderdecoder}}}}
			& \adjustbox{valign=m}{{\rotatebox{90}{none}}}
			& \includegraphics[width=7em, valign=m]{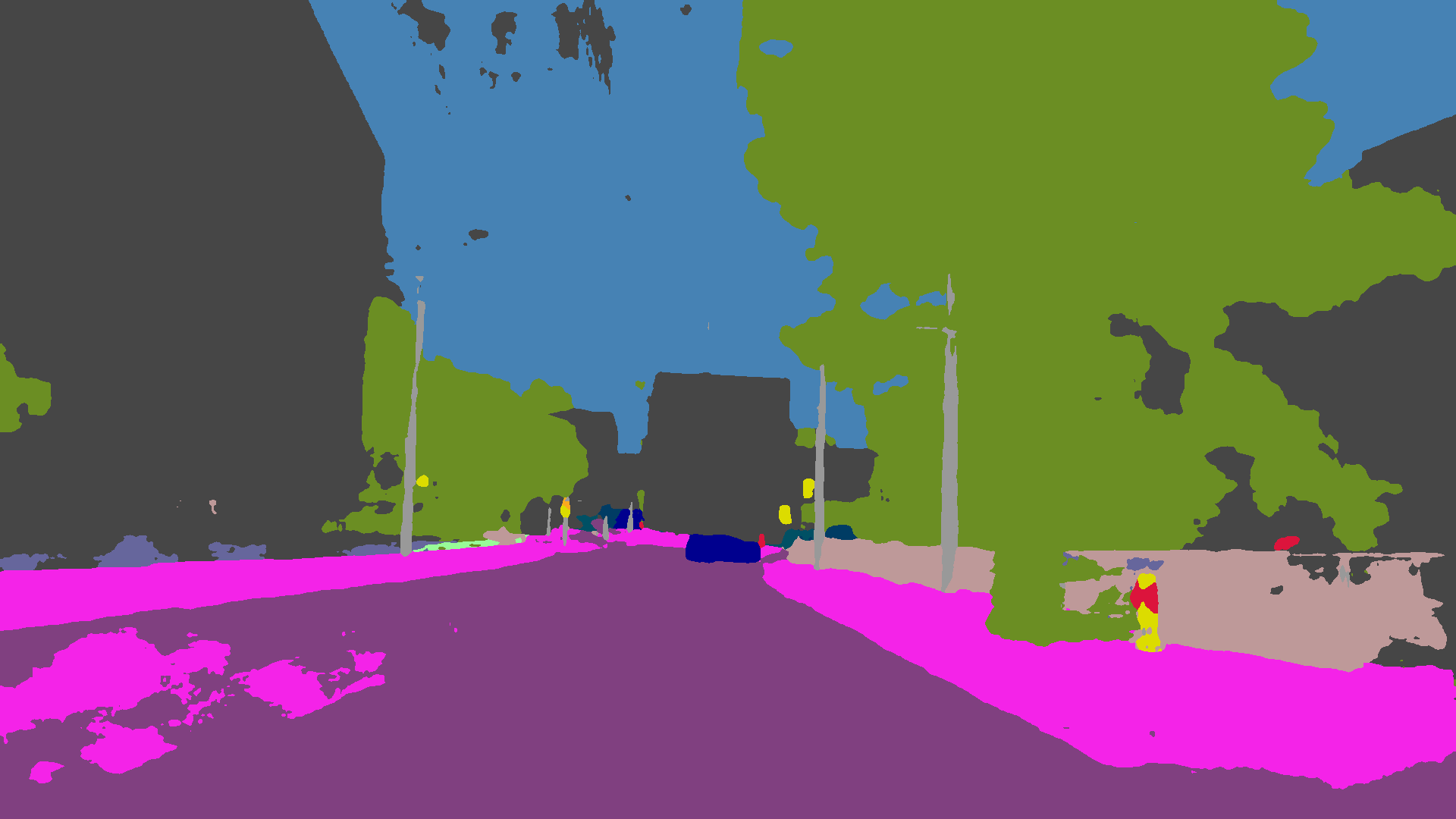}
			& \includegraphics[width=7em, valign=m]{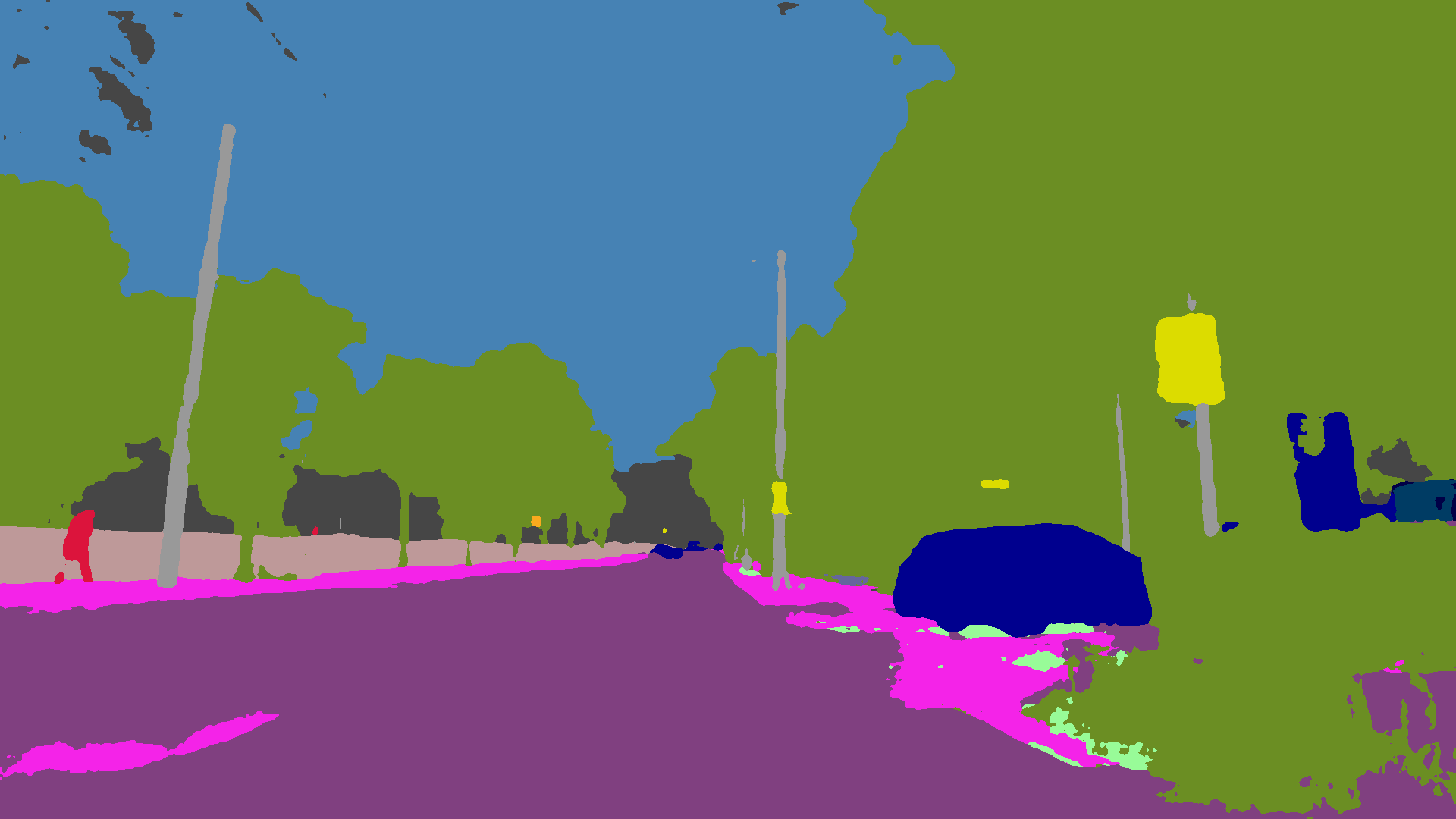}
			& \includegraphics[width=7em, valign=m]{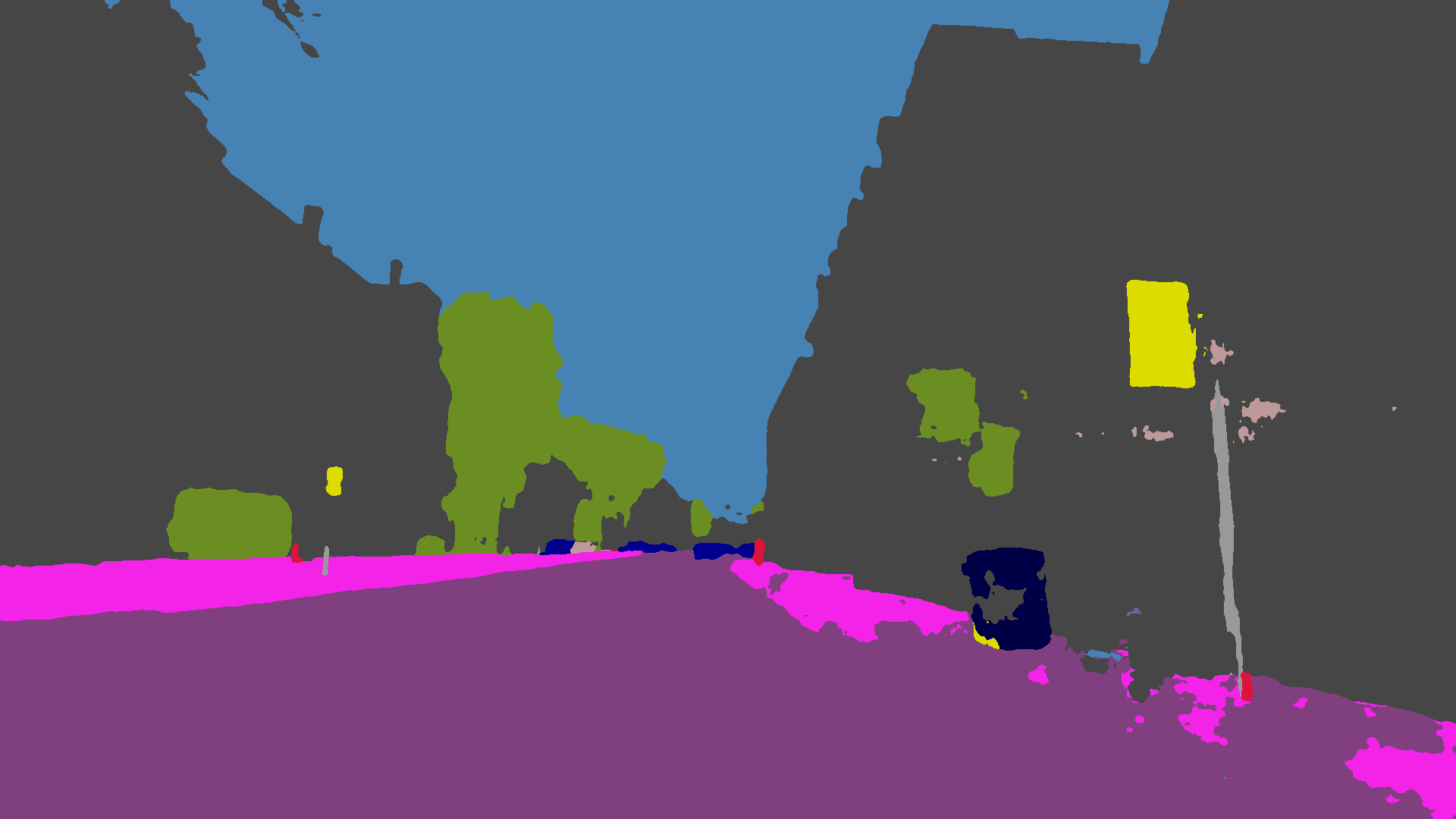}
			& \includegraphics[width=7em, valign=m]{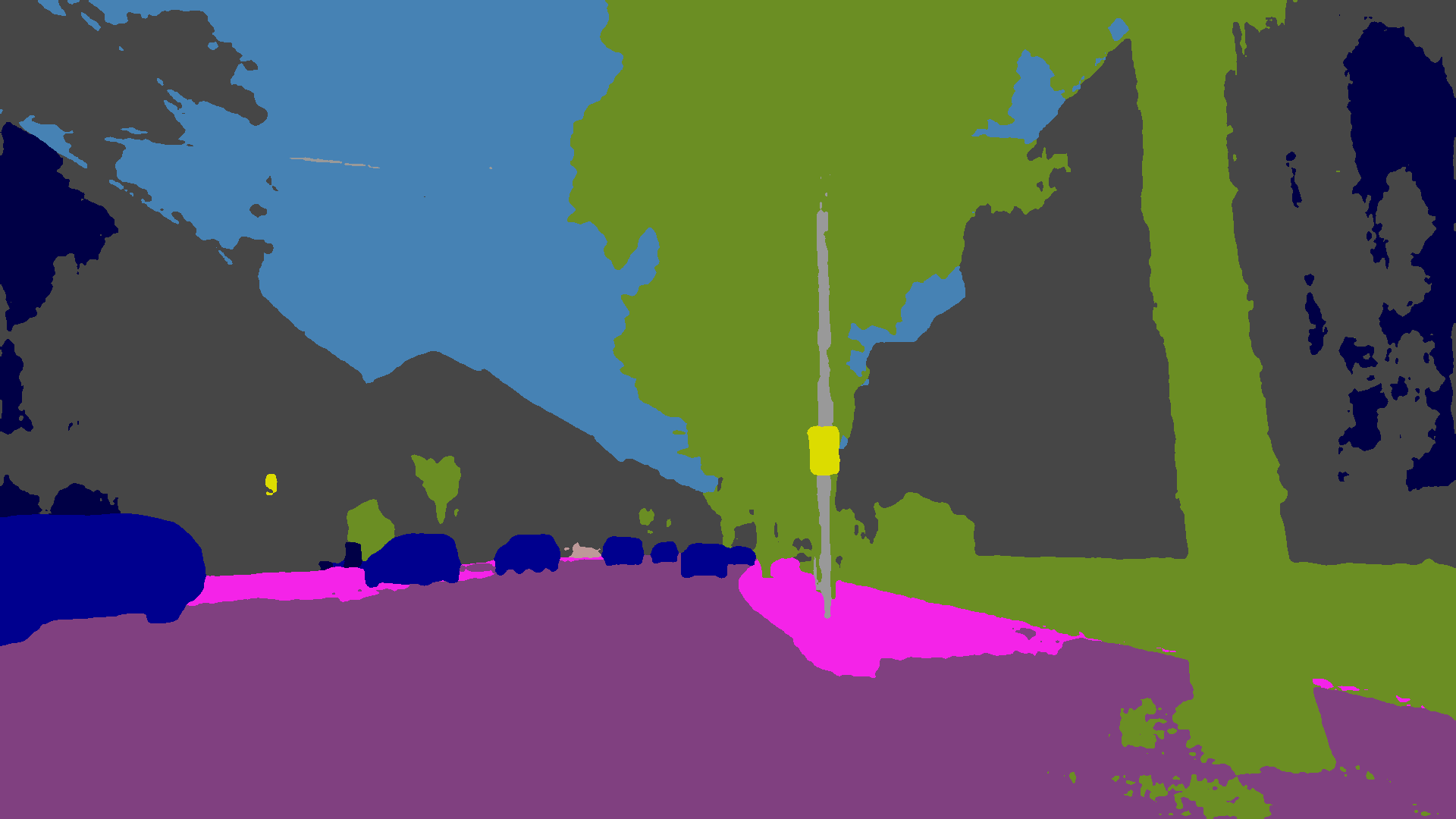}
			& \includegraphics[width=7em, valign=m]{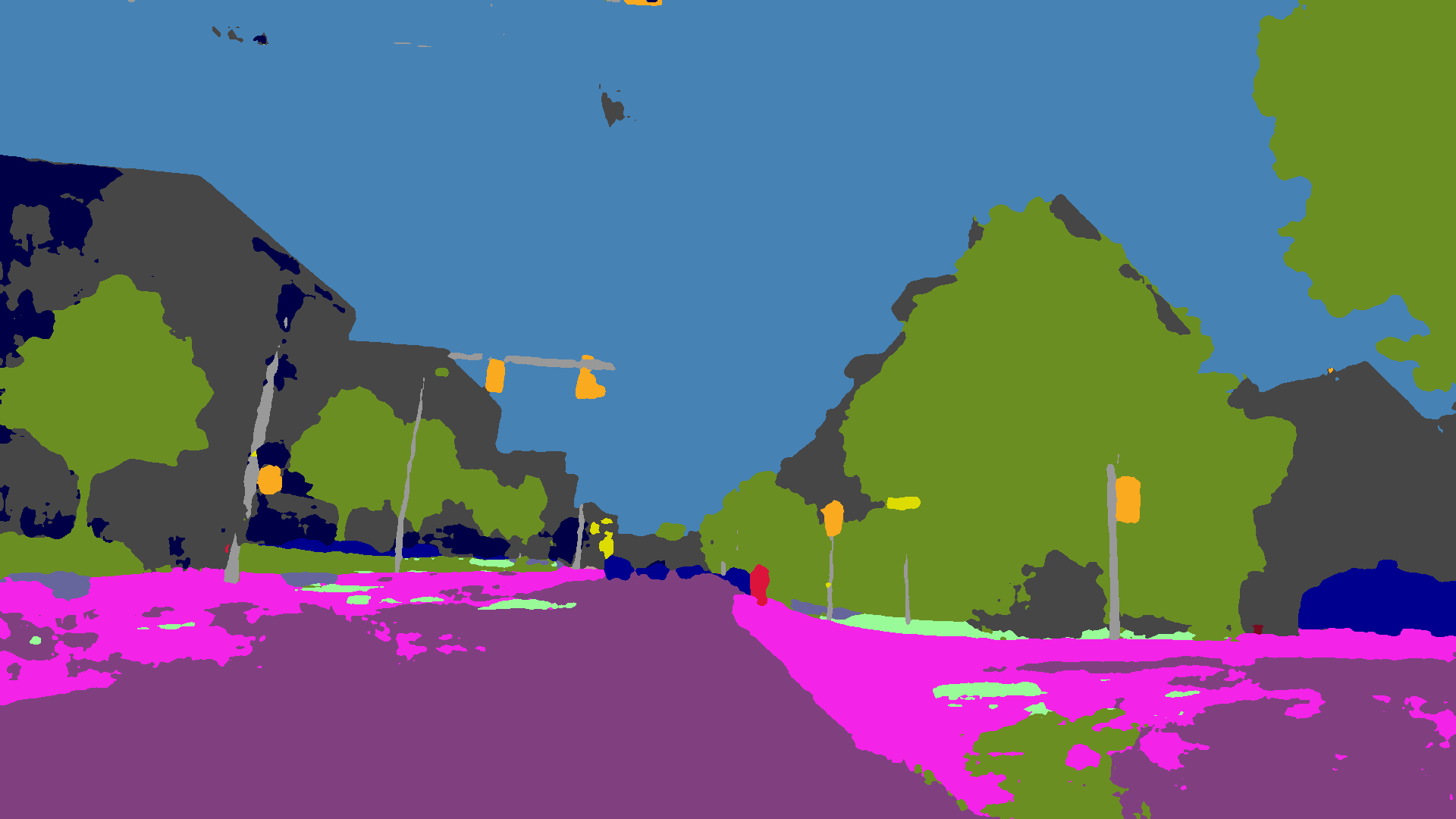}
			& \includegraphics[width=7em, valign=m]{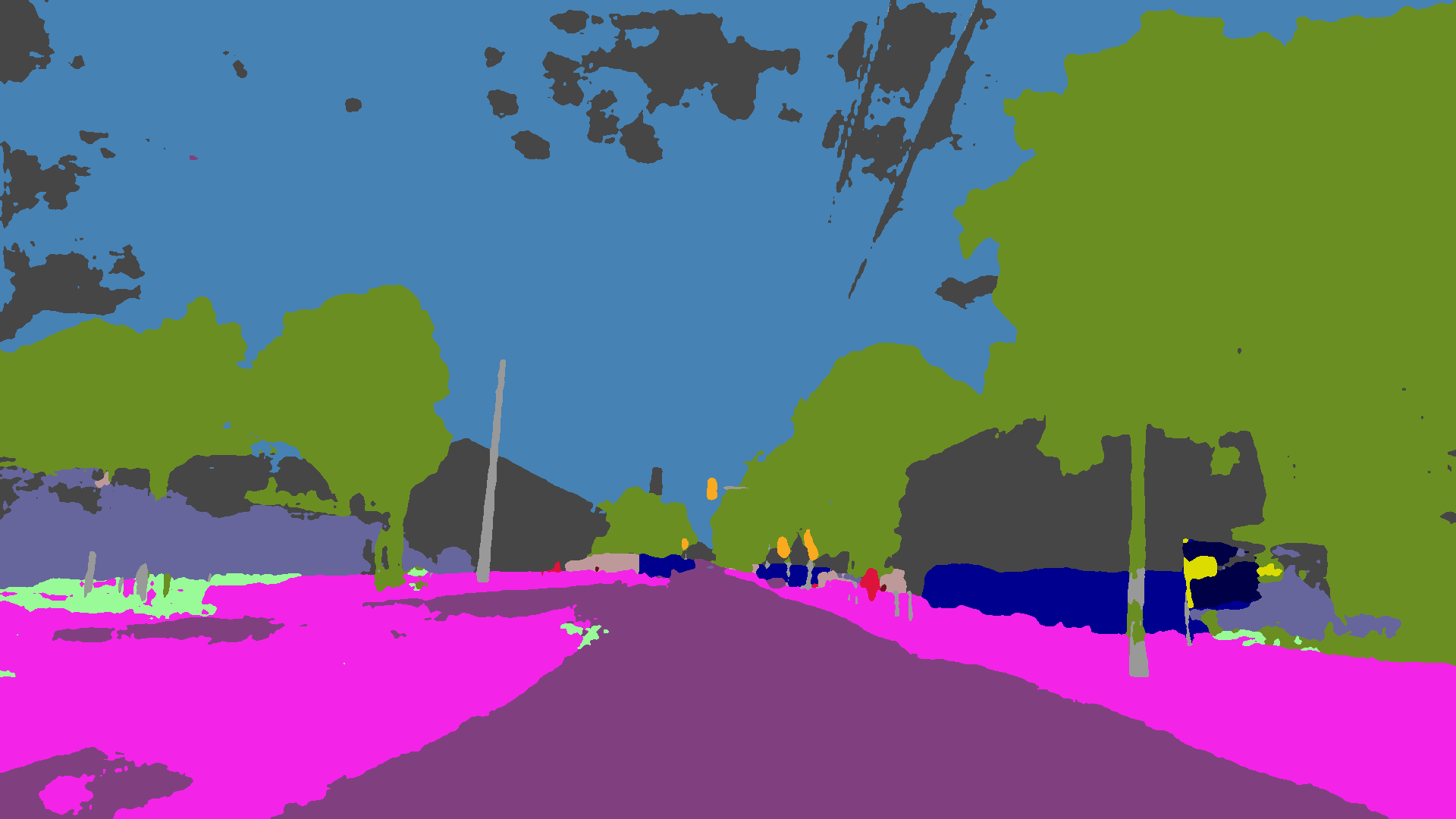}
			& \includegraphics[width=7em, valign=m]{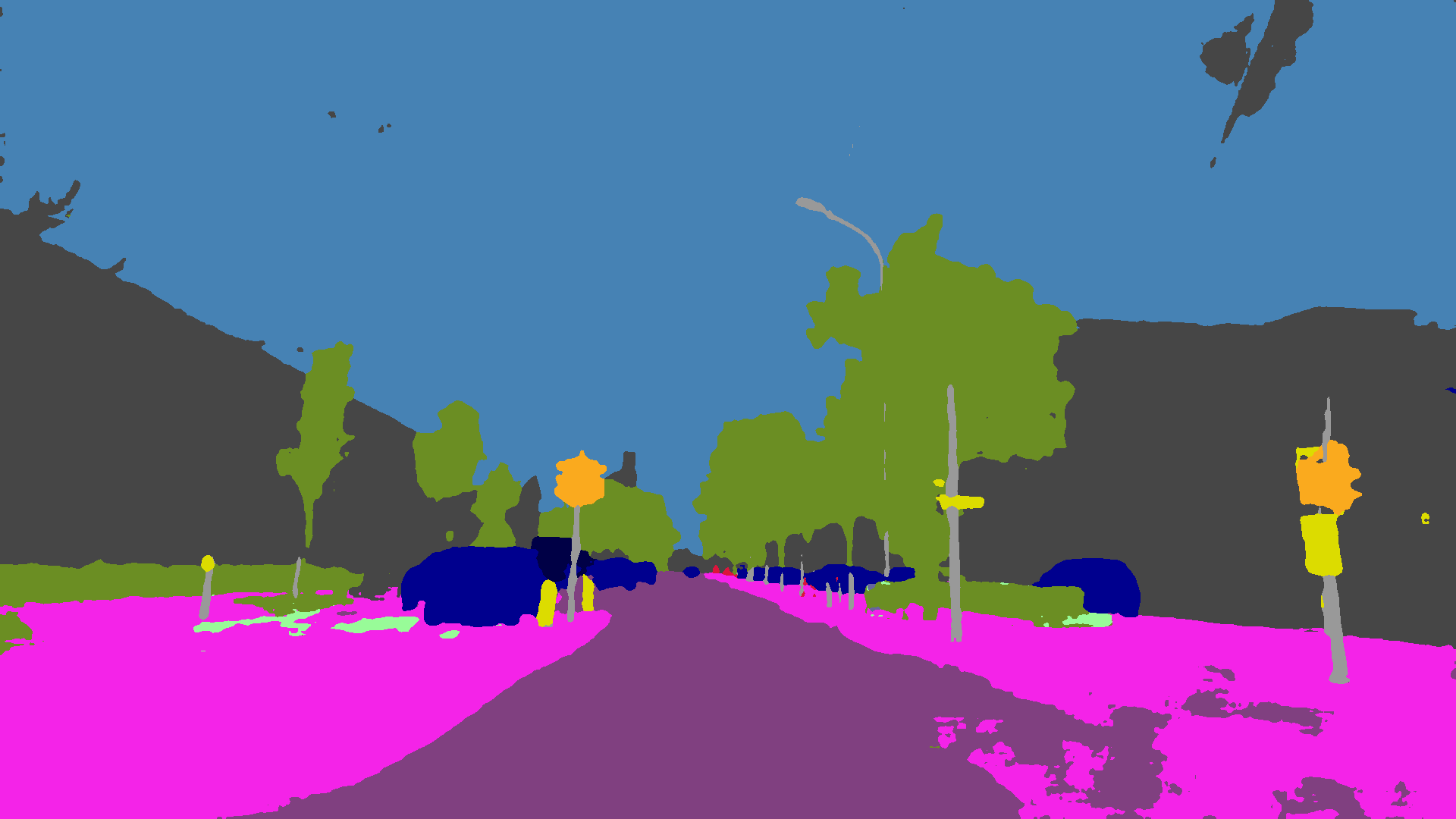}
			& \includegraphics[width=7em, valign=m]{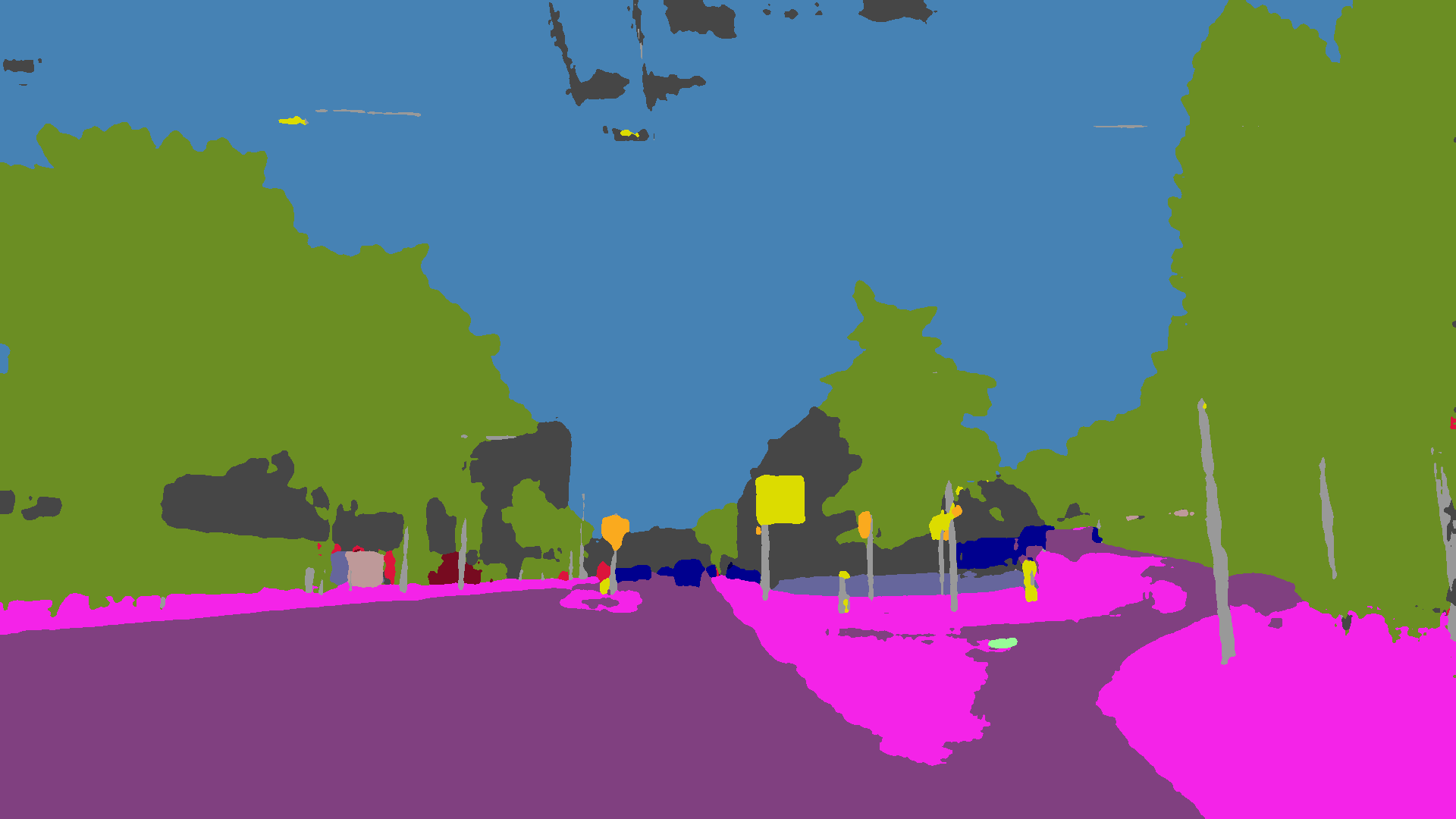}\vspace{0.3em}\\
			& \adjustbox{valign=m}{{\rotatebox{90}{Ours}}}
			& \includegraphics[width=7em, valign=m]{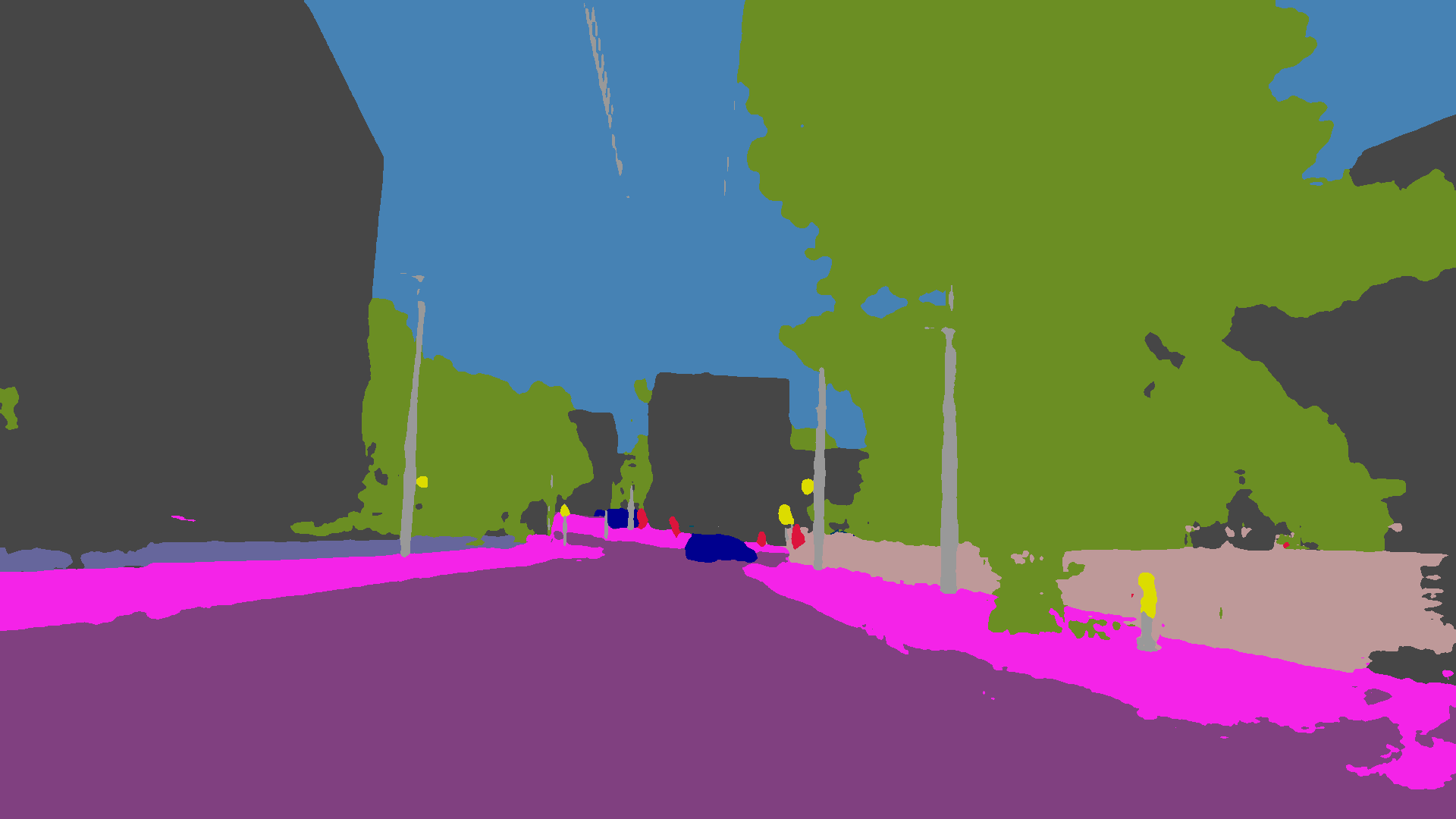}
			& \includegraphics[width=7em, valign=m]{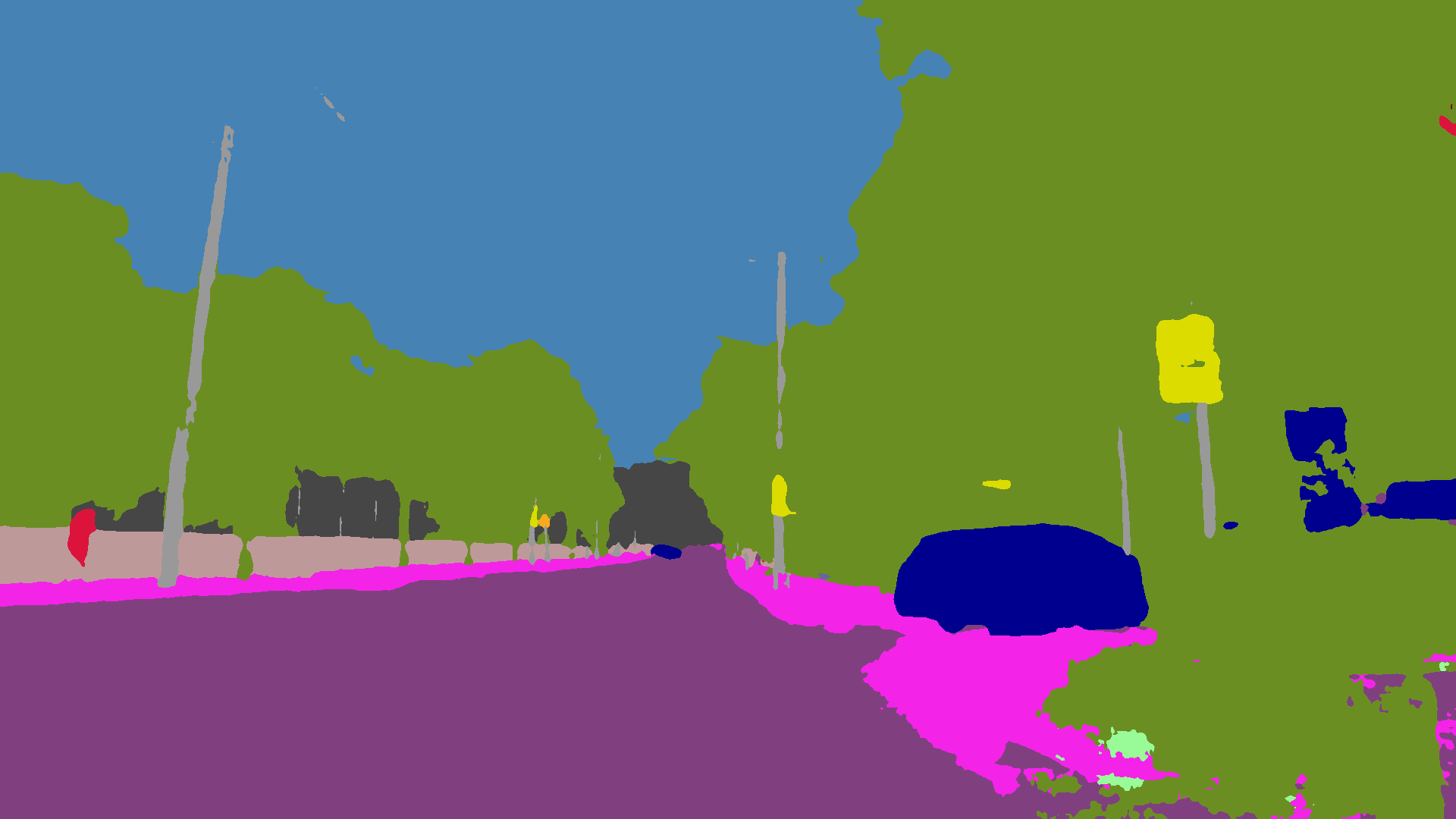}
			& \includegraphics[width=7em, valign=m]{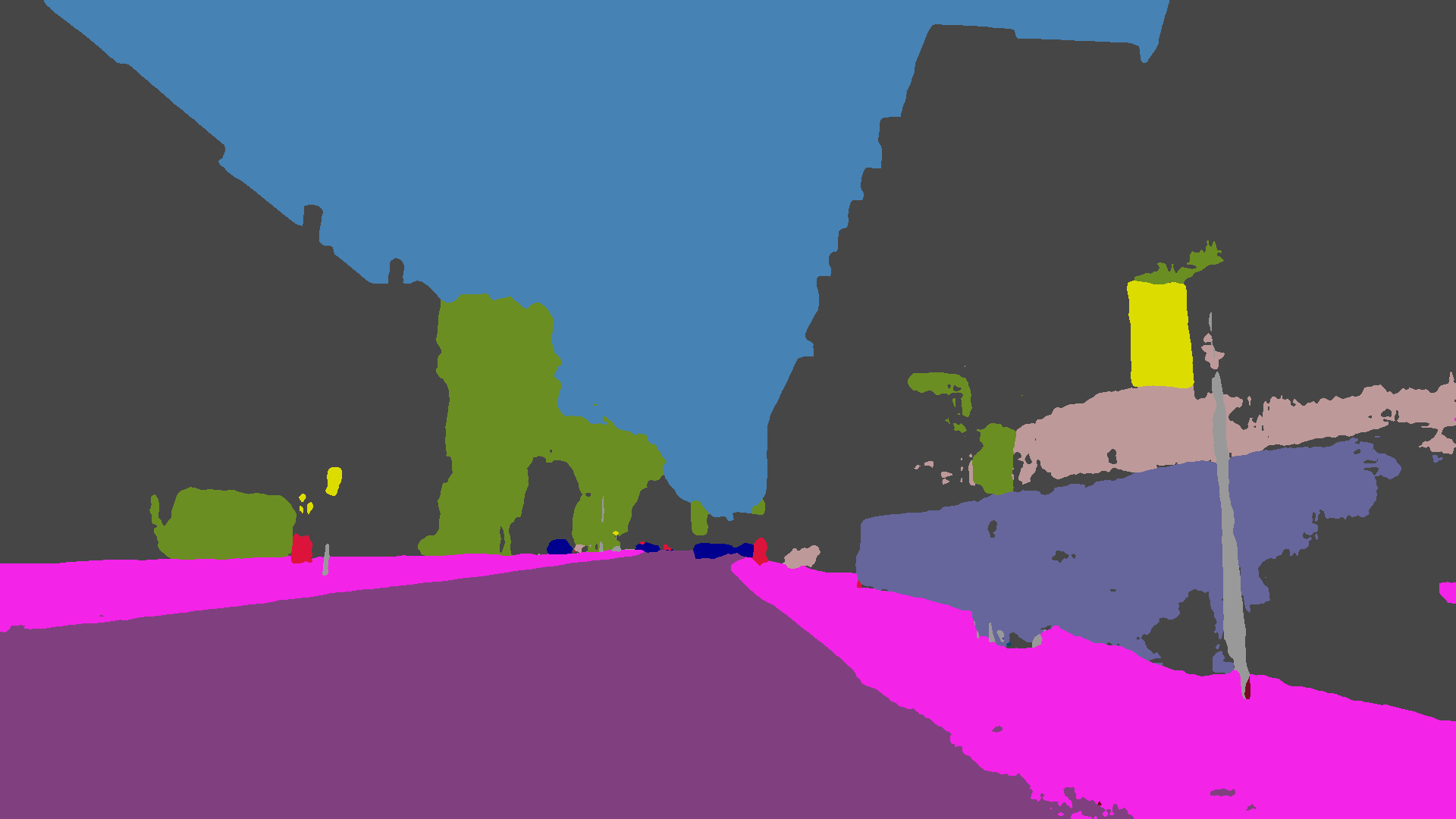}
			& \includegraphics[width=7em, valign=m]{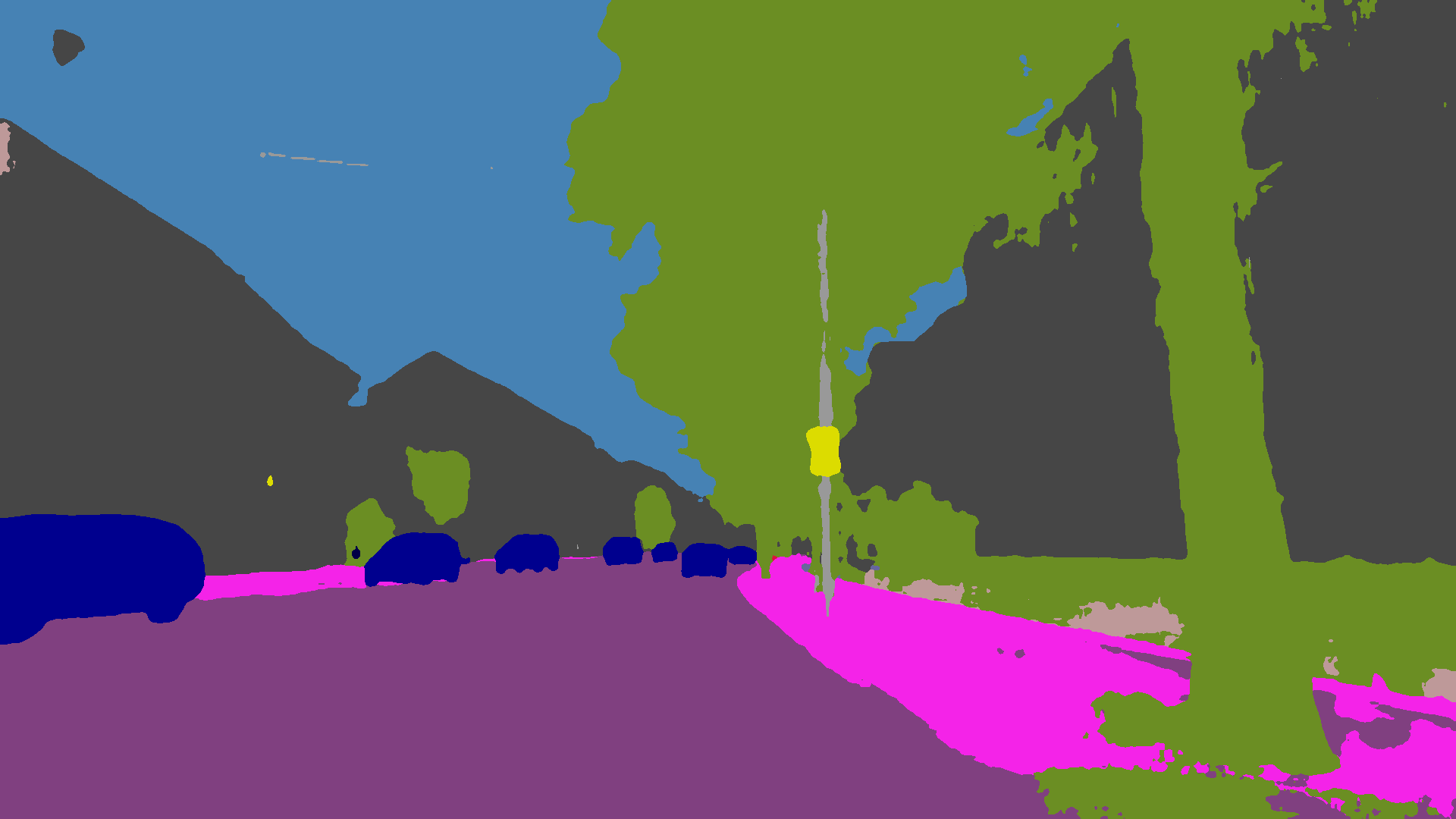}
			& \includegraphics[width=7em, valign=m]{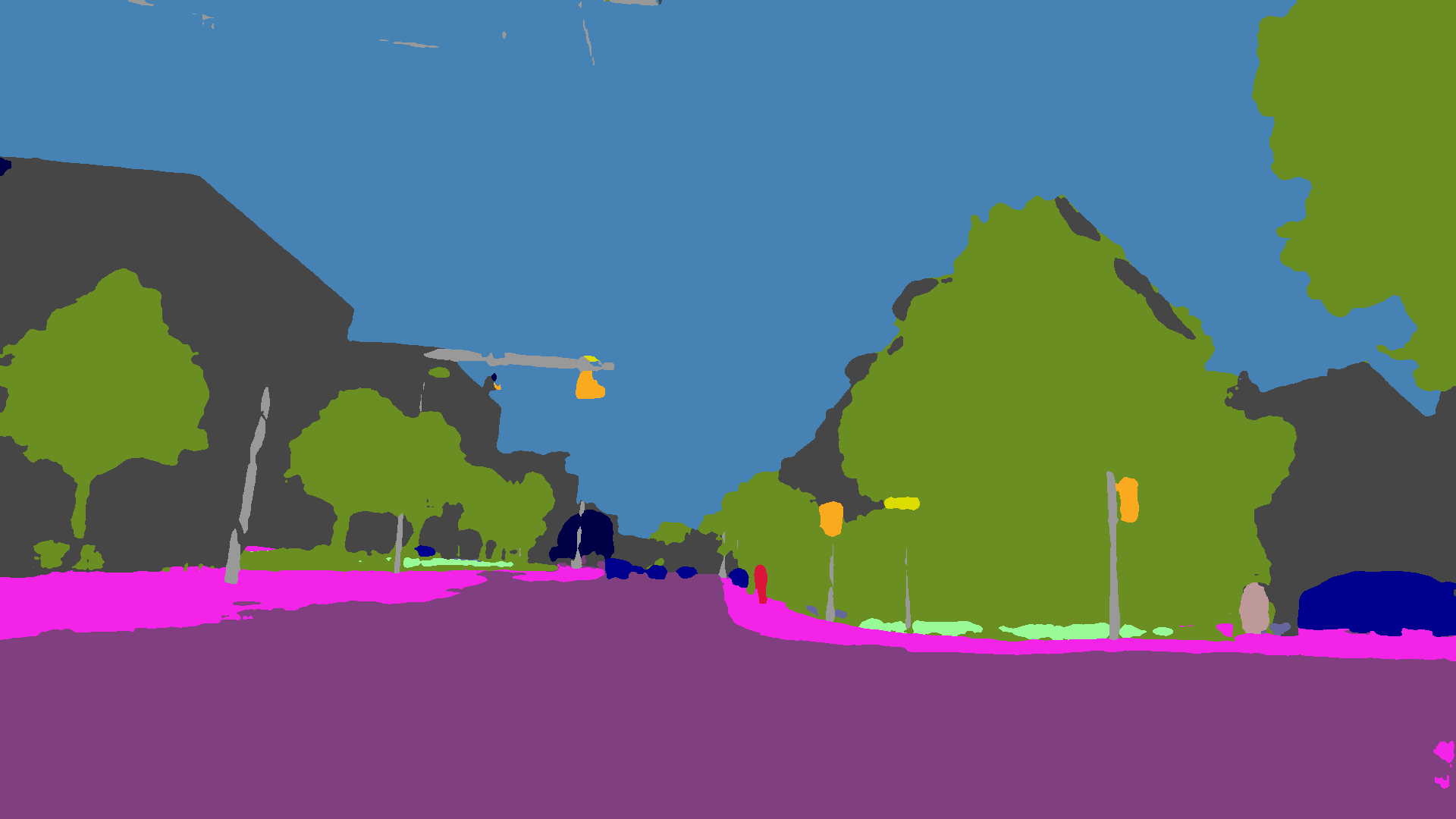}
			& \includegraphics[width=7em, valign=m]{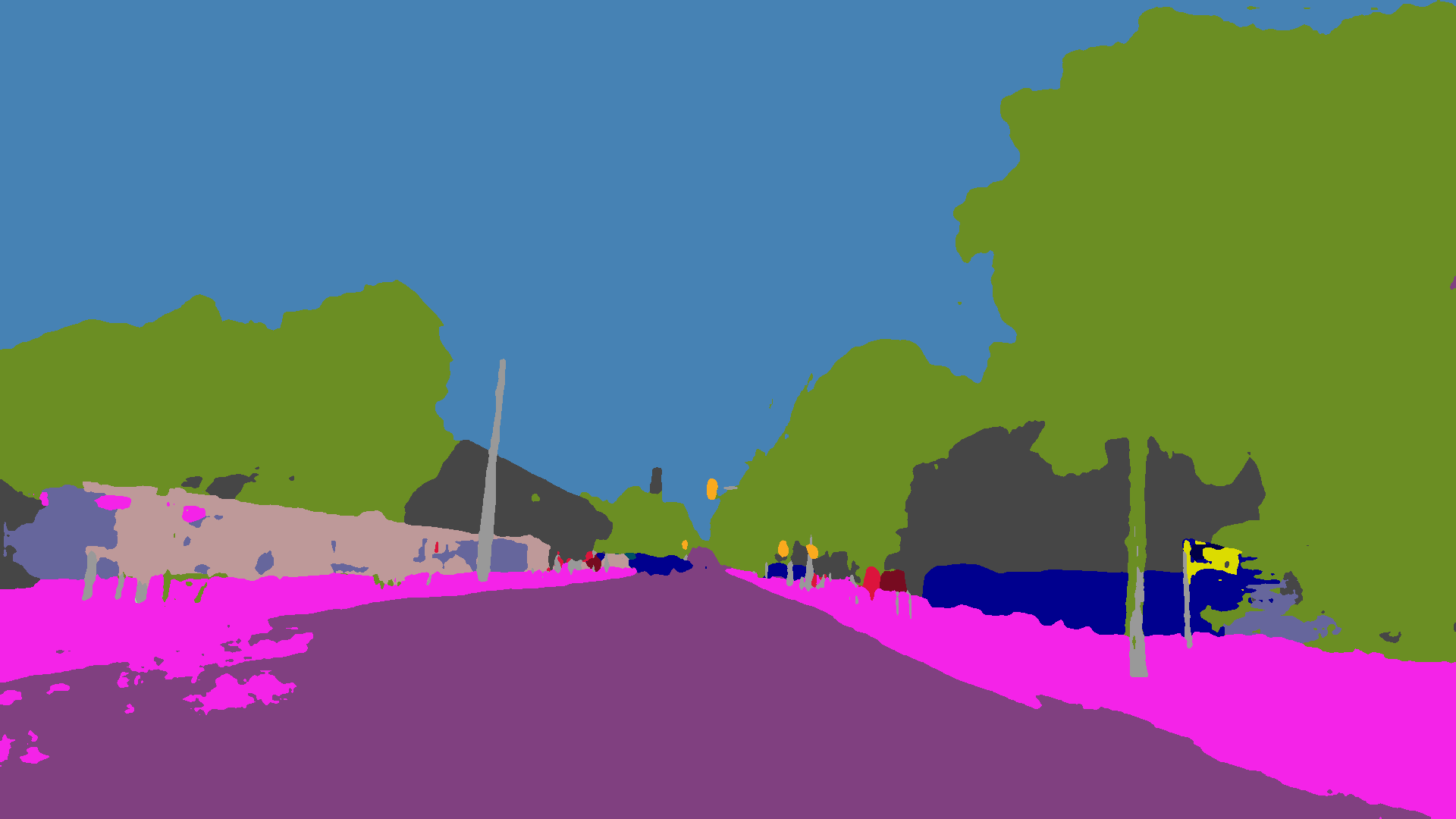}
			& \includegraphics[width=7em, valign=m]{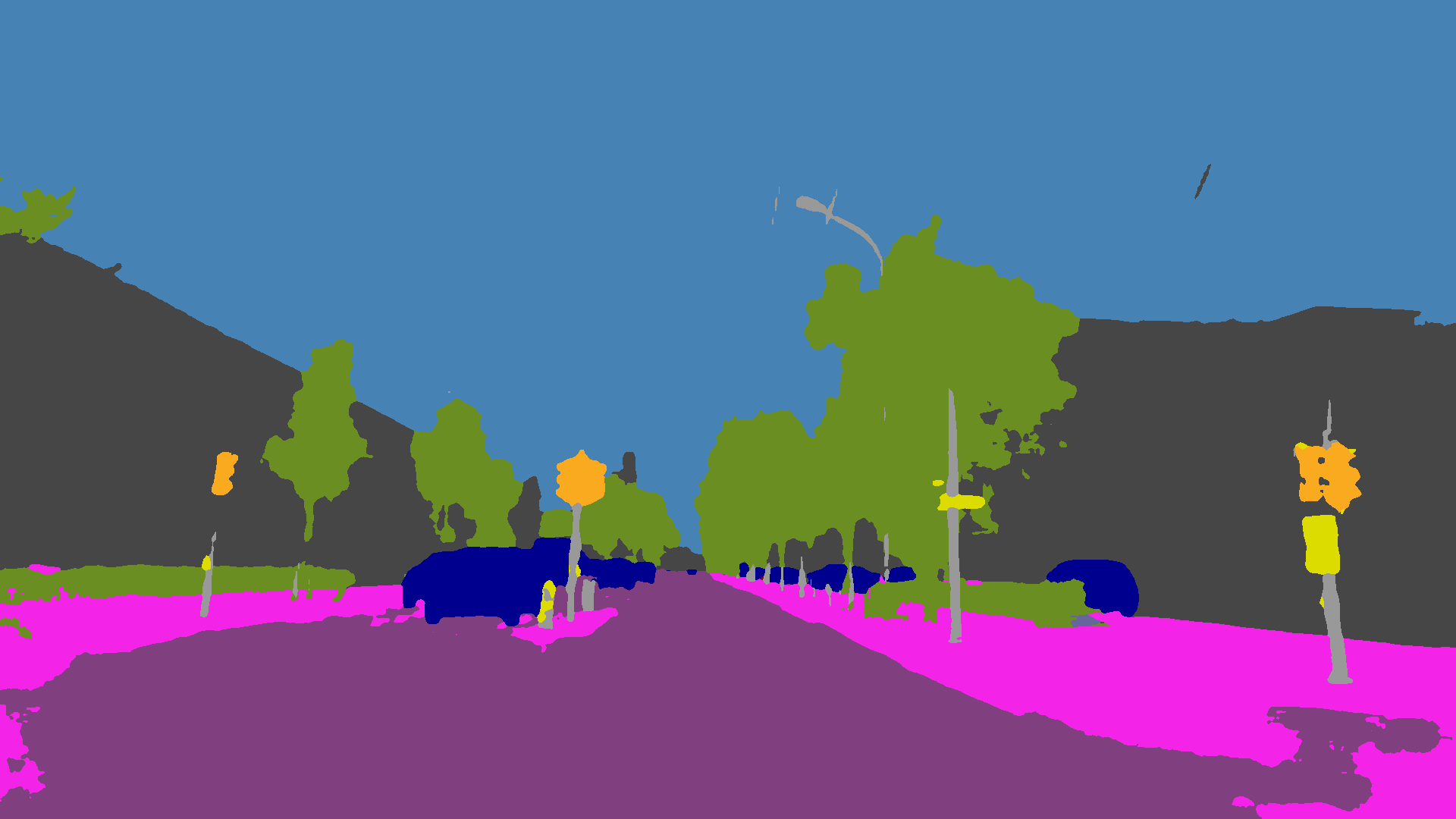}
			& \includegraphics[width=7em, valign=m]{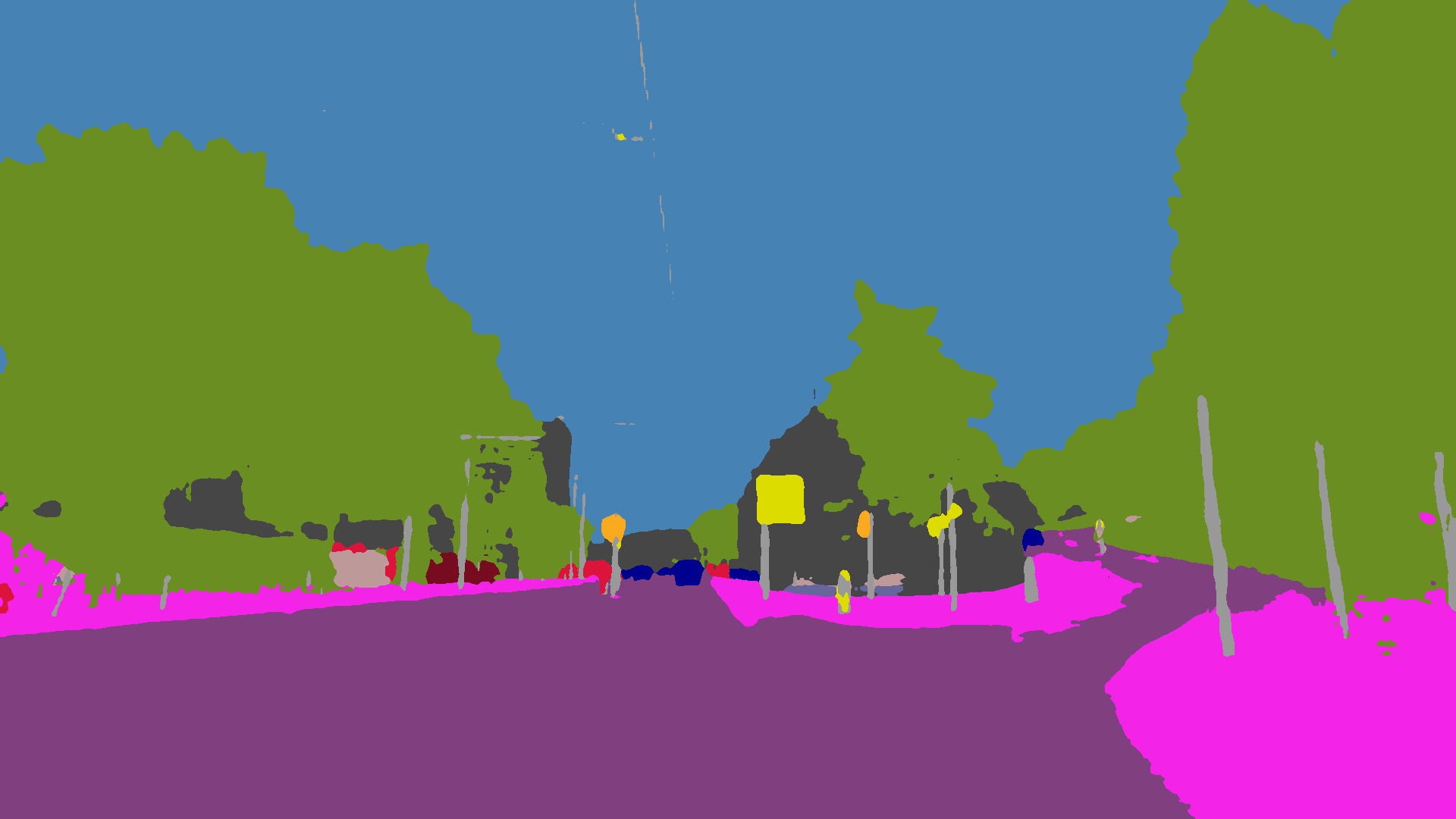}\vspace{0.3em}\\
			
			\midrule
			\adjustbox{valign=m}{\multirow{2}{*}[-0.3em]{\rotatebox{90}{\textbf{PSANet}~\cite{Zhao2018PSANetPS}}}}
			& \adjustbox{valign=m}{{\rotatebox{90}{none}}}
			& \includegraphics[width=7em, valign=m]{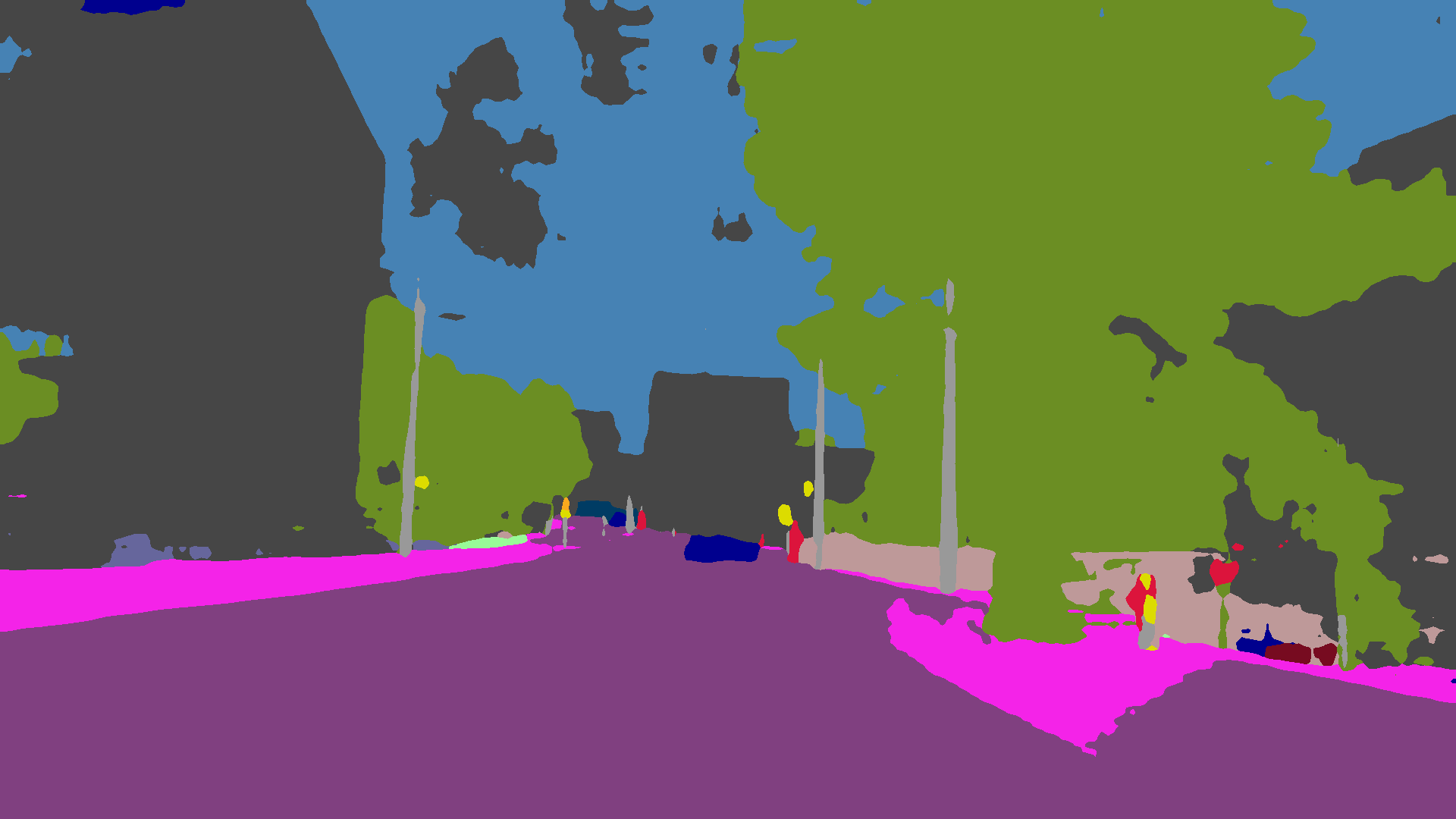}
			& \includegraphics[width=7em, valign=m]{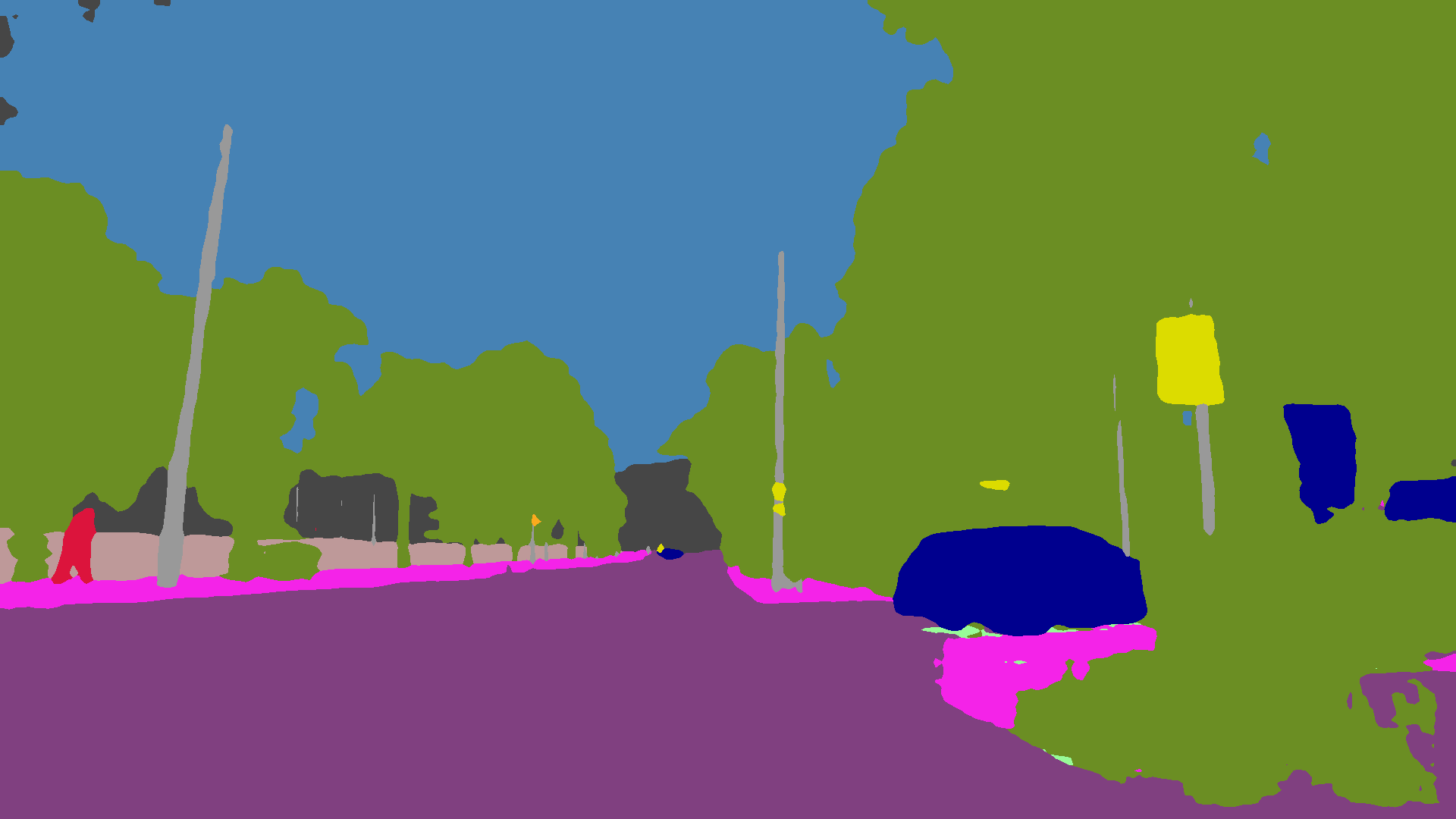}
			& \includegraphics[width=7em, valign=m]{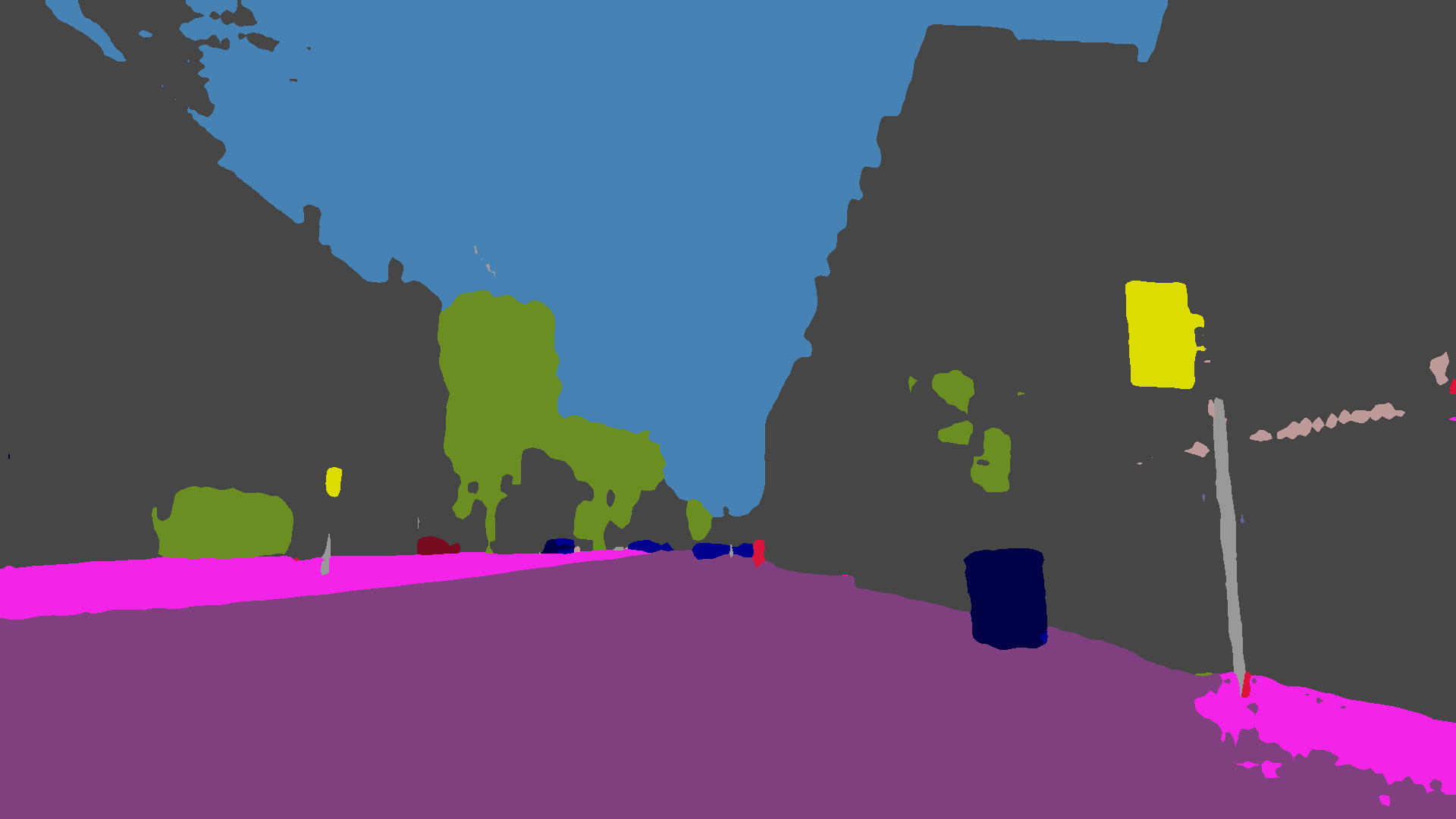}
			& \includegraphics[width=7em, valign=m]{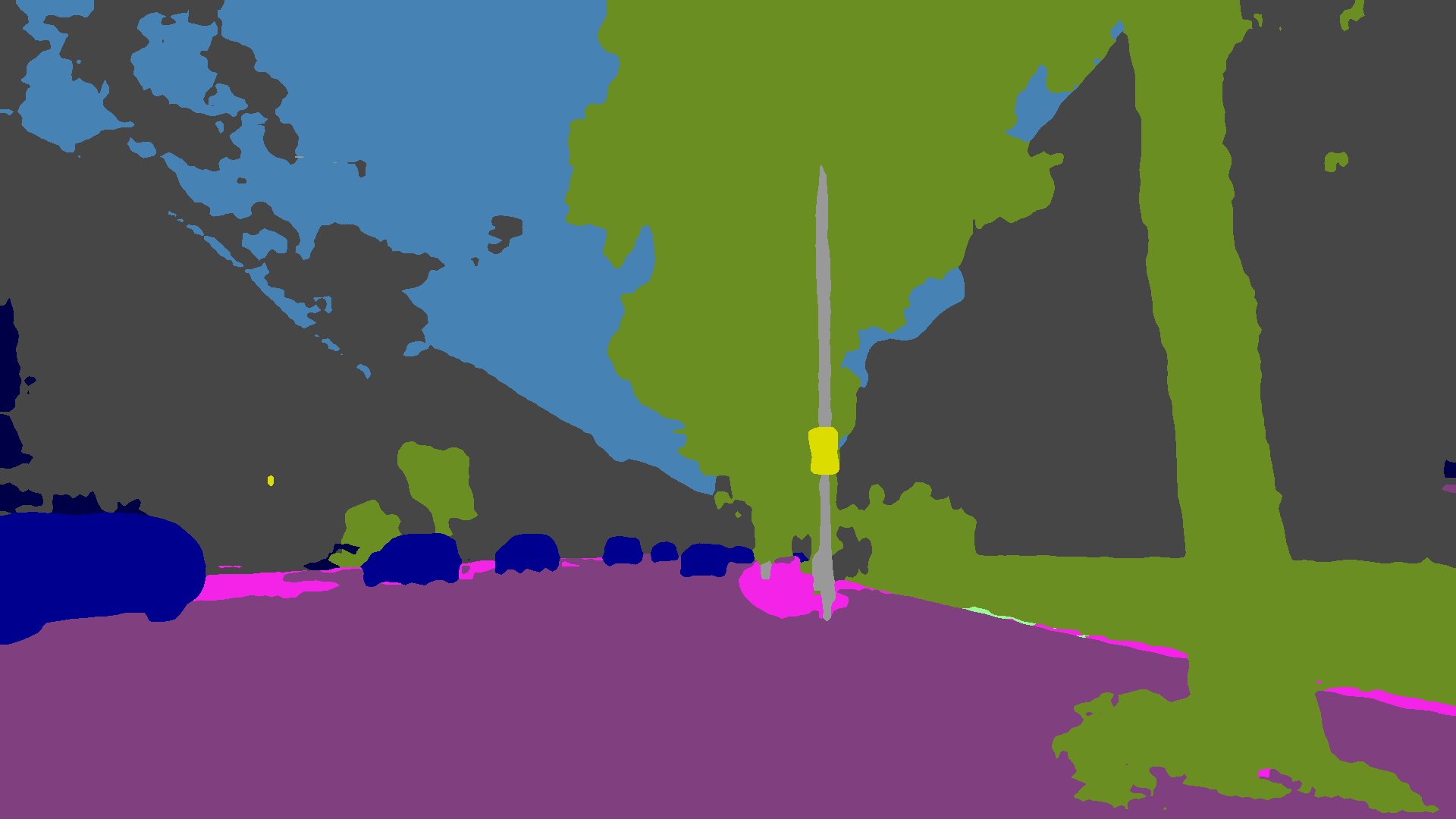}
			& \includegraphics[width=7em, valign=m]{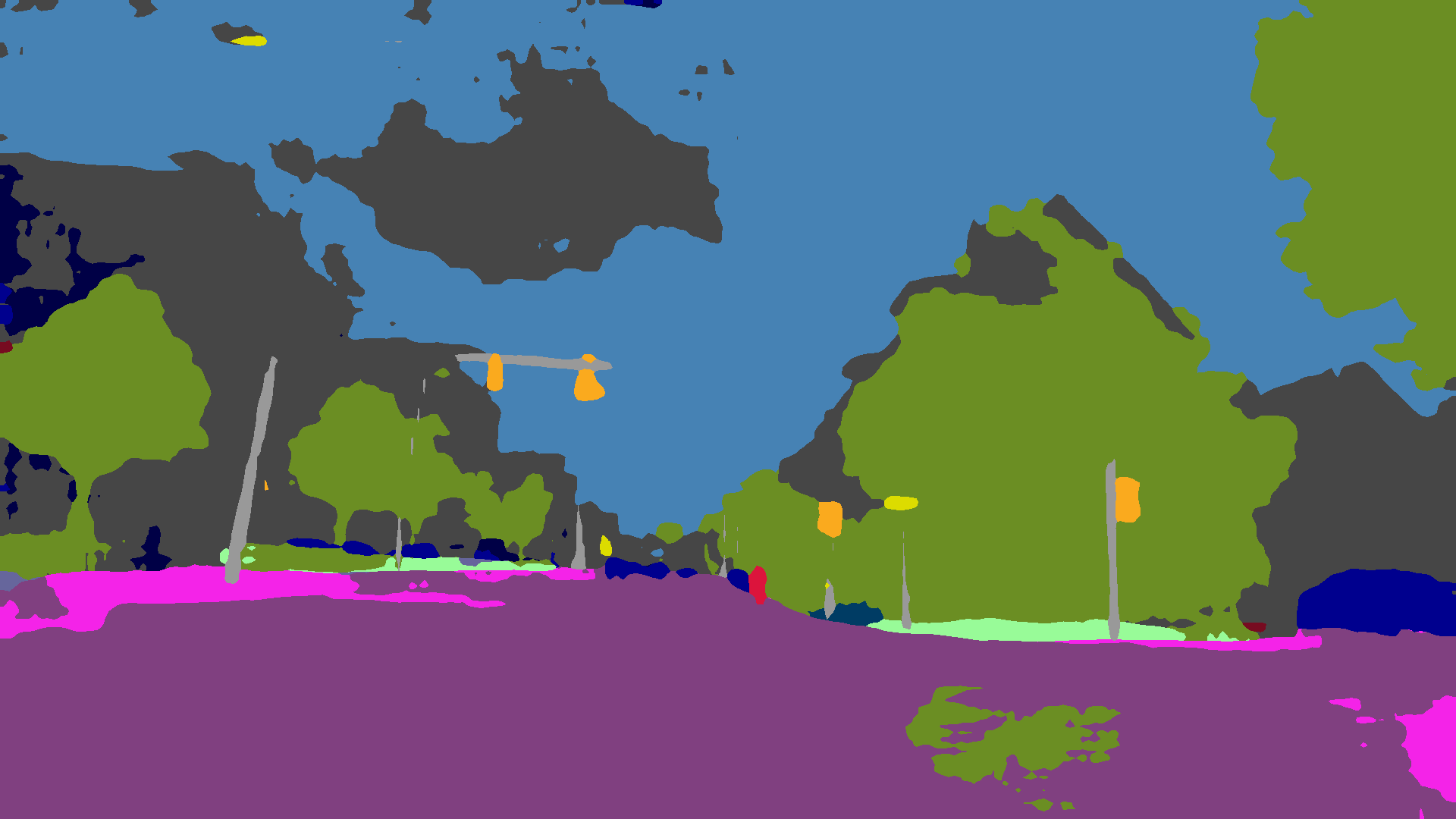}
			& \includegraphics[width=7em, valign=m]{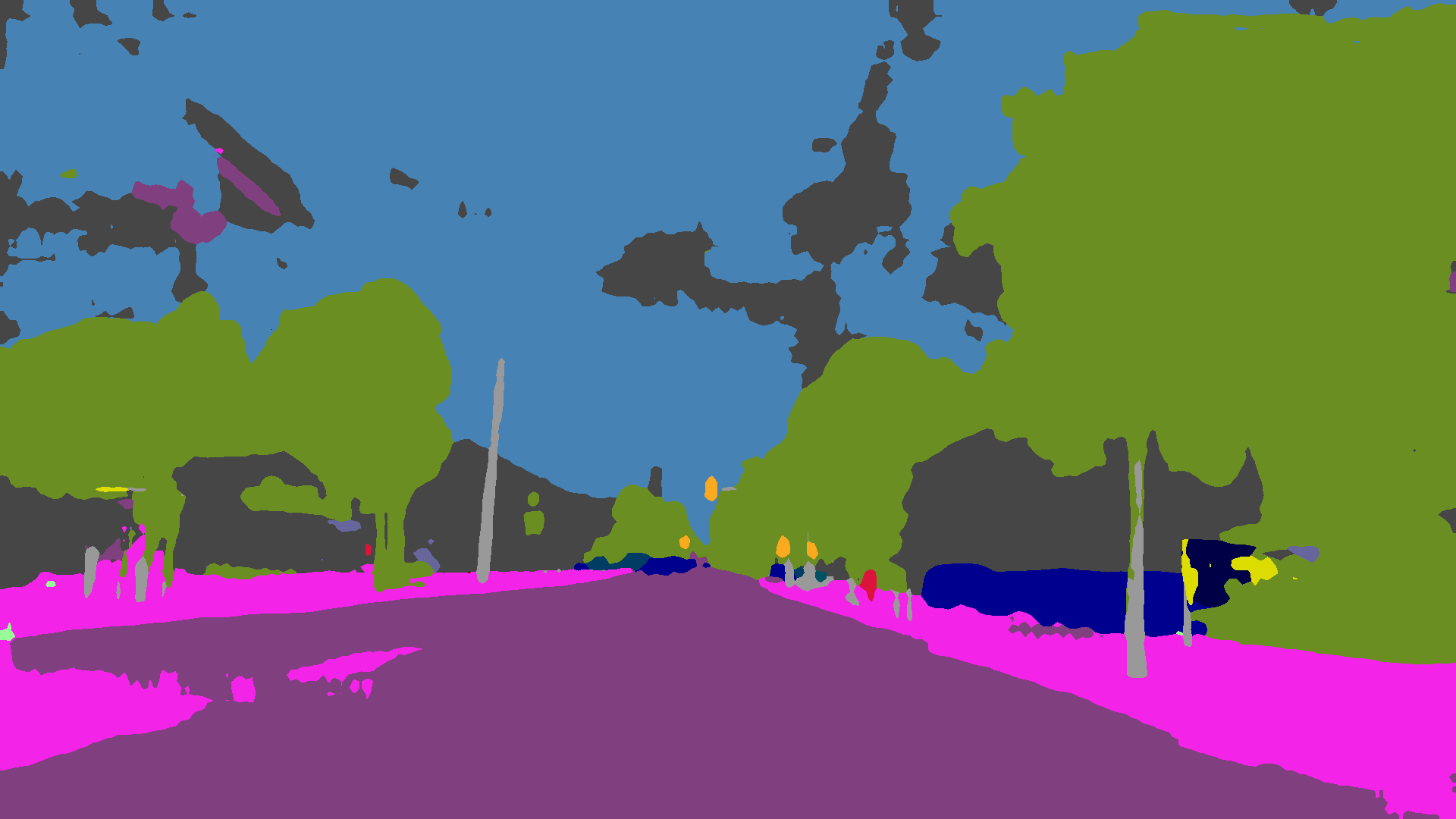}
			& \includegraphics[width=7em, valign=m]{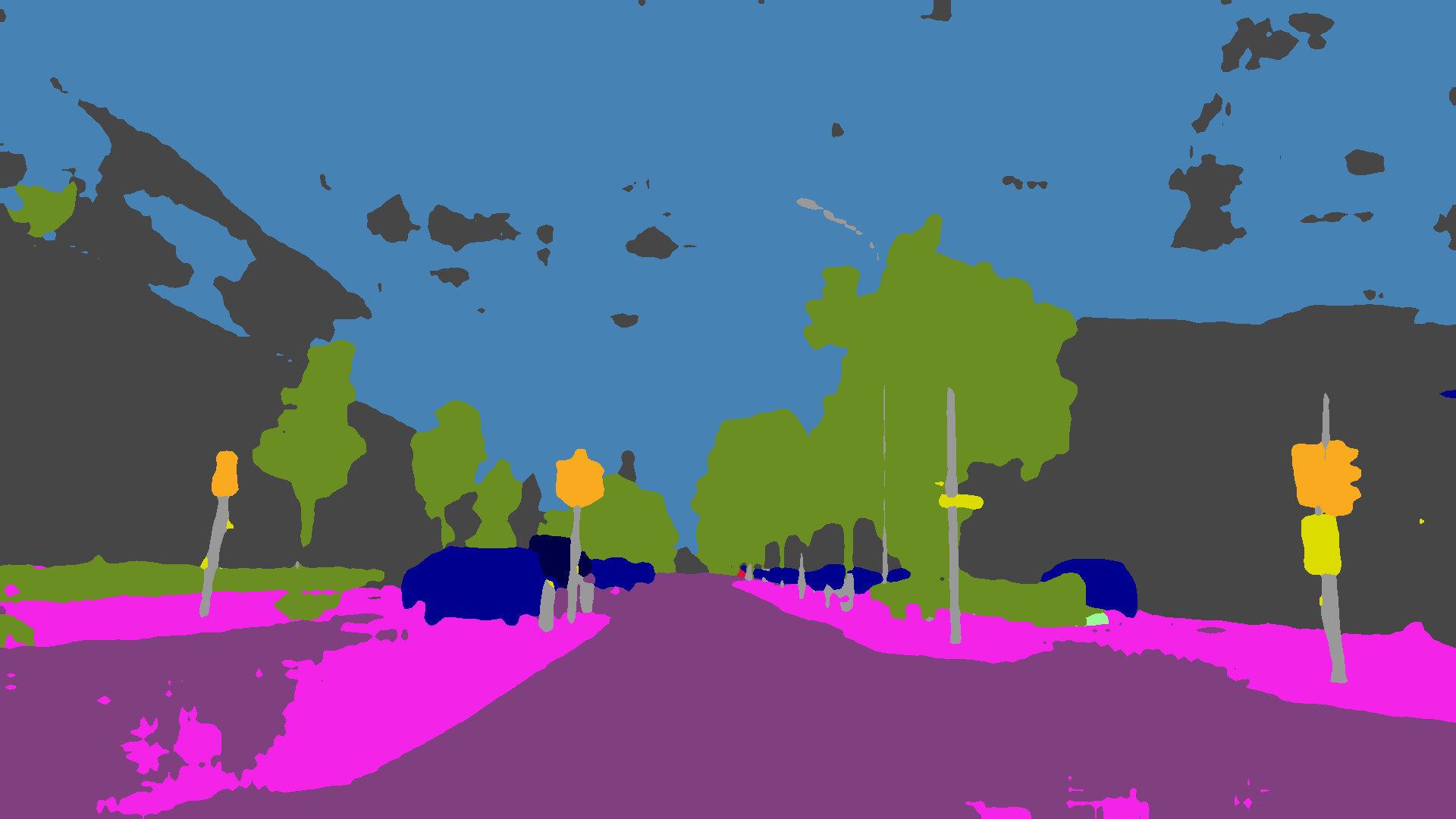}
			& \includegraphics[width=7em, valign=m]{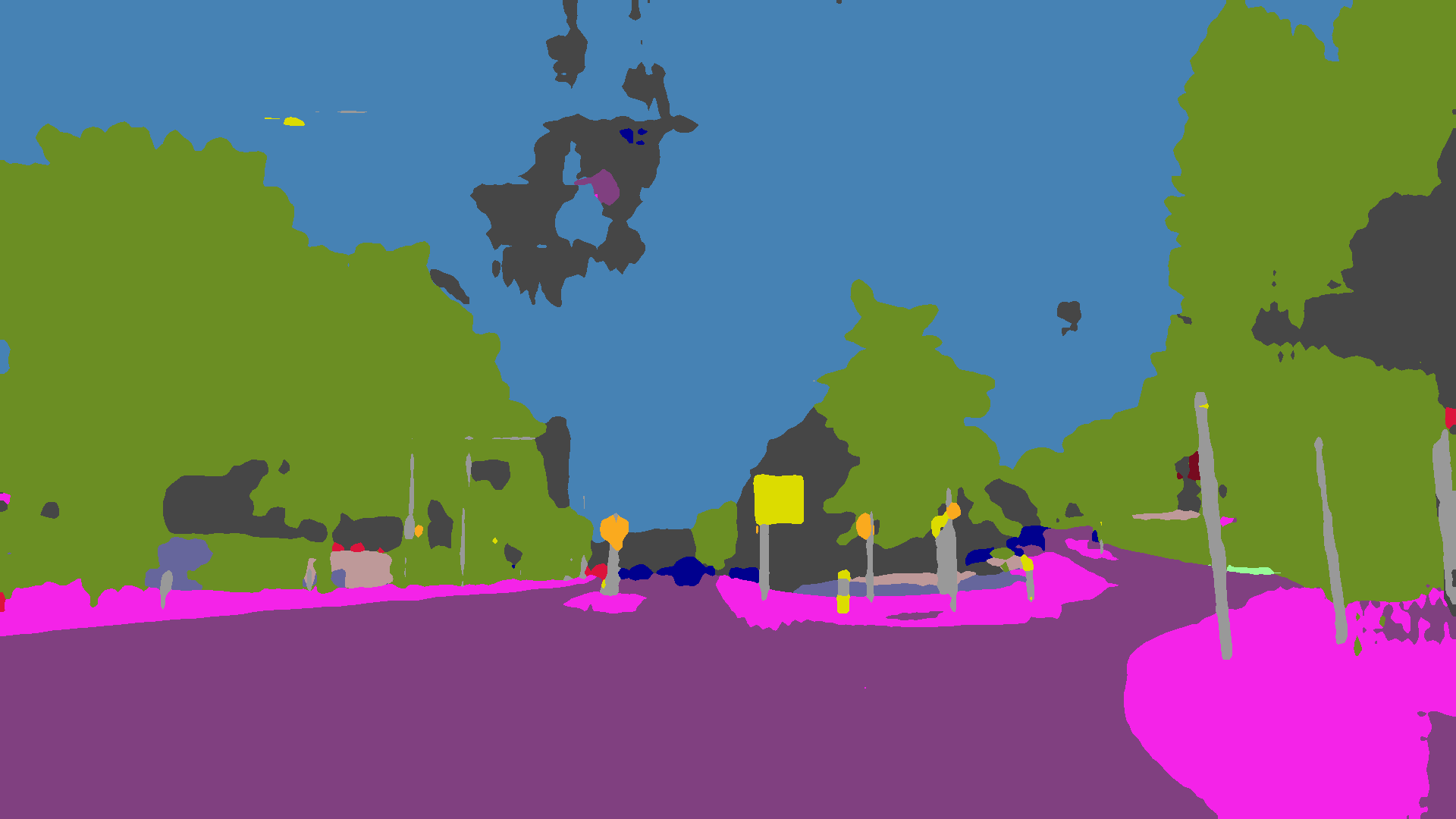}\vspace{0.3em}\\
			
			& \adjustbox{valign=m}{{\rotatebox{90}{Ours}}}
			& \includegraphics[width=7em, valign=m]{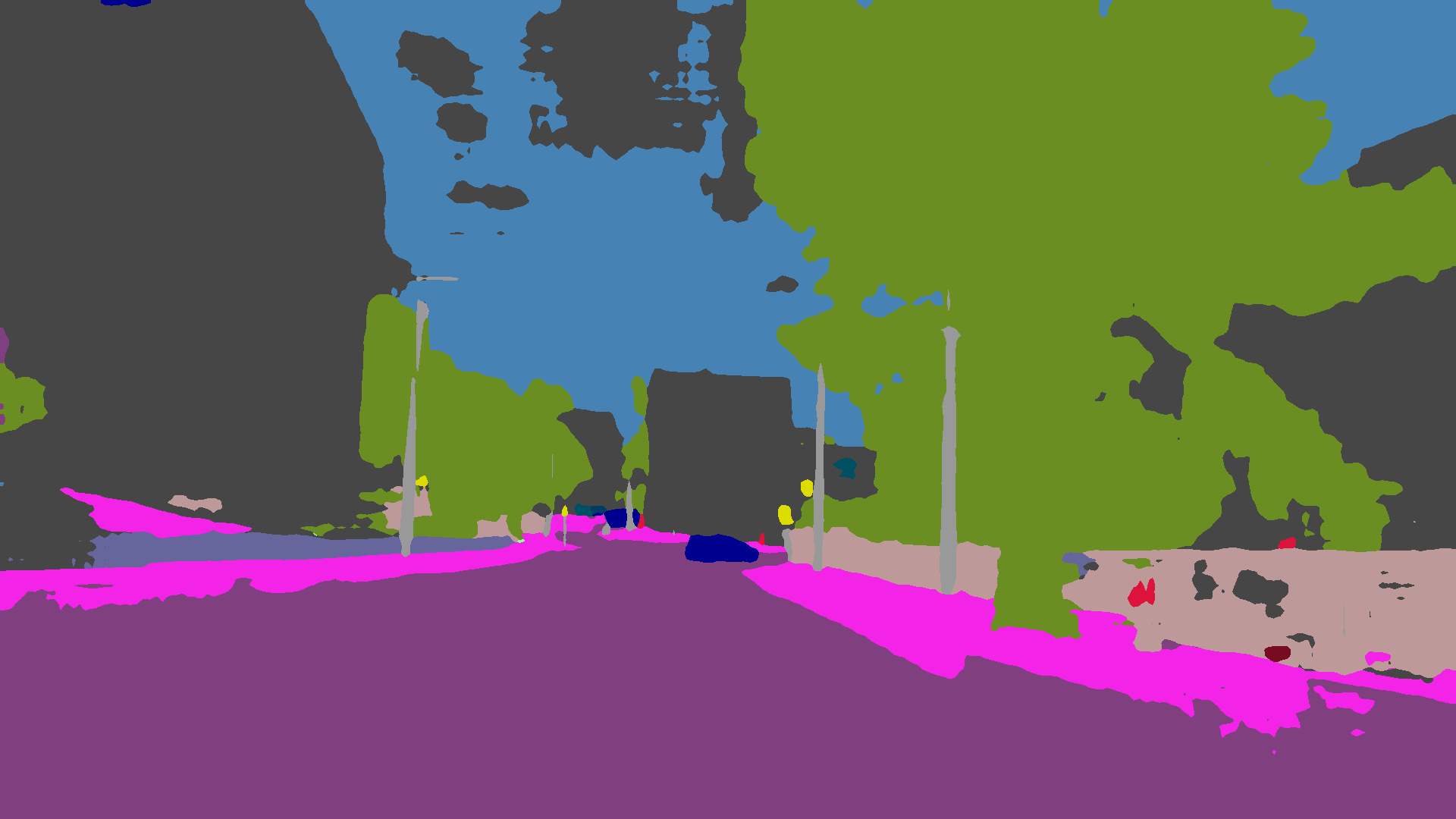}
			& \includegraphics[width=7em, valign=m]{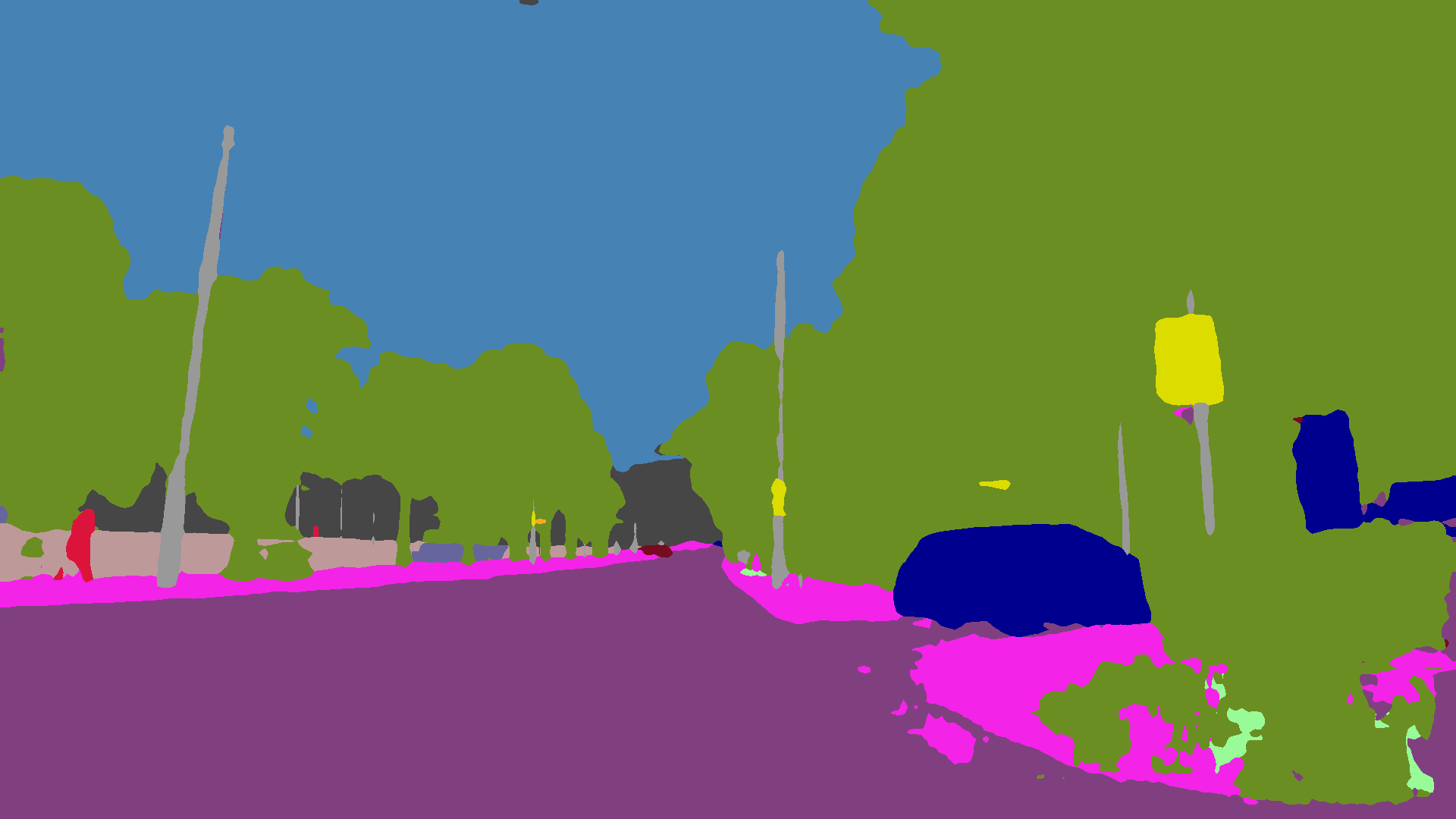}
			& \includegraphics[width=7em, valign=m]{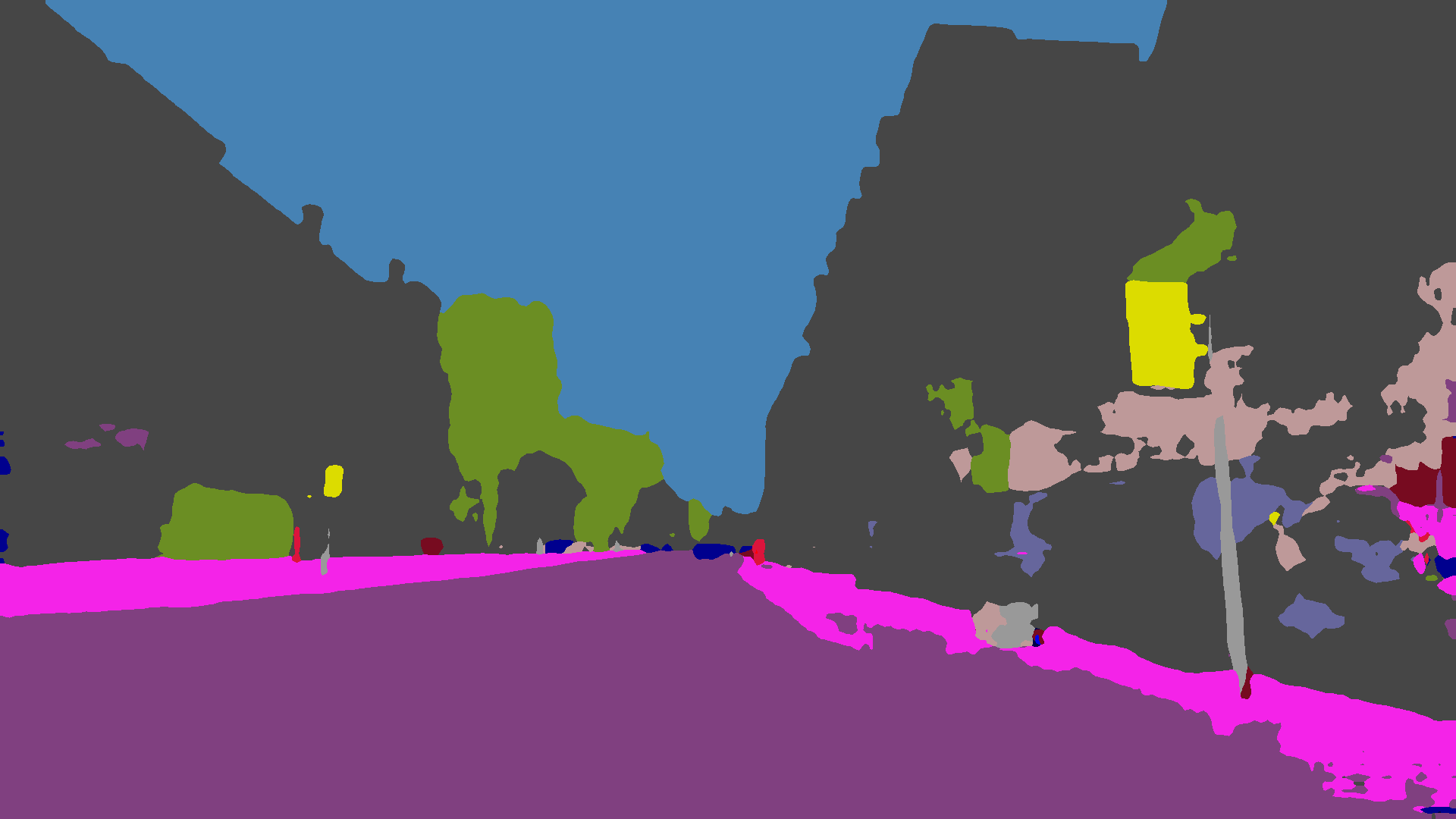}
			& \includegraphics[width=7em, valign=m]{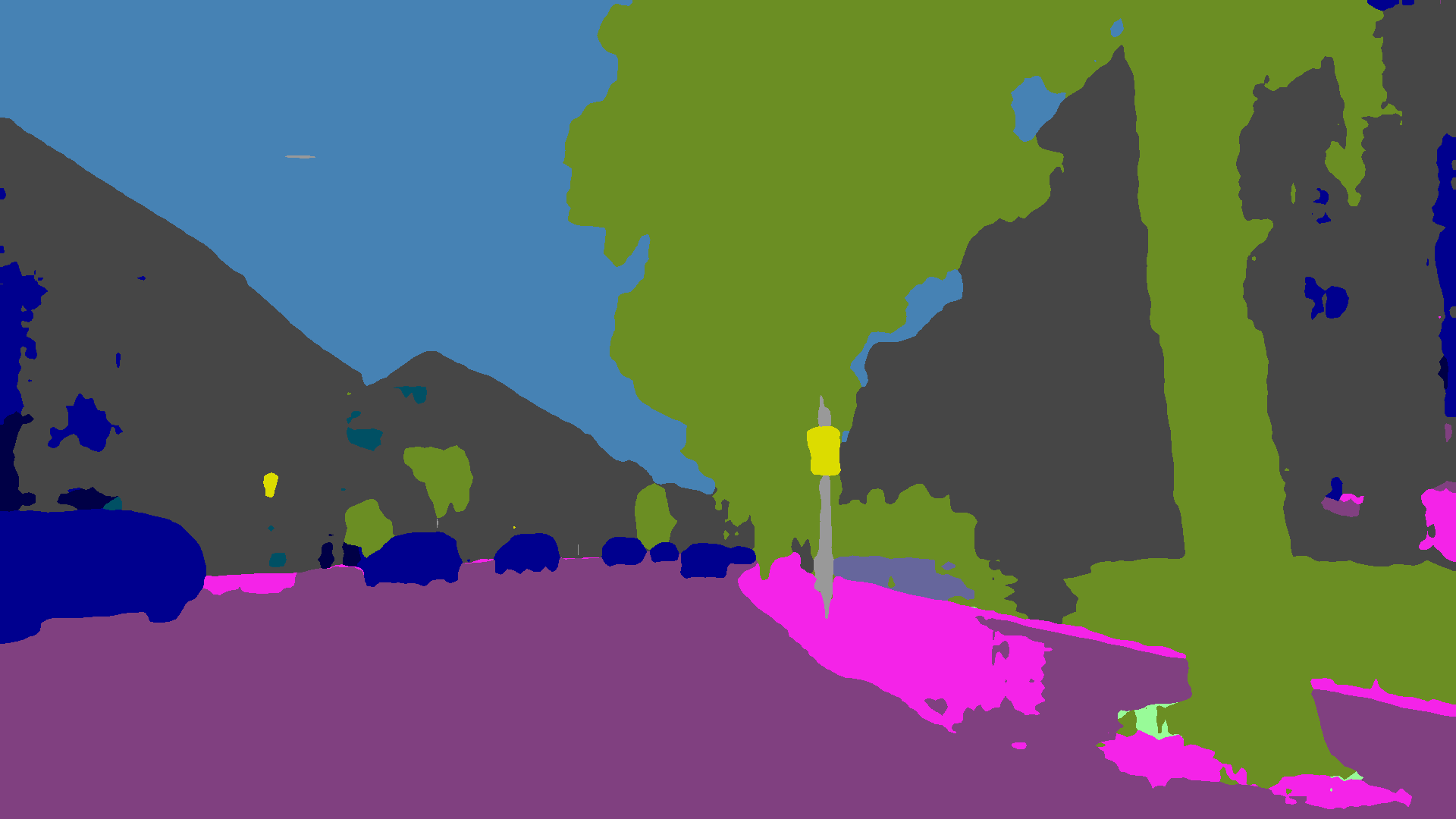}
			& \includegraphics[width=7em, valign=m]{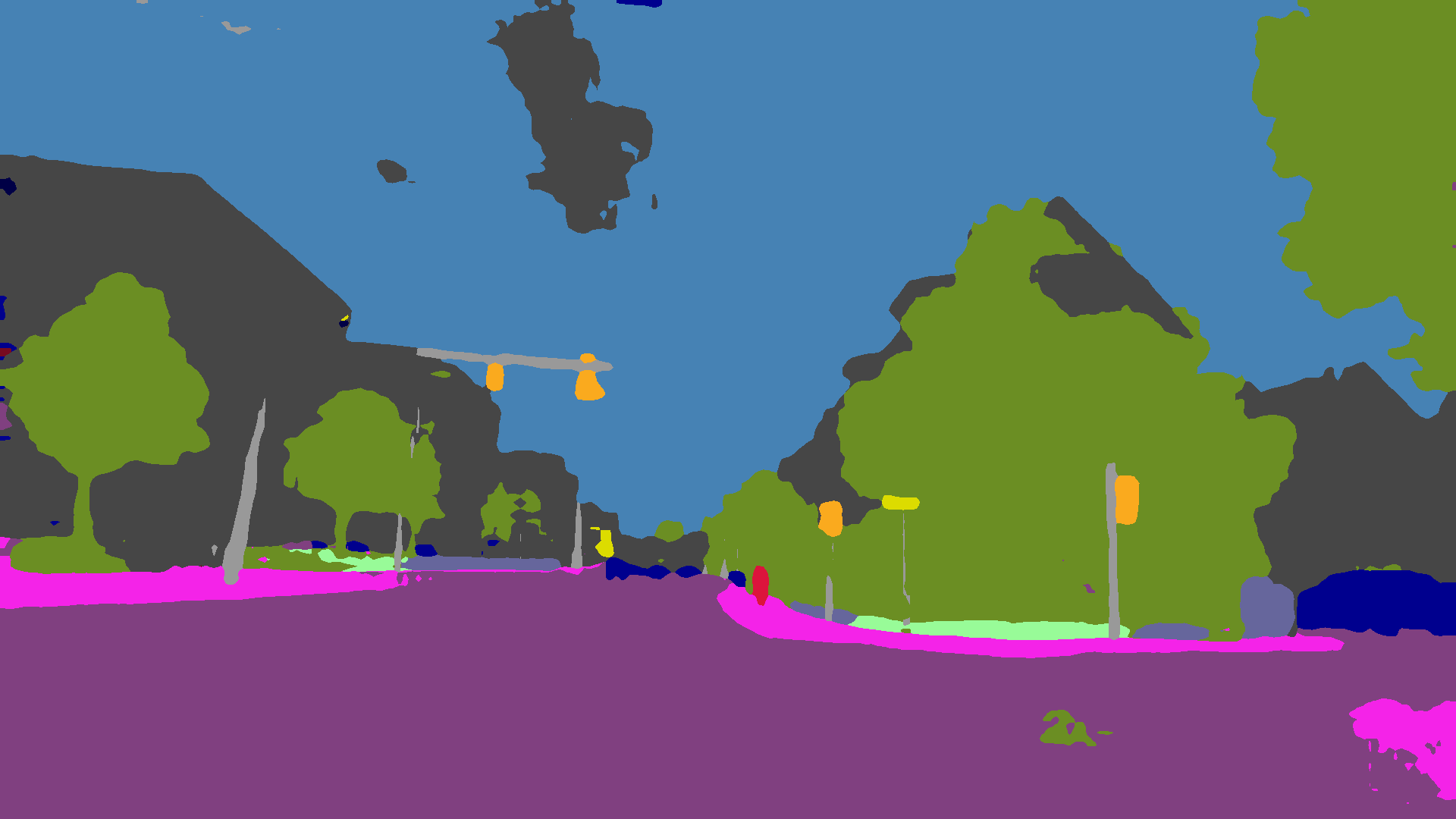}
			& \includegraphics[width=7em, valign=m]{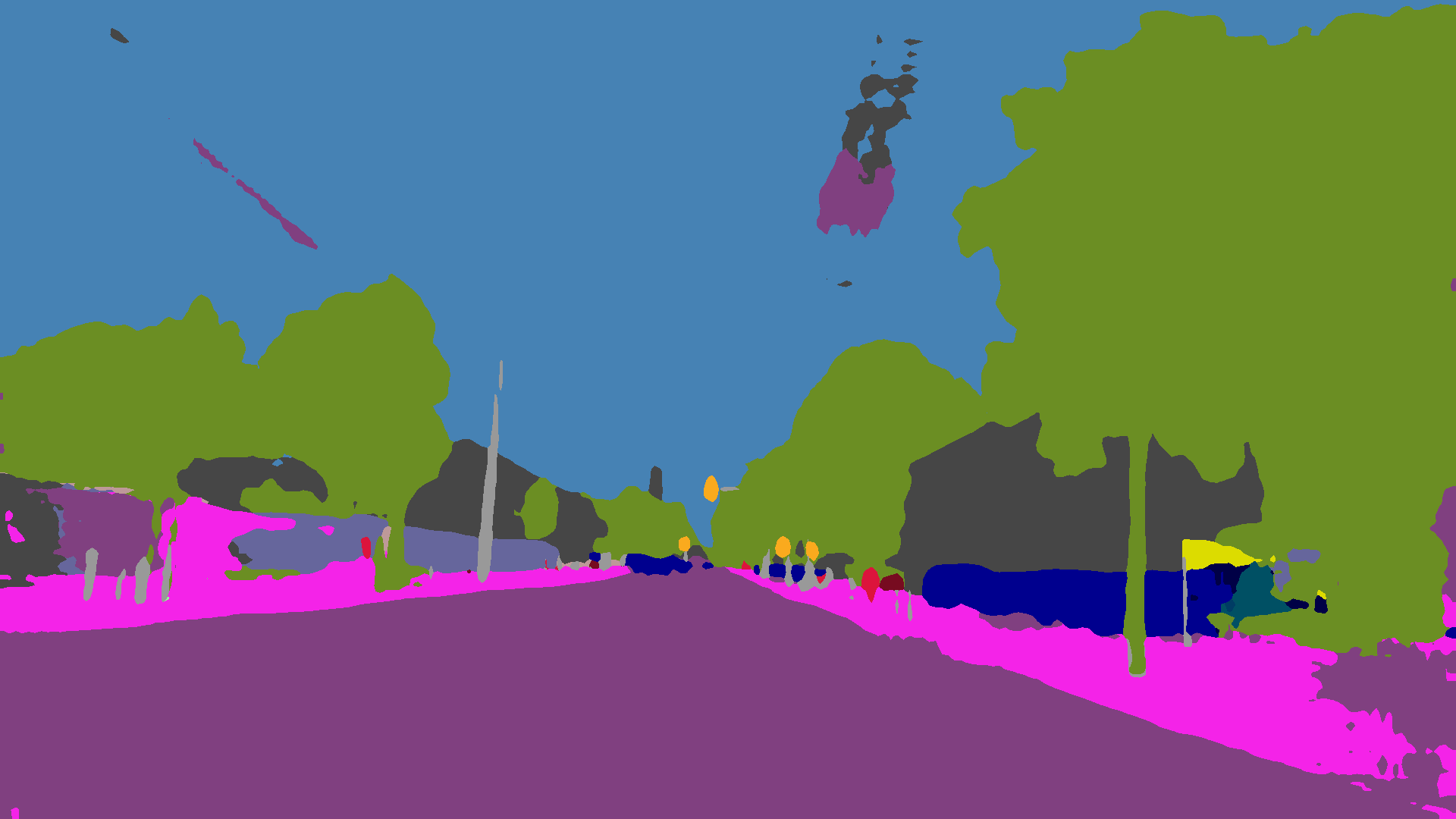}
			& \includegraphics[width=7em, valign=m]{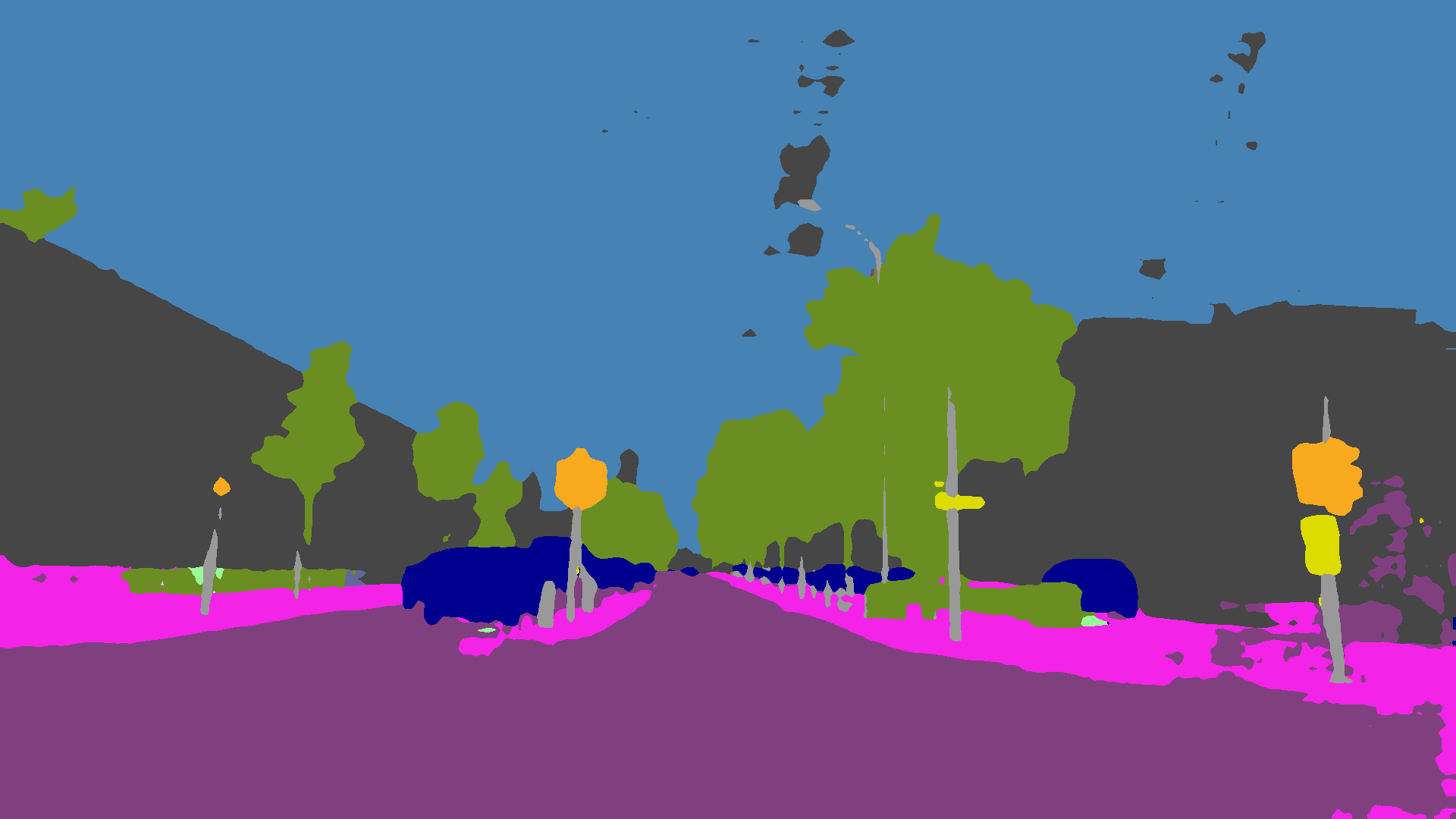}
			& \includegraphics[width=7em, valign=m]{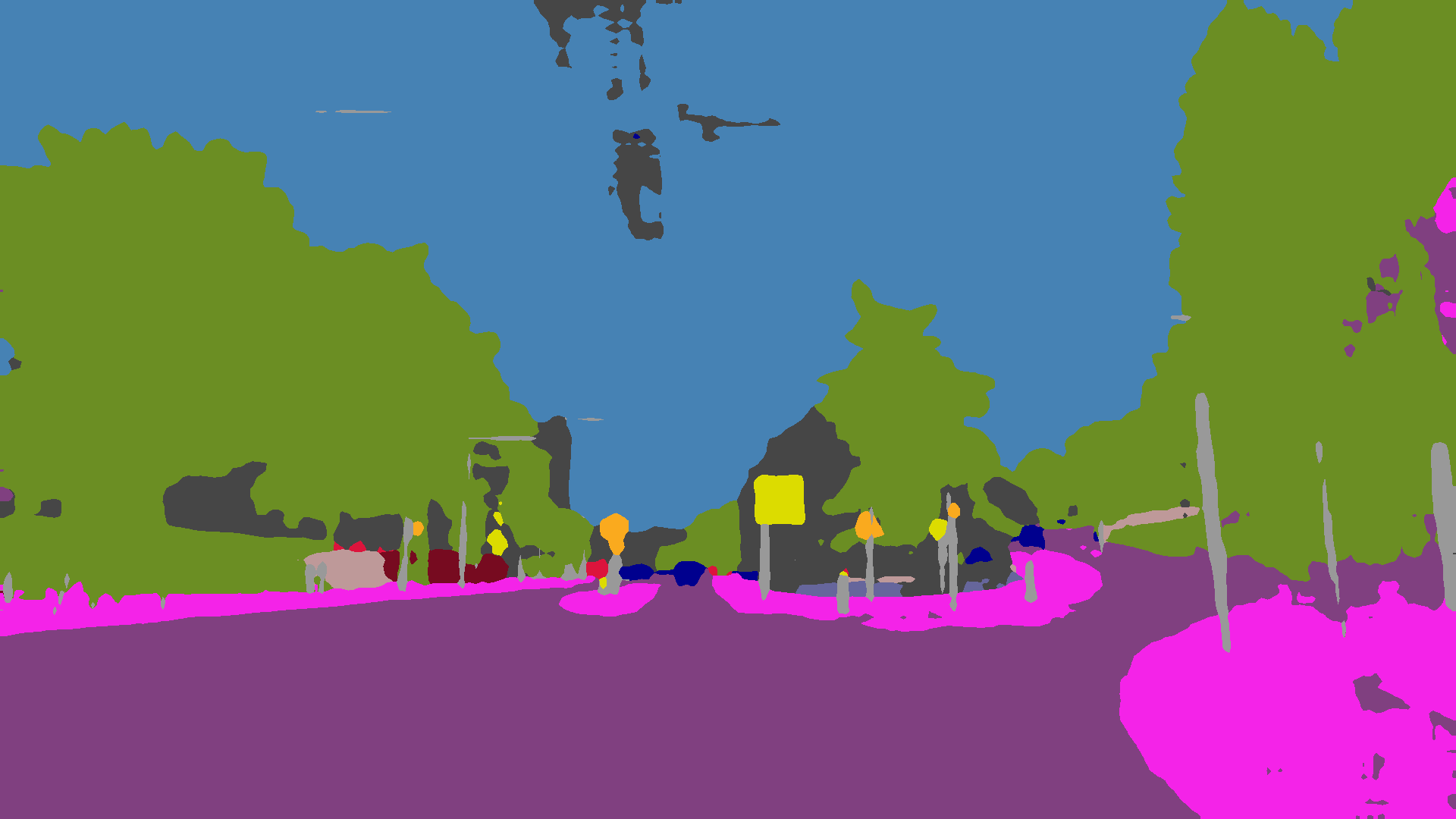}\vspace{0.3em}\\
			
			\midrule
			\adjustbox{valign=m}{\multirow{2}{*}{\rotatebox{90}{\textbf{OCRNet}~\cite{YuanCW20}}}}
			& \adjustbox{valign=m}{{\rotatebox{90}{none}}}
			& \includegraphics[width=7em, valign=m]{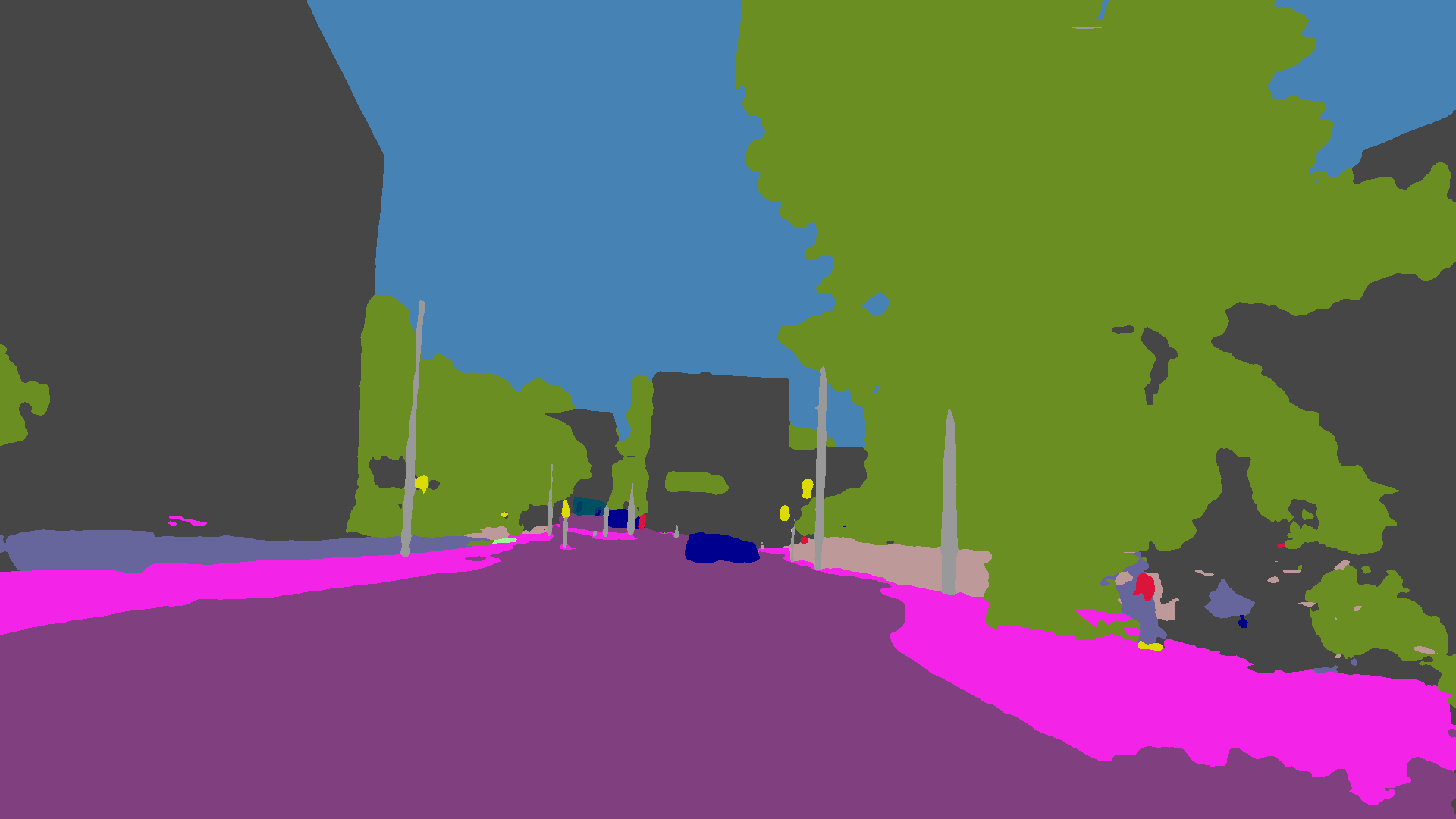}
			& \includegraphics[width=7em, valign=m]{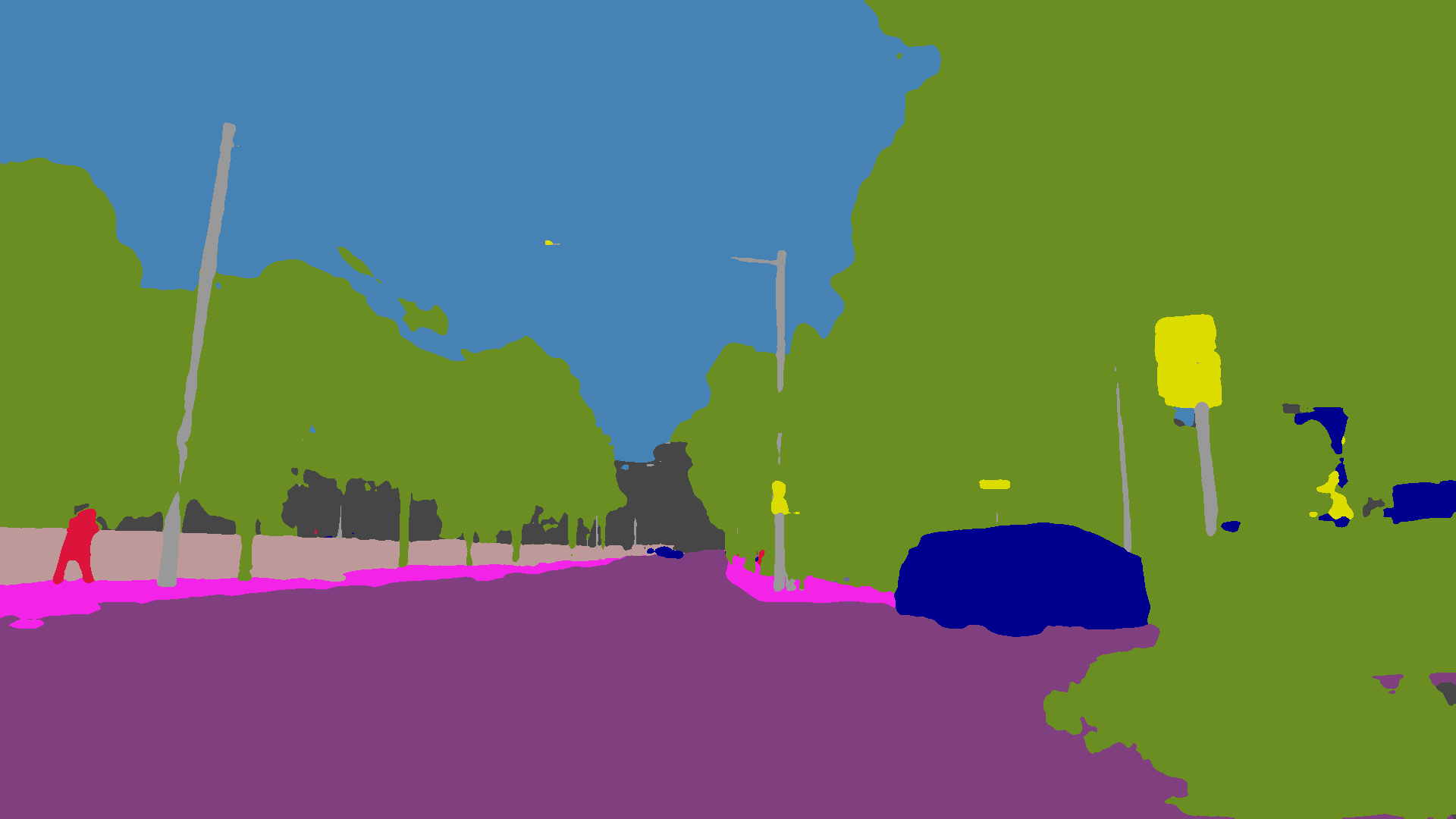}
			& \includegraphics[width=7em, valign=m]{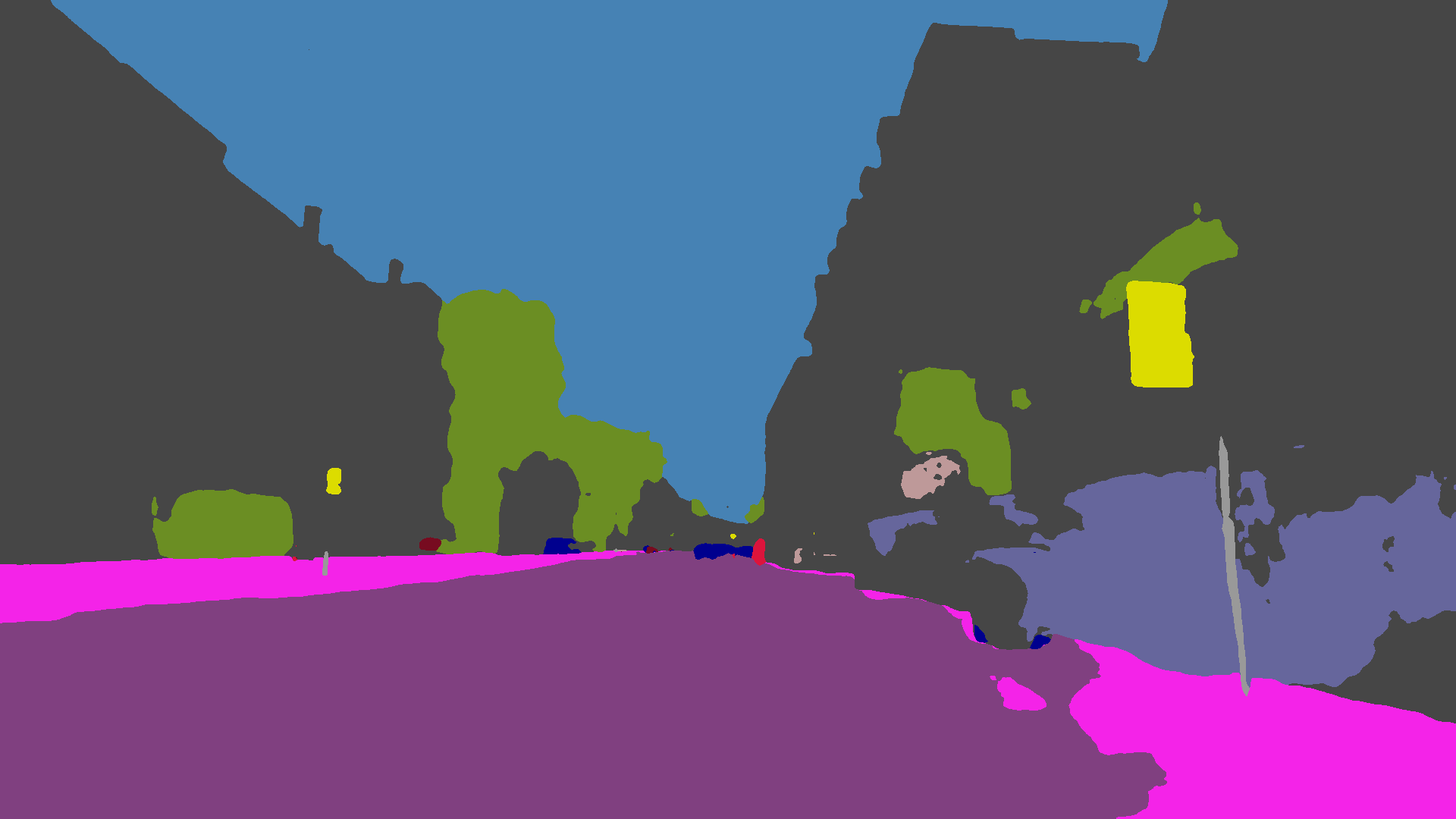}
			& \includegraphics[width=7em, valign=m]{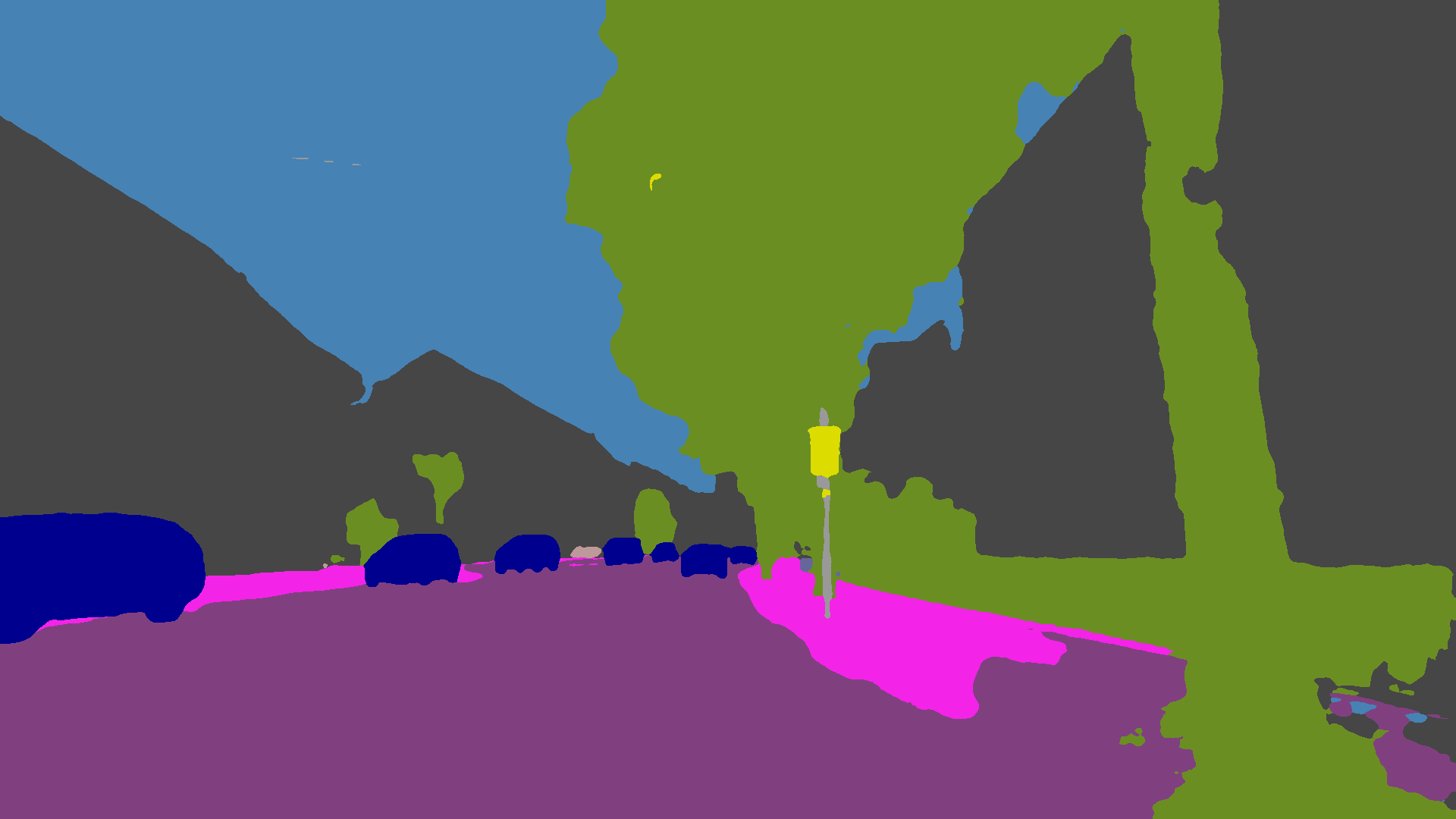}
			& \includegraphics[width=7em, valign=m]{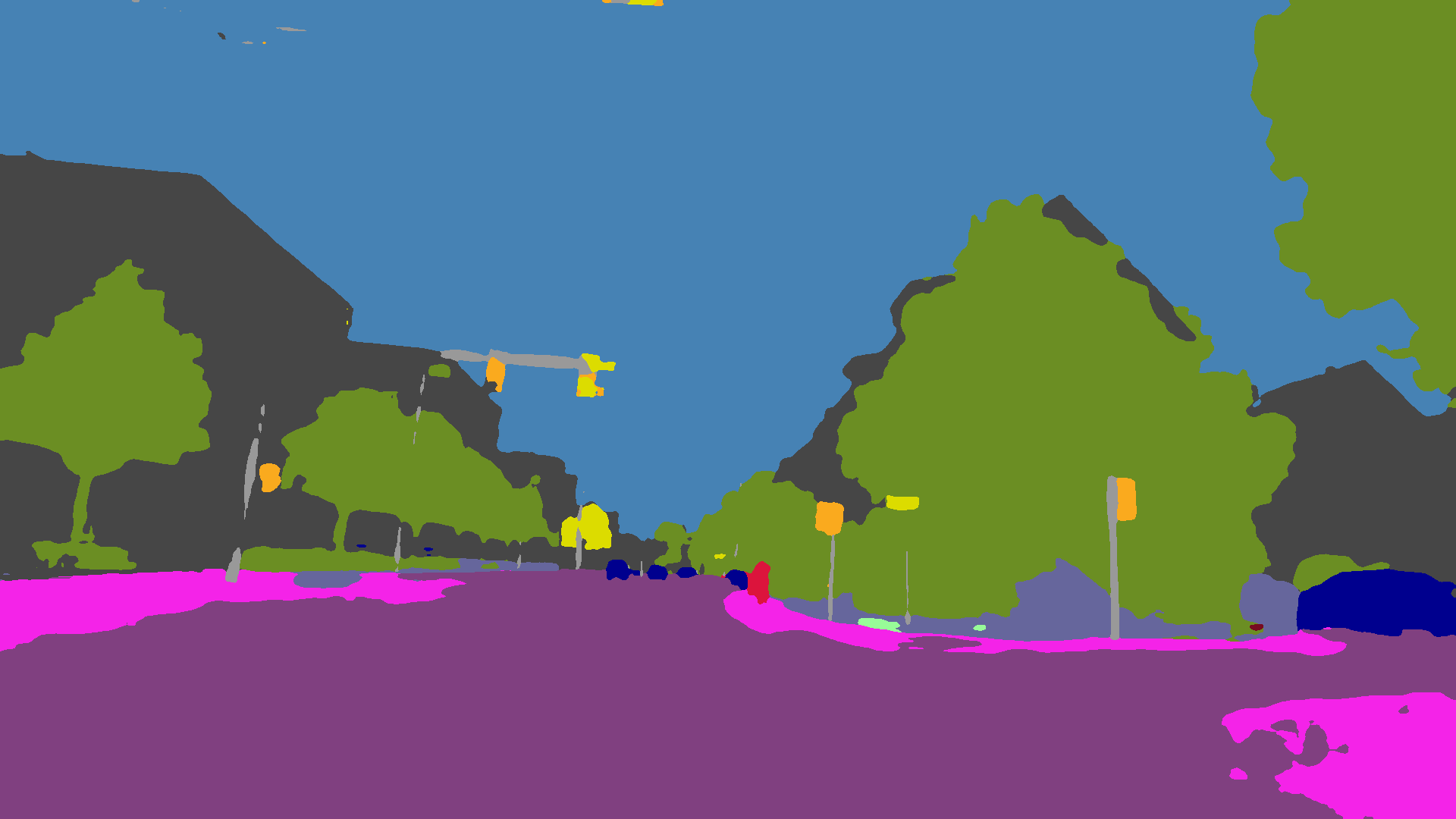}
			& \includegraphics[width=7em, valign=m]{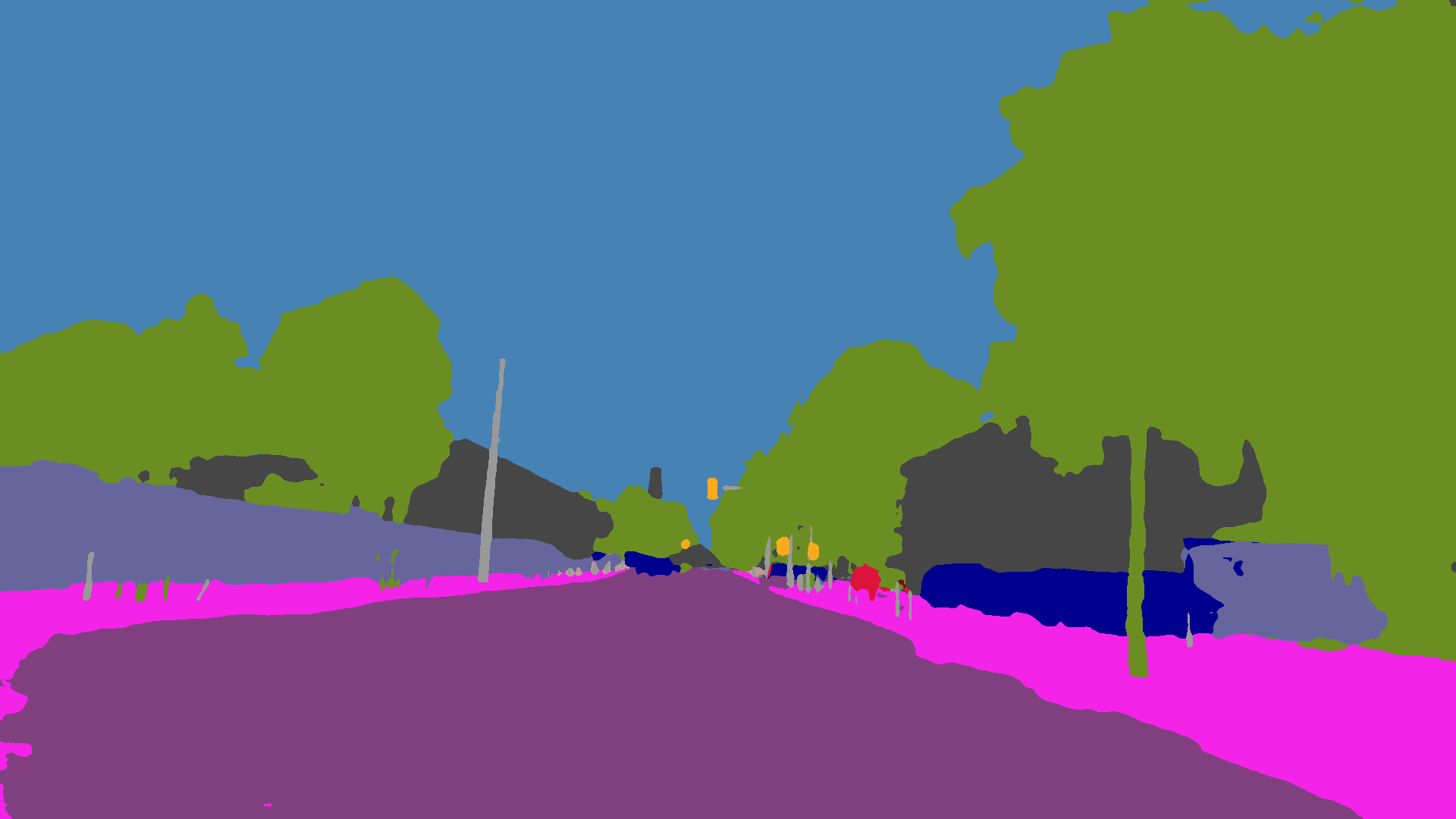}
			& \includegraphics[width=7em, valign=m]{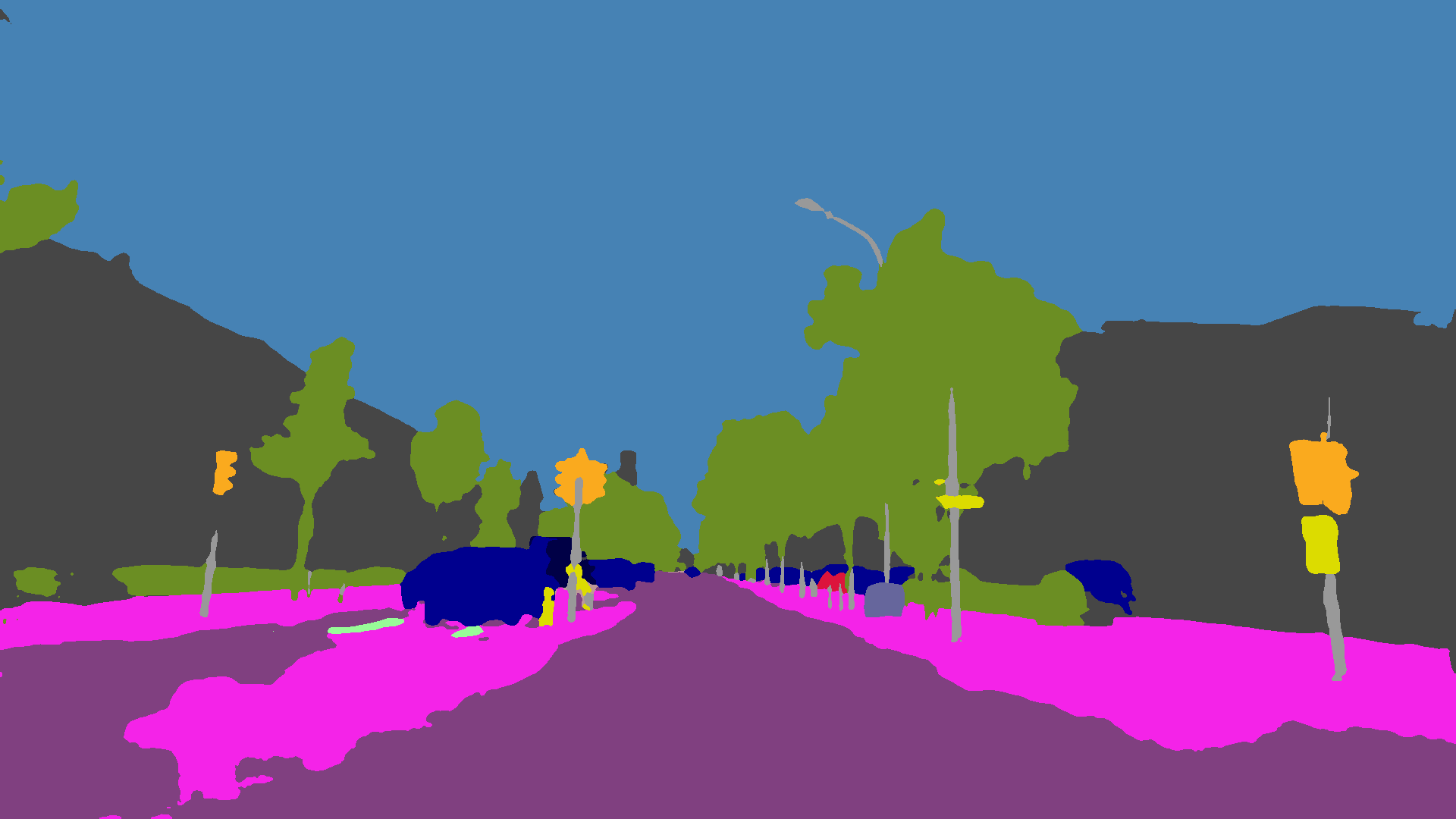}
			& \includegraphics[width=7em, valign=m]{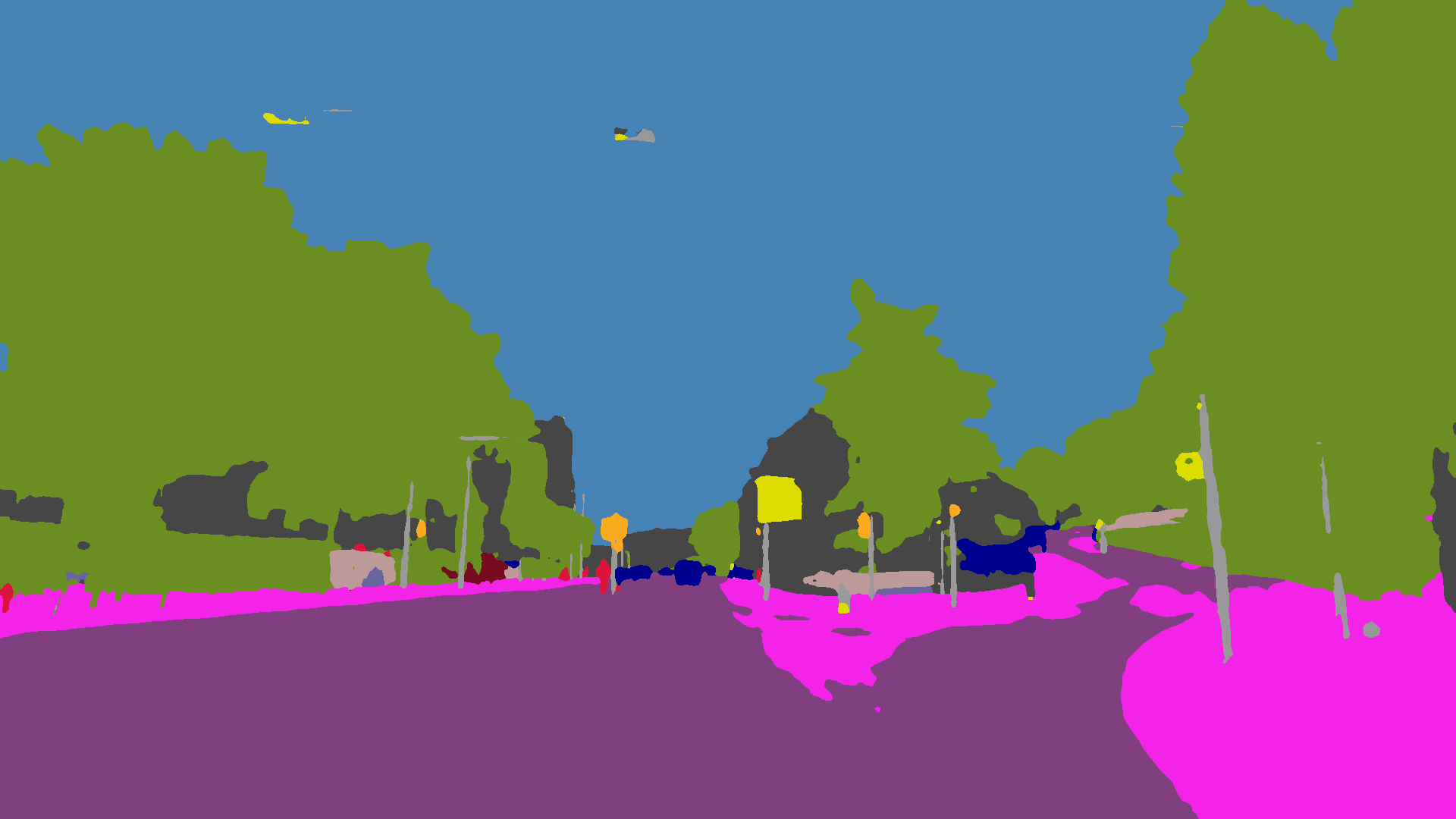}\vspace{0.3em}\\
			
			& \adjustbox{valign=m}{{\rotatebox{90}{Ours}}}
			& \includegraphics[width=7em, valign=m]{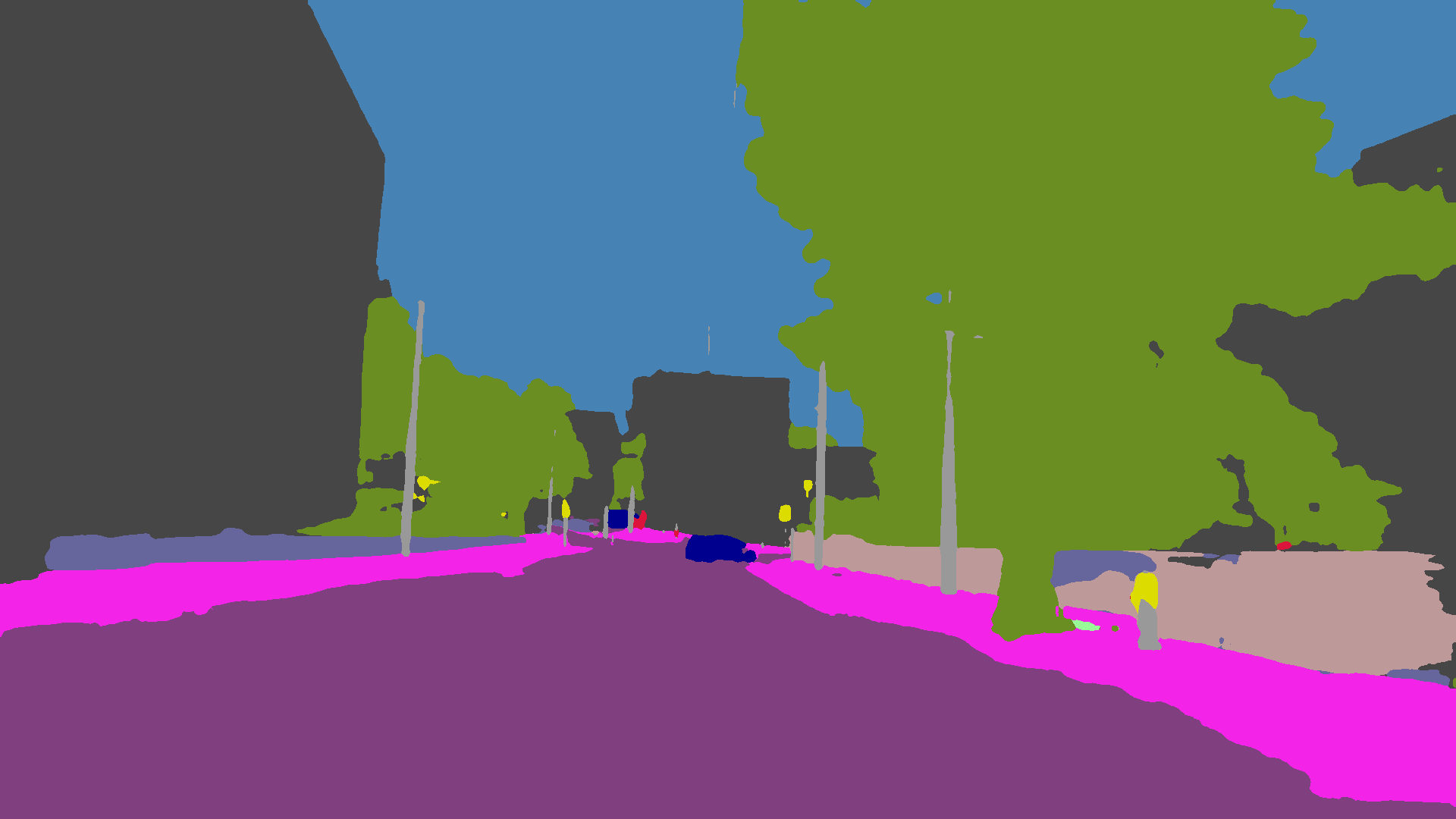}
			& \includegraphics[width=7em, valign=m]{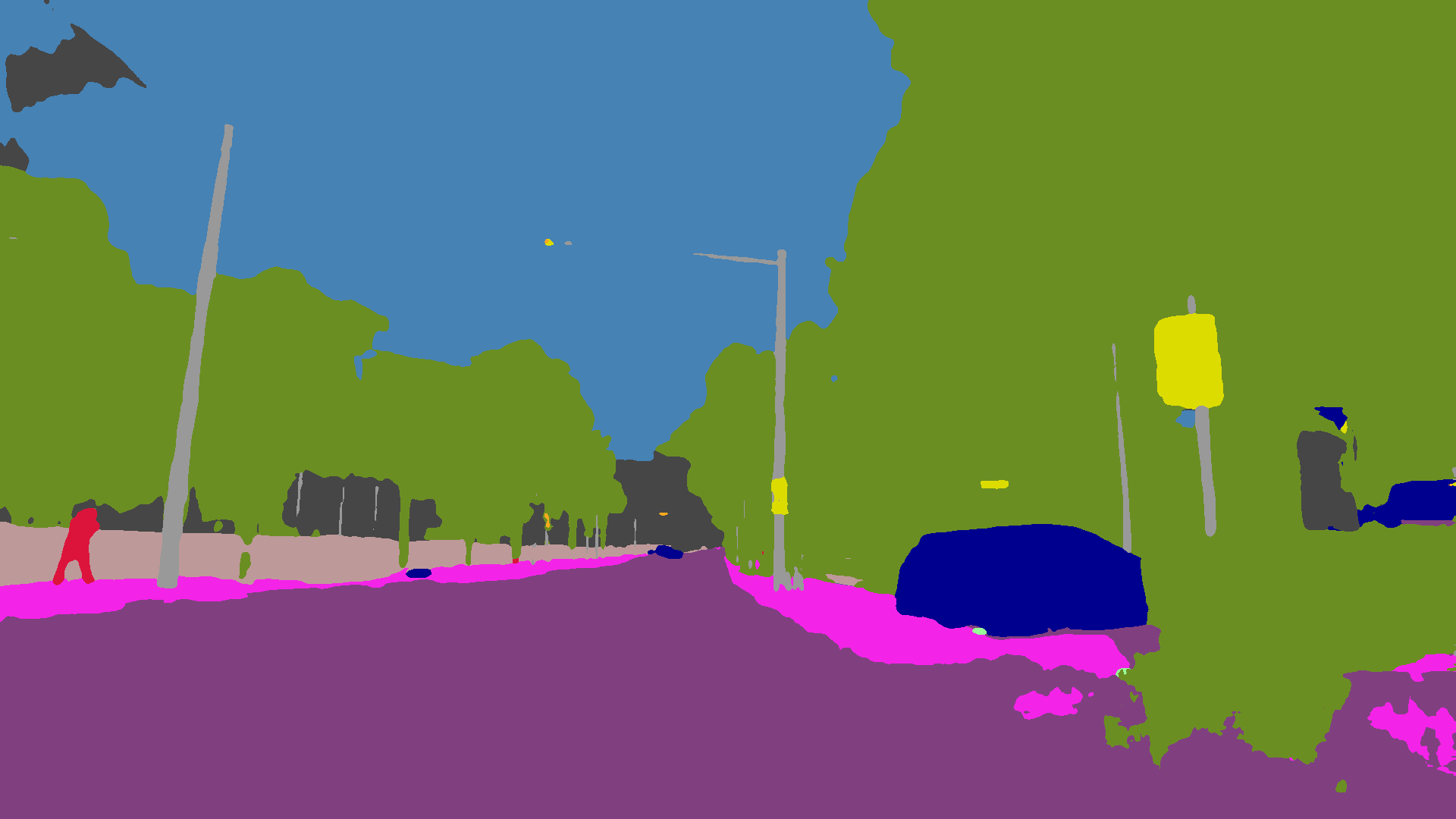}
			& \includegraphics[width=7em, valign=m]{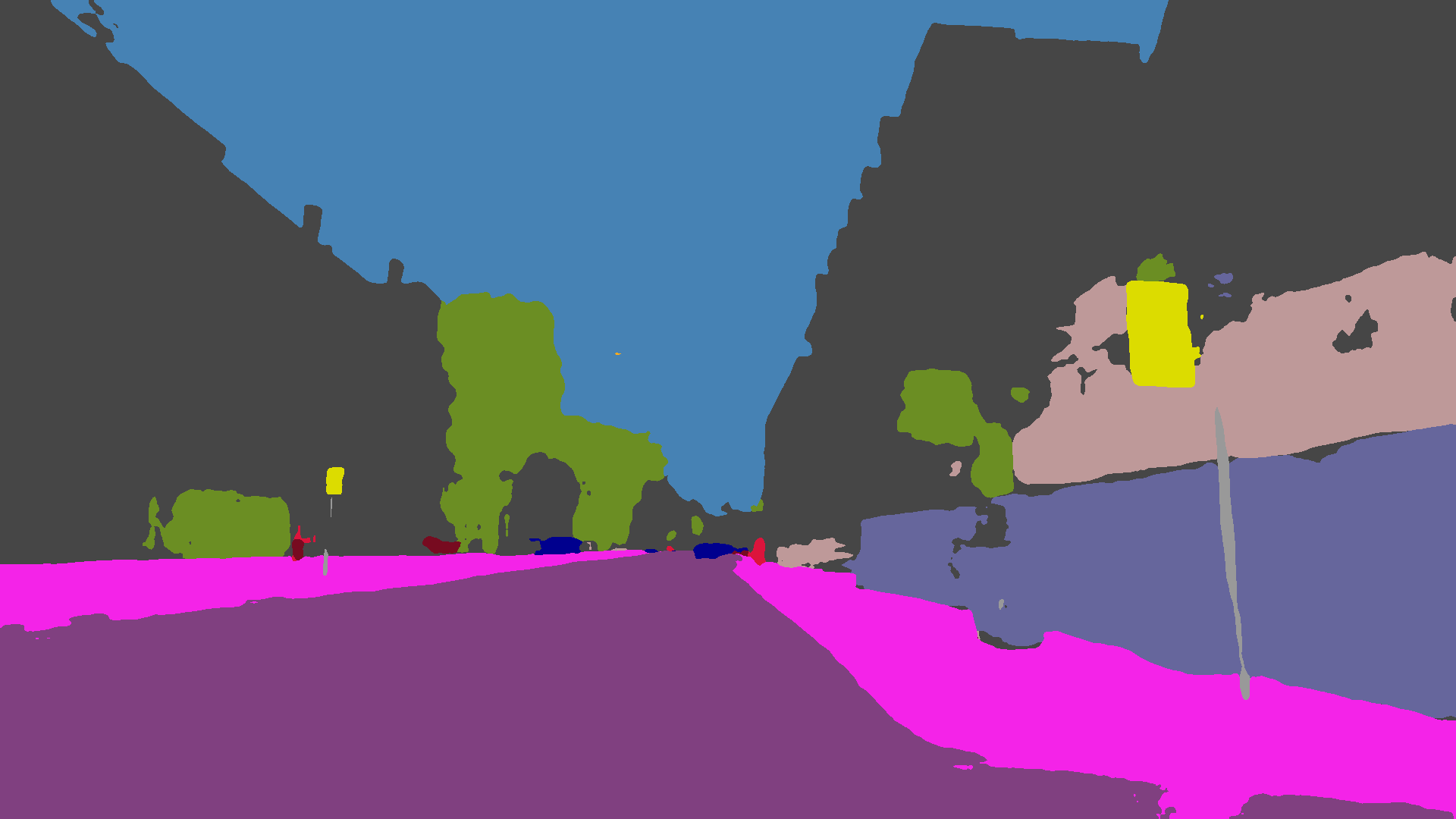}
			& \includegraphics[width=7em, valign=m]{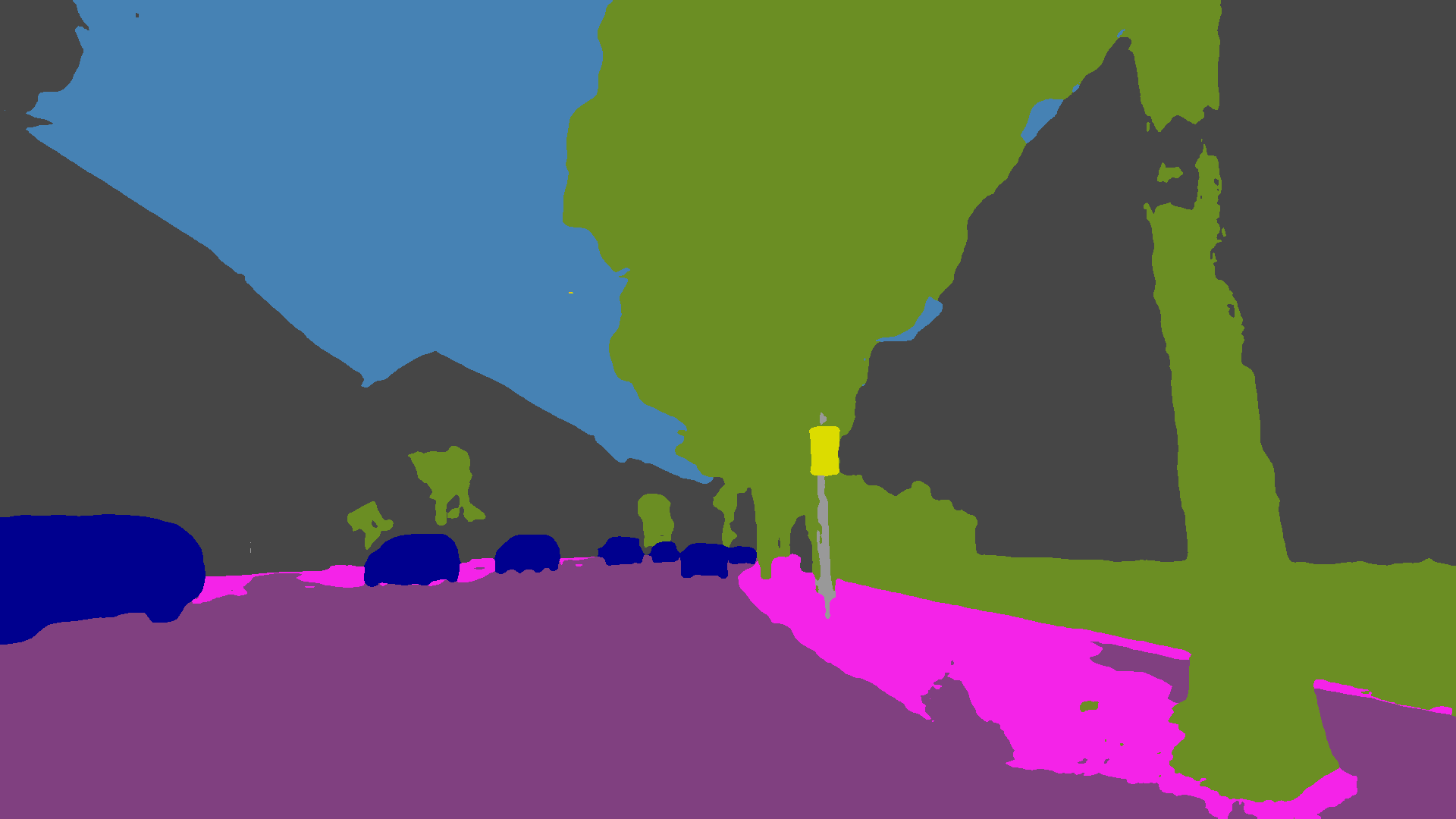}
			& \includegraphics[width=7em, valign=m]{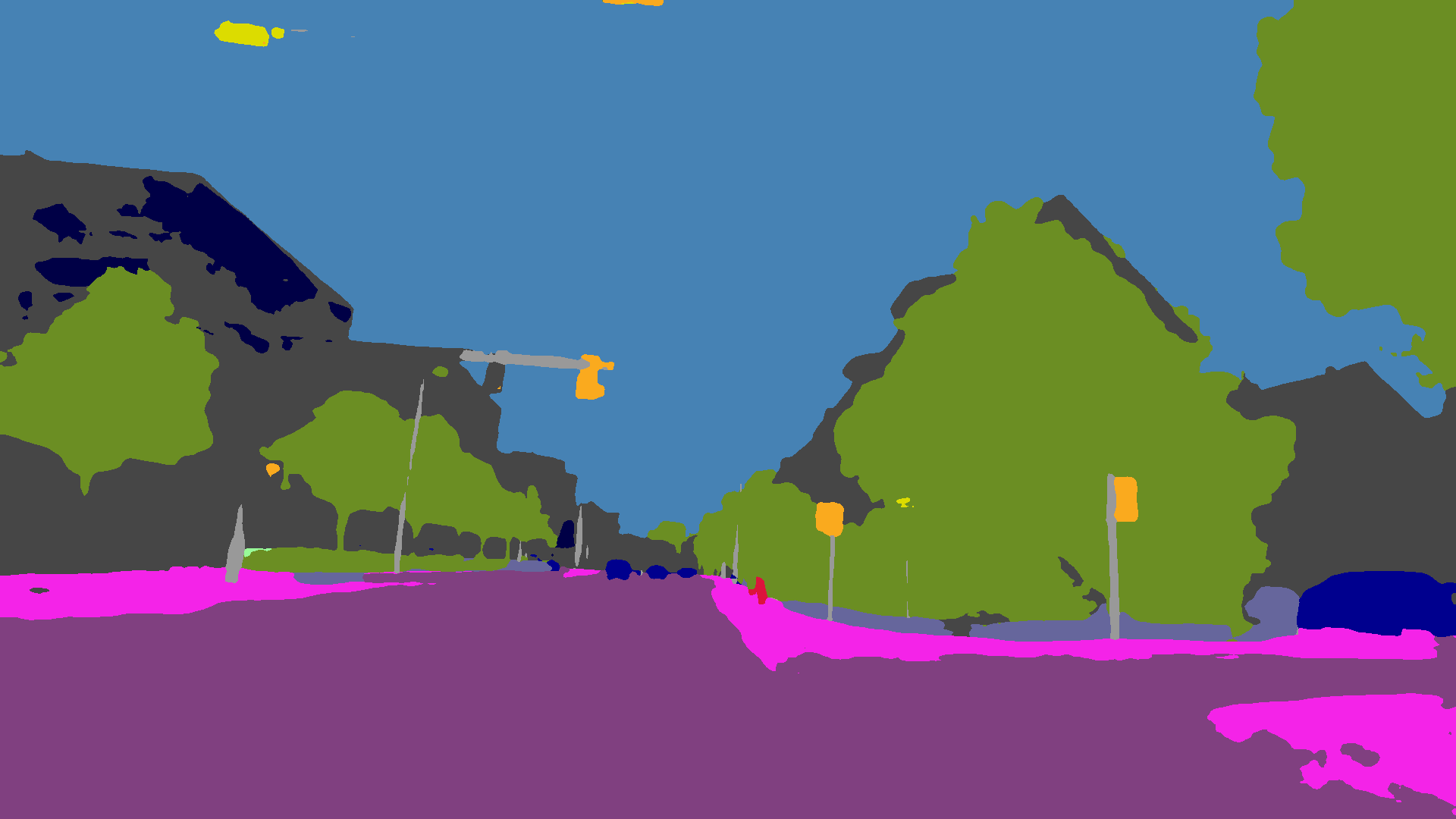}
			& \includegraphics[width=7em, valign=m]{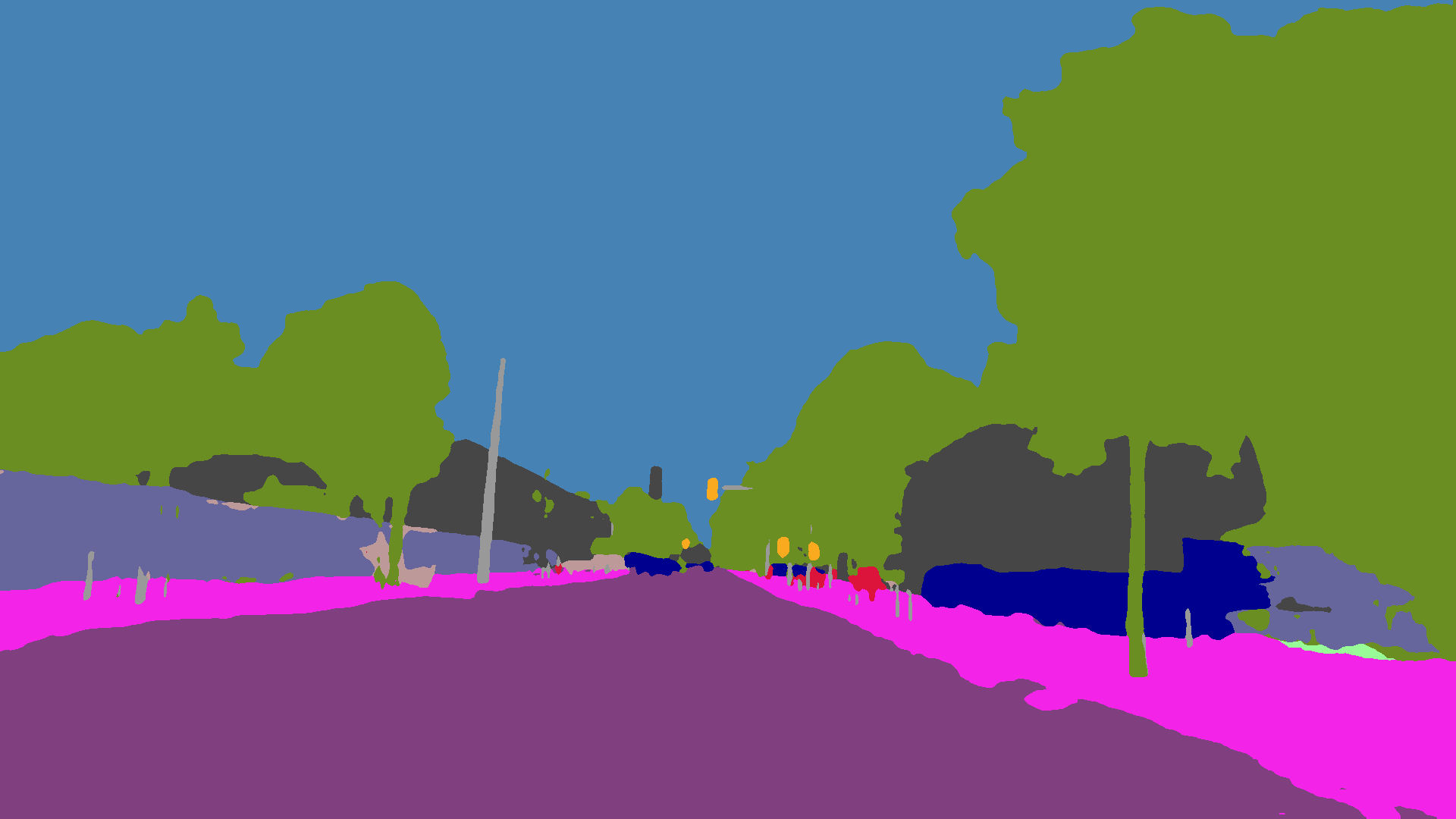}
			& \includegraphics[width=7em, valign=m]{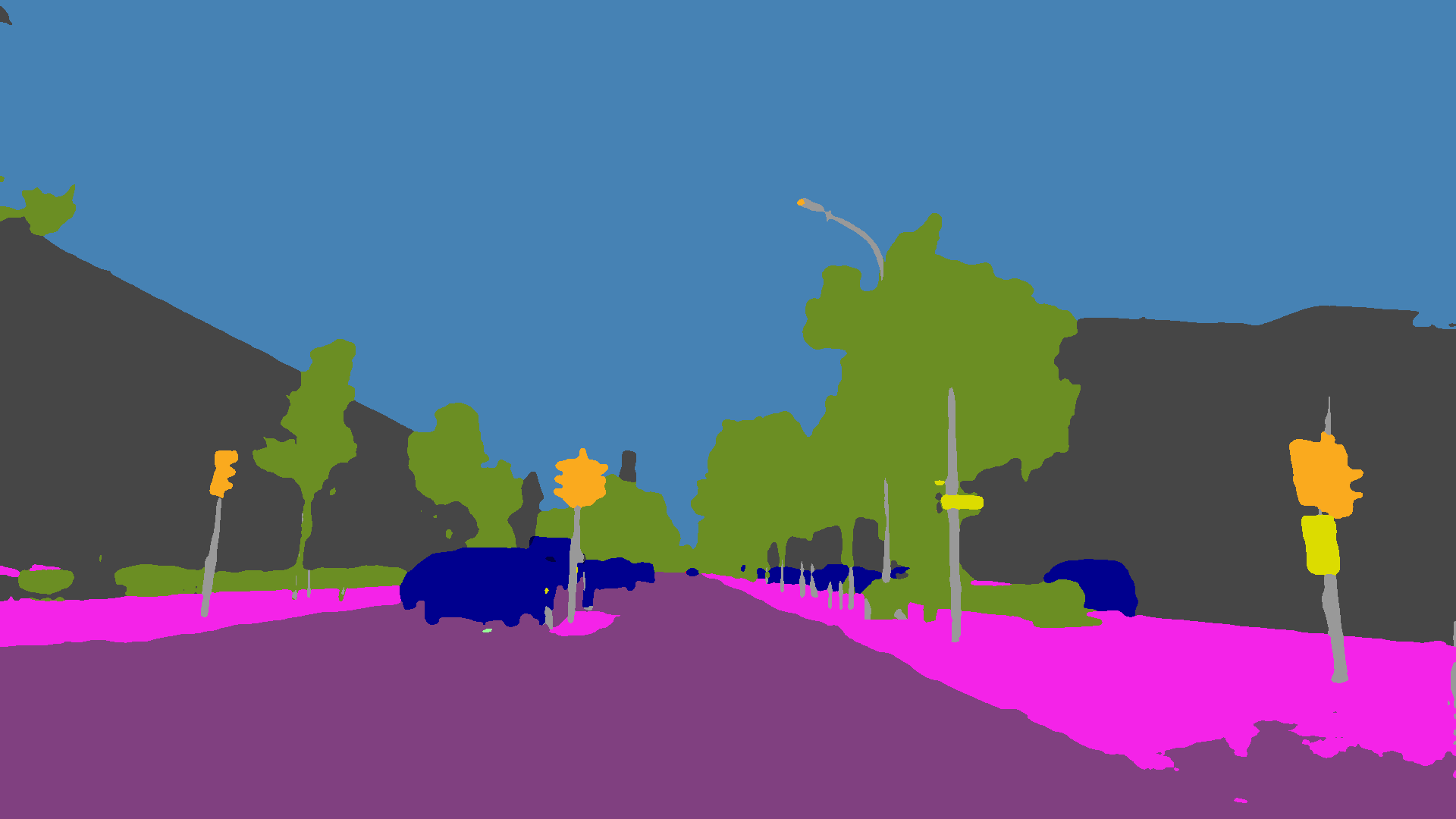}
			& \includegraphics[width=7em, valign=m]{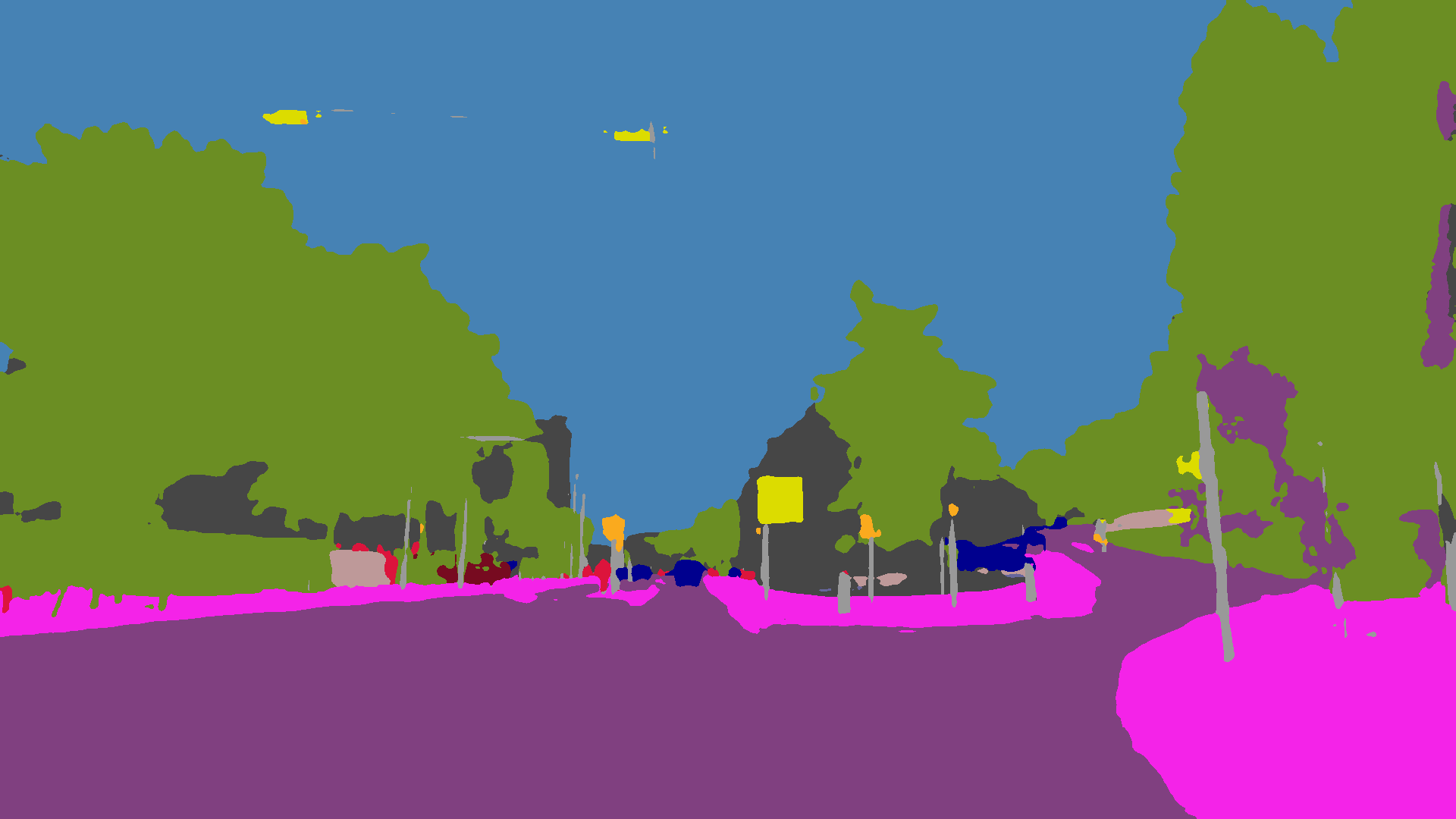}\vspace{0.3em}\\
			  \bottomrule

	\end{tabular}}
	\caption{\textbf{Additional qualitative results of semantic segmentation on ACDC$_\text{snow}$.~\cite{sakaridis2021acdc}.} Using our generated images with added snow during training improves resistance of existing segmentation networks (DeepLabv3+~\cite{chen2018encoderdecoder}, PSANet~\cite{Zhao2018PSANetPS} and OCRNet~\cite{YuanCW20}) on snowy scenarios.} \label{fig:qualit-semanticseg_supp}
\end{figure*}

Additional proxy tasks qualitative results are shown in Fig.~\ref{fig:qualit-lanedetIDD_supp} for lane detection and in Fig.~\ref{fig:qualit-semanticseg_supp} for semantic segmentation. From Fig.~\ref{fig:qualit-lanedetIDD_supp}, it is evident the robustness on degraded or even  absent street lines of lane detectors trained with our translated samples. Similarly, in Fig.~\ref{fig:qualit-semanticseg_supp} we demonstrate the improved generalization of semantic segmentation models on snowy weather conditions. In both cases, we follow the strategy described in the main paper, Sec.~4.2.3.

\end{appendices}

\end{document}